\documentclass[10pt,twocolumn,letterpaper]{article}

\usepackage[pagenumbers]{wacv} %
\usepackage{graphicx}
\usepackage{subcaption}
\usepackage{adjustbox}
\usepackage{booktabs}
\usepackage{multicol}

\definecolor{wacvblue}{rgb}{0.21,0.49,0.74}
\usepackage[pagebackref,breaklinks,colorlinks,allcolors=wacvblue]{hyperref}

\def\wacvPaperID{331} %
\def\confName{WACV}
\def\confYear{2027}

\title{Segmentation of Bovid Dentition Under Imperfect Annotations:\\A Comparative Study of Convolutional and Attention Models}

\author{Keith G. Mills\\
LSU ATHENA Lab\\
Baton Rouge, LA\\
{\tt\small keith.mills@lsu.edu}
\and
Evan B. Sanders\thanks{Summer research work with LSU as part of Episcopal High School of Baton Rouge's ESTAAR Program.}\\
LSU ATHENA Lab \& EHSBR\\
Baton Rouge, LA\\
{\tt\small SandersE27@ehsbr.org}
\and
Gregory J. Matthews\\
Loyola Dept. Math \& Stats\\
Chicago, IL\\
{\tt\small gmatthews1@luc.edu}
\and 
Juliet K. Brophy\\
LSU Dept. Geography \& Anthropology\\
Baton Rouge, LA\\
{\tt\small jbrophy@lsu.edu}
}

\begin{document}
\maketitle

\begin{abstract}

Semantic segmentation decomposes an image into distinct mask regions corresponding to different object categories, such as people, cars, signs or buildings. Advances in machine learning (ML) have shifted this task away from traditional rule-based heuristics such as edge detection, towards deep neural networks (DNN) that learn to classify pixels directly. However, semantic segmentation DNNs crucially depend on expertly designed mask targets to learn from, and imperfect or misaligned masks can interfere with a model's ability to learn effectively. 

This paper presents a comparative study of segmentation architectures, ranging from convolutional backbones to vision transformers, applied to the B.O.V.I.D. dataset, a corpus of high-resolution bovid dental photographs paired with hand-made segmentation masks not originally designed for ML-based training. We evaluate a range of preprocessing and alignment techniques to mitigate the resulting label imperfections. We find that while these preprocessing choices have limited effect on quantitative metrics such as Dice score and mIoU, their qualitative impact on predicted masks is substantial. 

\end{abstract}

\section{Introduction}

Paleoanthropologists often use the animals associated with human ancestors in order to reconstruct their past environment. Specifically, researchers rely on the teeth of animals in the Family Bovidae because these animals have relatively strict ecological tendencies. Bovid teeth, in particular isolated teeth, are also abundant in the fossil record. In order to help taxonomically identify these teeth, previous studies have relied upon morphometric analyses~\cite{brophy2021comparative} on the form (shape and size) and shape of the occlusal, or chewing, surface in order to differentiate between the different taxonomic tribes and species~\cite{brophy2014quantitative, brophy2024classification}. 

A photograph of the tooth of a known species was manually scaled and converted into a black and white image, binarized, using programs such as GIMP. The teeth were statistically analyzed and shown to differentiate from each other based on tribe, genus, and species. The success of this approach led to the application of this method to the fossil record. Using machine learning, the known teeth were compared to fossil teeth in order to look for fossil representatives of their modern counterparts and fossil relatives. The identifications of the fossil teeth have been used in paleoenvironmental reconstruction. 
The process of digitizing the teeth to binarize them has been performed by placing points around the occlusal surface of a tooth~\cite{brophy2014quantitative}. More points could be placed in areas where the shape is more complex. Once the outline was complete, the shape was filled in with black and the outside was converted to white. This method, while effective, is quite time consuming and creates imperfections compared to a ML-geared semantic segmentation dataset~\cite{Cordts2016Cityscapes, zhou2017ADE}. 

The purpose of this paper is to investigate to what kind of machine learning solution is required to adequately recognize and isolate the bovid tooth in a raw, color image, and, with proper training, create a binarized image needed for analysis. As such, this paper performs a comparative study of different segmentation model designs~\cite{iakubovskii2019segmentation} and rule-based processing heuristic techniques~\cite{pizer1990clahe, gonzalez2009image} in order to combat these issues.  Our detailed contributions are as follows:

We first compare a breadth of semantic segmentation architectures, ranging from traditional ResNets~\cite{he2016deep} to more lightweight MobileNets~\cite{sandler2018mobilenetv2}, EfficientNets~\cite{tan2019efficientnet} and Vision Transformer-based~\cite{dosovitskiy2020image, xie2021segformer} backbones to profile the efficacy of different encoder designs across architecture type, model size and technical complexity. 

Second, we conduct our experiments on B.O.V.I.D.~\cite{brophy2022bovid}, a small dataset consisting of high-resolution bovid dental photos and black/white masks which we cast as a semantic segmentation task. A key challenge to working with this dataset is that %
the masks are handmade without the original intention of being used as training data, and thus contain imperfections which make traditional ML-based generalization techniques cumbersomely difficult if applied in a straightforward manner. To combat this, we consider a slew of preprocessing techniques such as downsampling, correction, and histogram equalization~\cite{pizer1990clahe} alongside a range of different backbone feature extractor types.

Third, we compare and contrast our best methods and techniques on wholly new and out-of-distribution bovid dentation data to demonstrate the generalizability and efficacy of our approach towards labeling future data samples. Moreover, we provide a taxonomic analysis of segmentation performance across different tribes of Bovidae species. 

Experimental results demonstrate a surprising finding in the segmentation ability of different encoders. Specifically, we find that while using different preprocessing techniques may not show much difference in terms of quantitative metrics such as Dice score and mIoU, the qualitative impact is much more substantial. Further, we find that lightweight, moderately complex CNN encoders such as MobileNetV2~\cite{sandler2018mobilenetv2} are surprisingly effective in their ability to handle this task, and are comparable to much newer, attention-drive SegFormer~\cite{xie2021segformer} encoders that are several times their computational size. Finally, we provide some insights on segmentation ability across different Bovidae tribes.

\section{Bovidae Extant Dataset}
\label{sec:dataset}

\begin{figure}[t!]
    \centering
    \includegraphics[width=0.3\columnwidth]{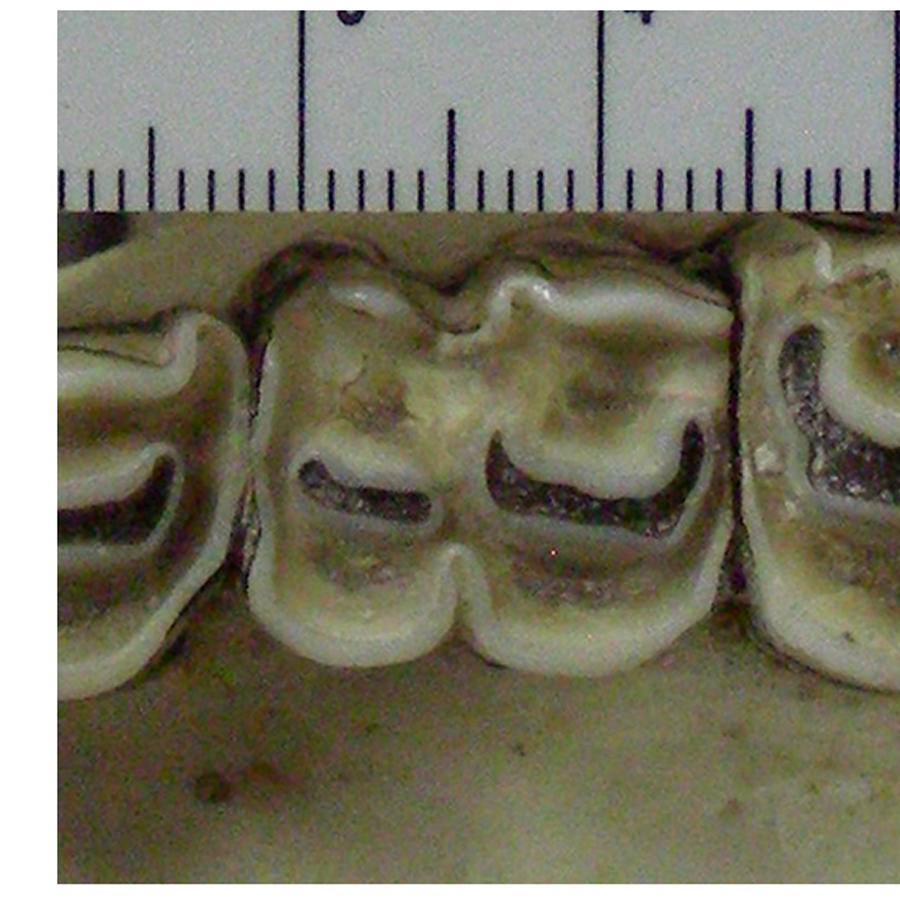} 
    \includegraphics[width=0.3\columnwidth]{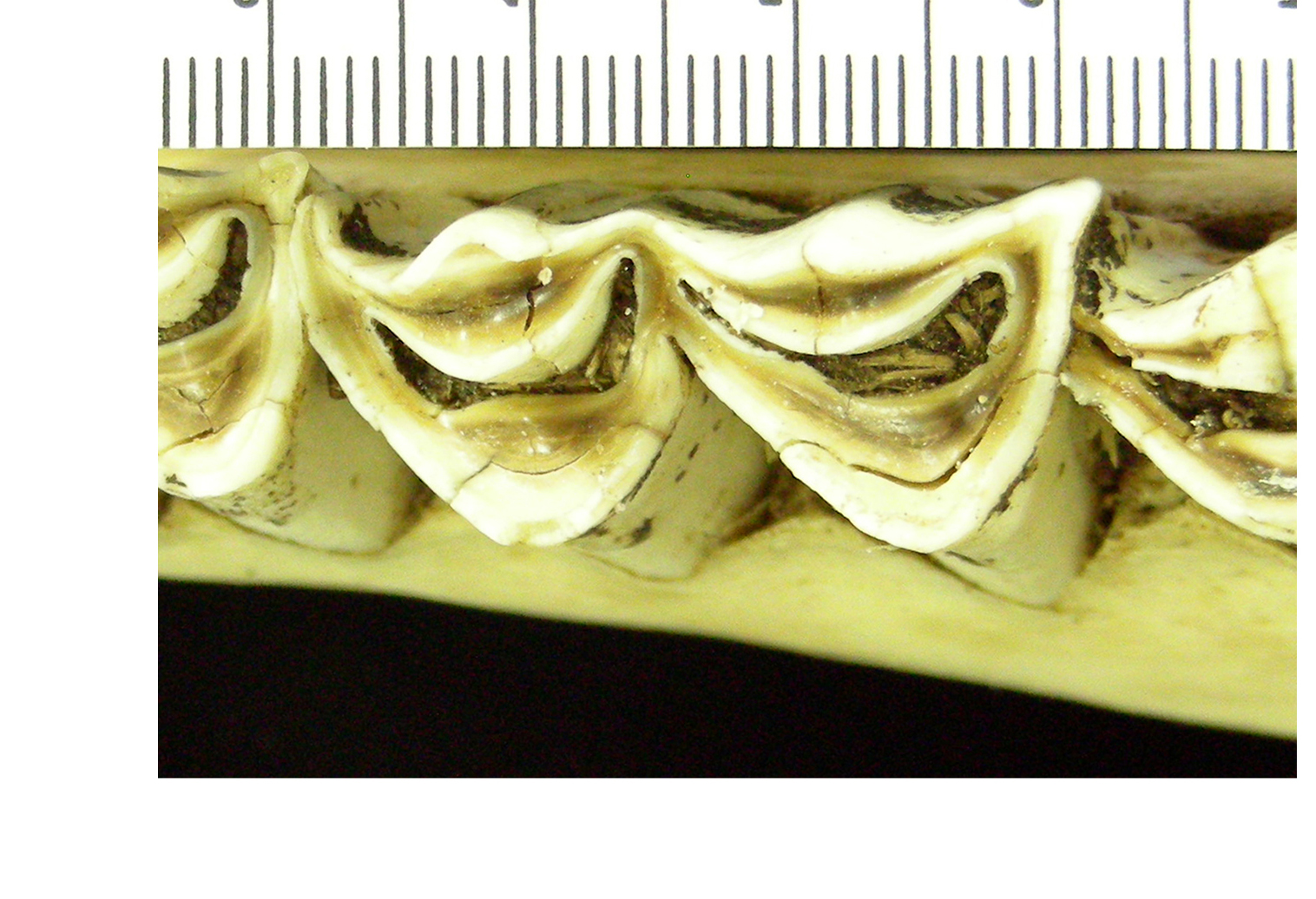}
    \includegraphics[width=0.3\columnwidth]{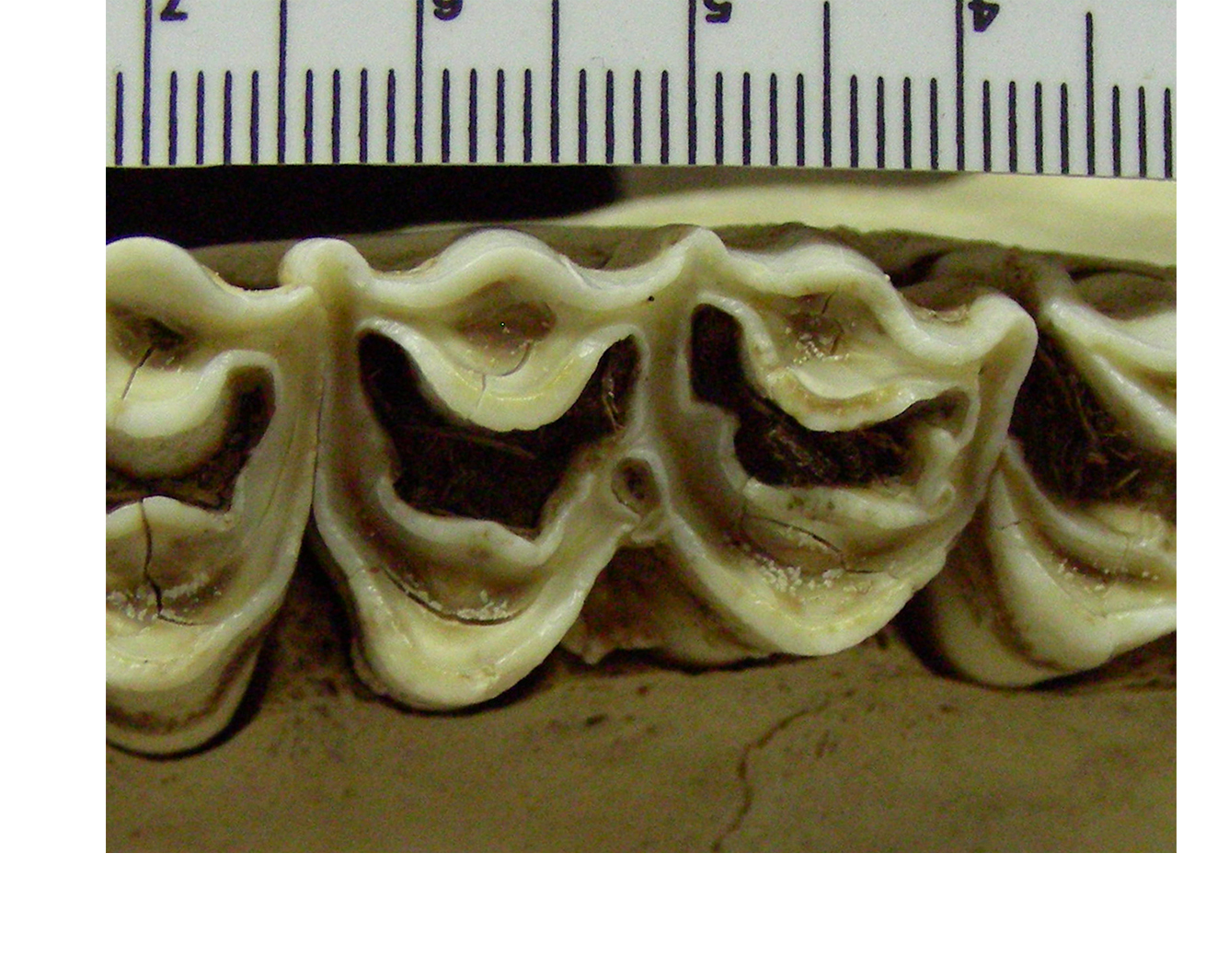}\\[-1mm]
    \includegraphics[width=0.3\columnwidth]{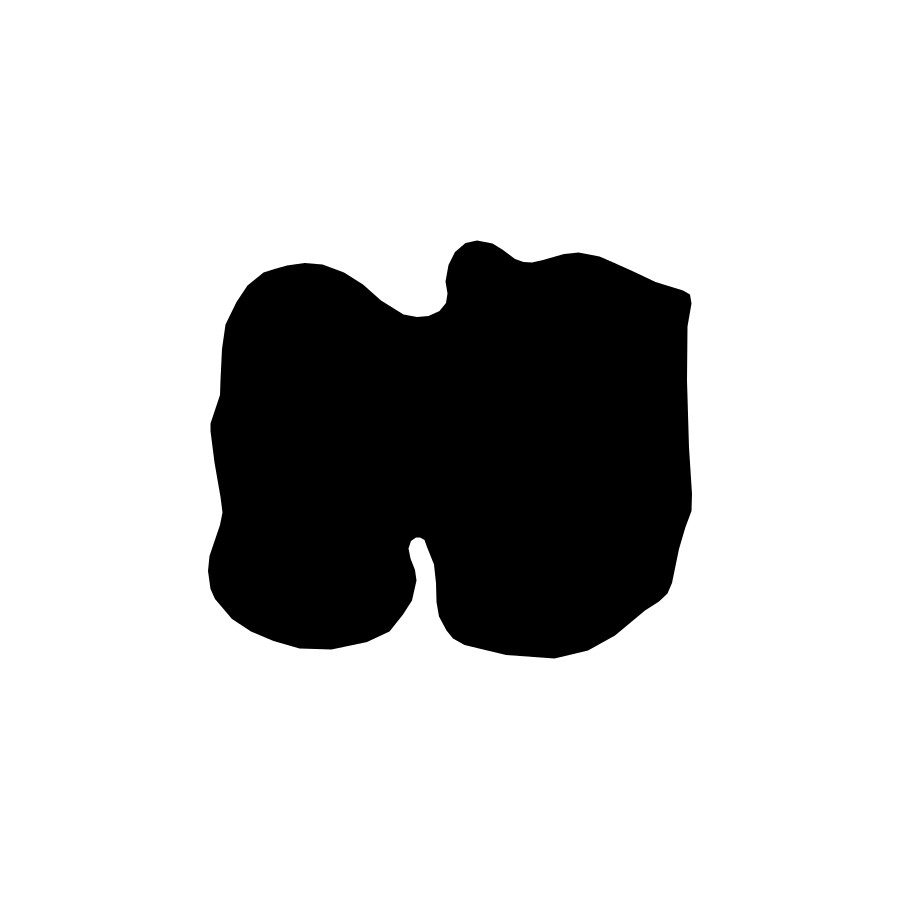}
    \includegraphics[width=0.3\columnwidth]{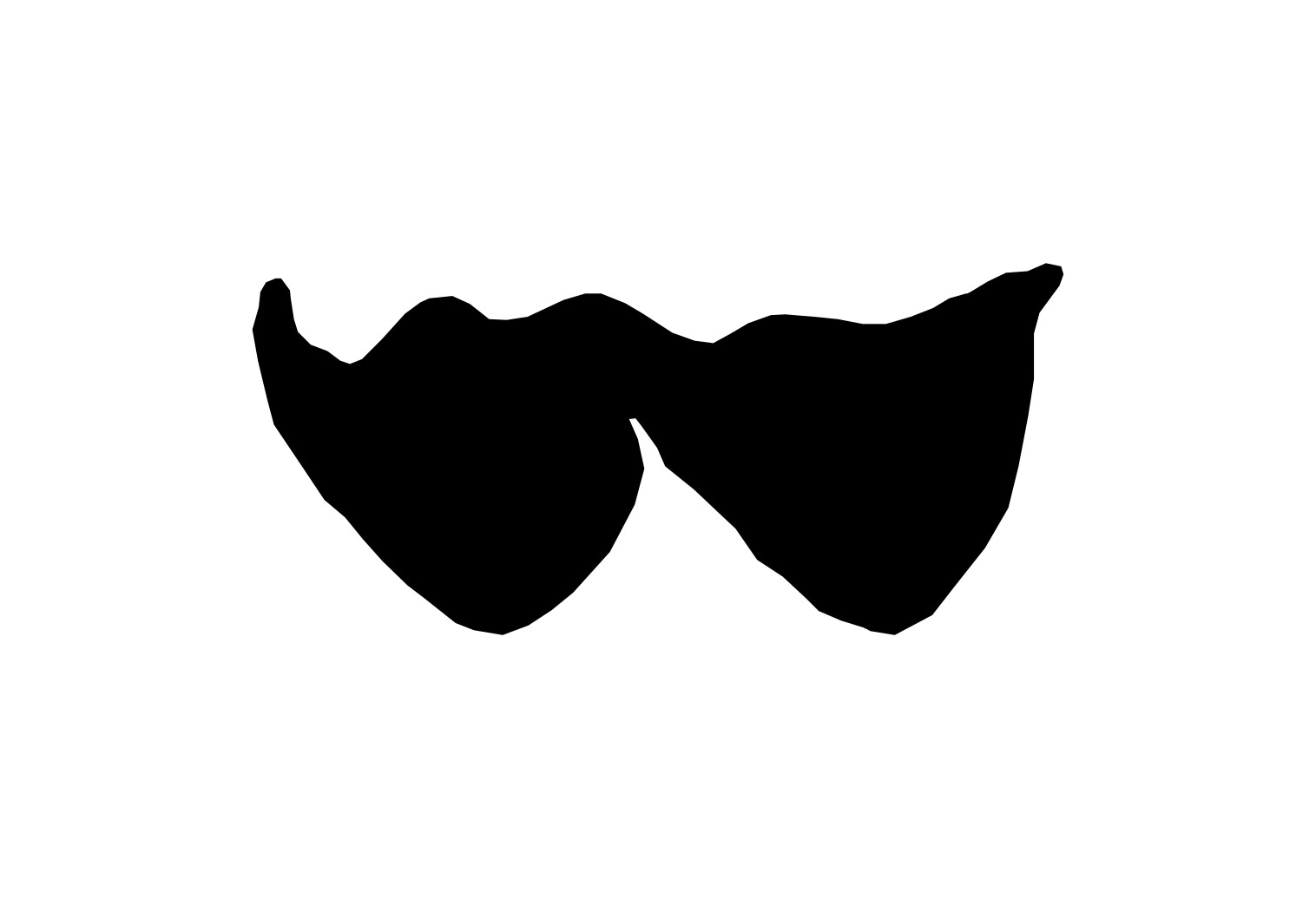}
    \includegraphics[width=0.3\columnwidth]{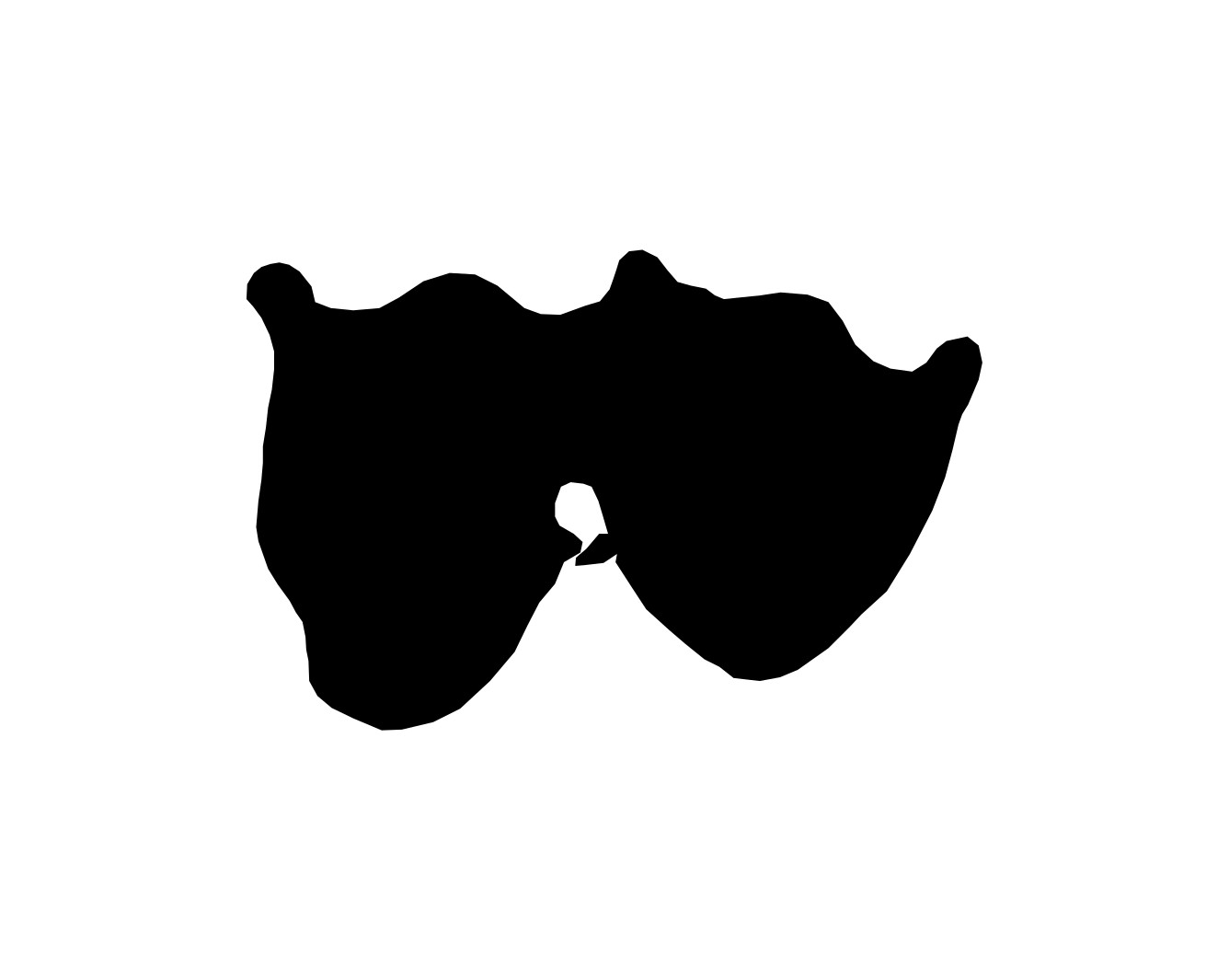}
    \caption{Example (raw, mask) image pairs from the B.O.V.I.D. dataset. Note how for each image, the masked region denotes the specific tooth in question although there may be other teeth shown.}
    \label{fig:dataset_examples} 
\end{figure}

We introduce and discuss the segmentation dataset considered in this paper. Specifically, we describe the original dataset on a high-level, the challenges imposed performing segmentation on the dataset, as well as how we prune it for suitability training and testing an ML segmentation model. 

\subsection{Dataset Characteristics}
In this paper we consider the B.O.V.I.D. dataset~\cite{brophy2022bovid} which consists of 3592 extant Bovidae teeth images of varying sizes and resolutions from 7 tribes: Alcelaphini, Antilopini, Bovini, Hippotragini, Neotragini, Reduncini and Tragelaphini. The average image resolution is $(1288 \times 1430) \pm (351 \times 492)$ pixels, with the smallest being $(407 \times 238)$ pixels while the largest image is $(5712 \times 4288)$. Each tooth is associated with two images:

\begin{enumerate}
    \item \textbf{Raw Image:} Scaled color image of the occlusal surface of the tooth. These images were primarily sourced from three South African institutions: National Museum, Bloemfontein; Ditsong Museum, Pretoria; and Amathole Museum, King William's Town. Other images were sourced from the Field Museum in Chicago, Illinois, U.S.A.
    \item \textbf{Mask Image:} Black and white image of the tooth's occlusal surface. These mask images were handcrafted from the original tooth image primarily by student workers using GIMP.
\end{enumerate}

Each image includes metadata pertaining to the taxonomic Family, Tribe, Genus, and Species, as well as the original tooth side and type, i.e., first lower molar (LM1), second upper molar (UM2), tooth's current location, accession number at institution, origin, etc. Figure~\ref{fig:dataset_examples} provides a sample illustration of several raw image-mask pairs. 

In this paper the Raw Images serve as the input to a segmentation model, while the Mask Images serve as the supervised labels. Given the size of the data corpus, we utilize pre-trained off-the-shelf segmentation models and fine-tune them for this task. 

\subsection{Dataset Imperfections}

Unlike classical semantic segmentation datasets~\cite{Cordts2016Cityscapes, zhou2017ADE}, B.O.V.I.D. was not initially 
created with ML-driven segmentation in mind. This imposes some unique challenges which make this task easier said than done.

\noindent\textbf{Resolution Mismatch.}
There are several instances where the resolutions in a given (Raw Image, Mask Image) pair are not the same nor even proportional to each other as the major concern when creating the dataset was DPI matching to match size components. These pairs of images are unsuitable for use in supervised ML-driven segmentation as the segmentation model will produce an output proportional to the input Raw Image, which must then also be proportional to the target Mask Image.

\begin{figure}[t!]
    \centering
    \includegraphics[width=0.45\columnwidth]{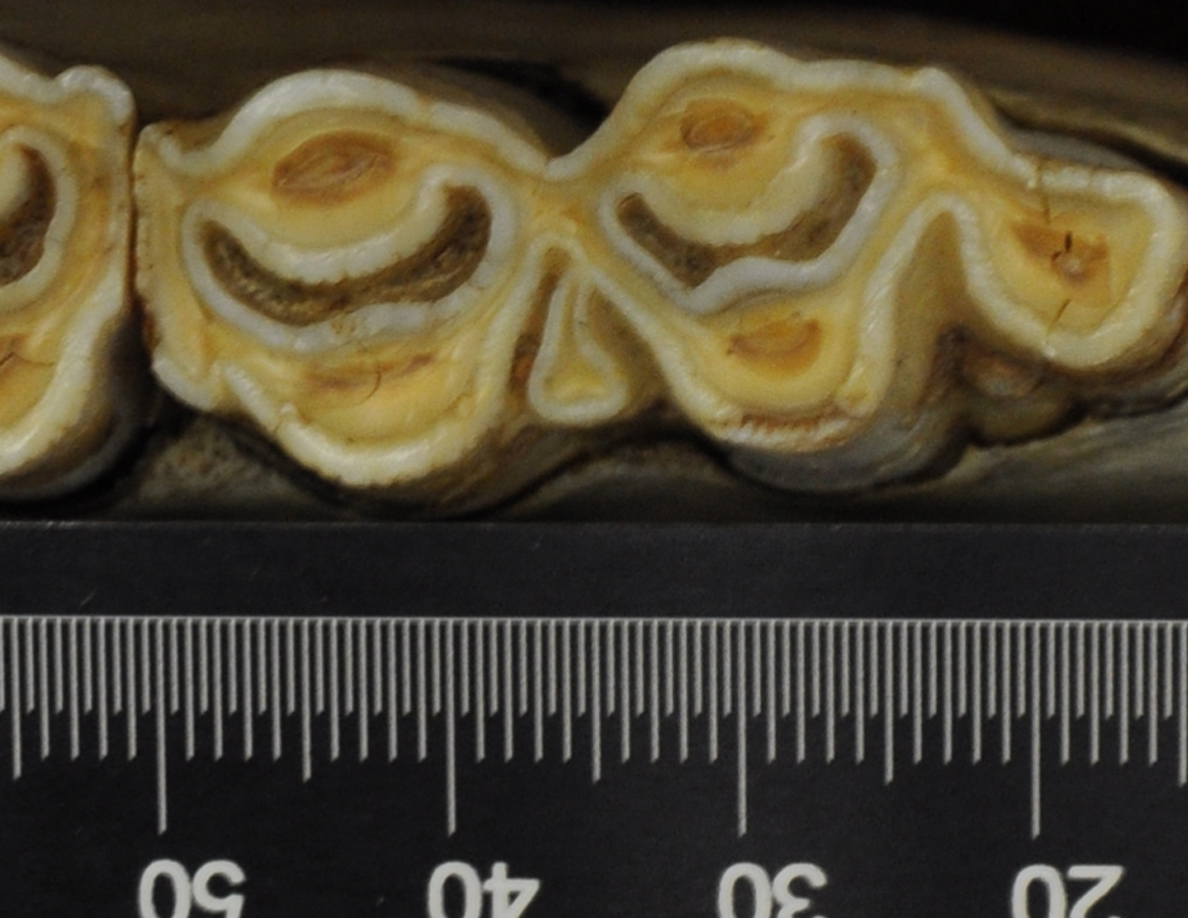} 
    \includegraphics[width=0.45\columnwidth]{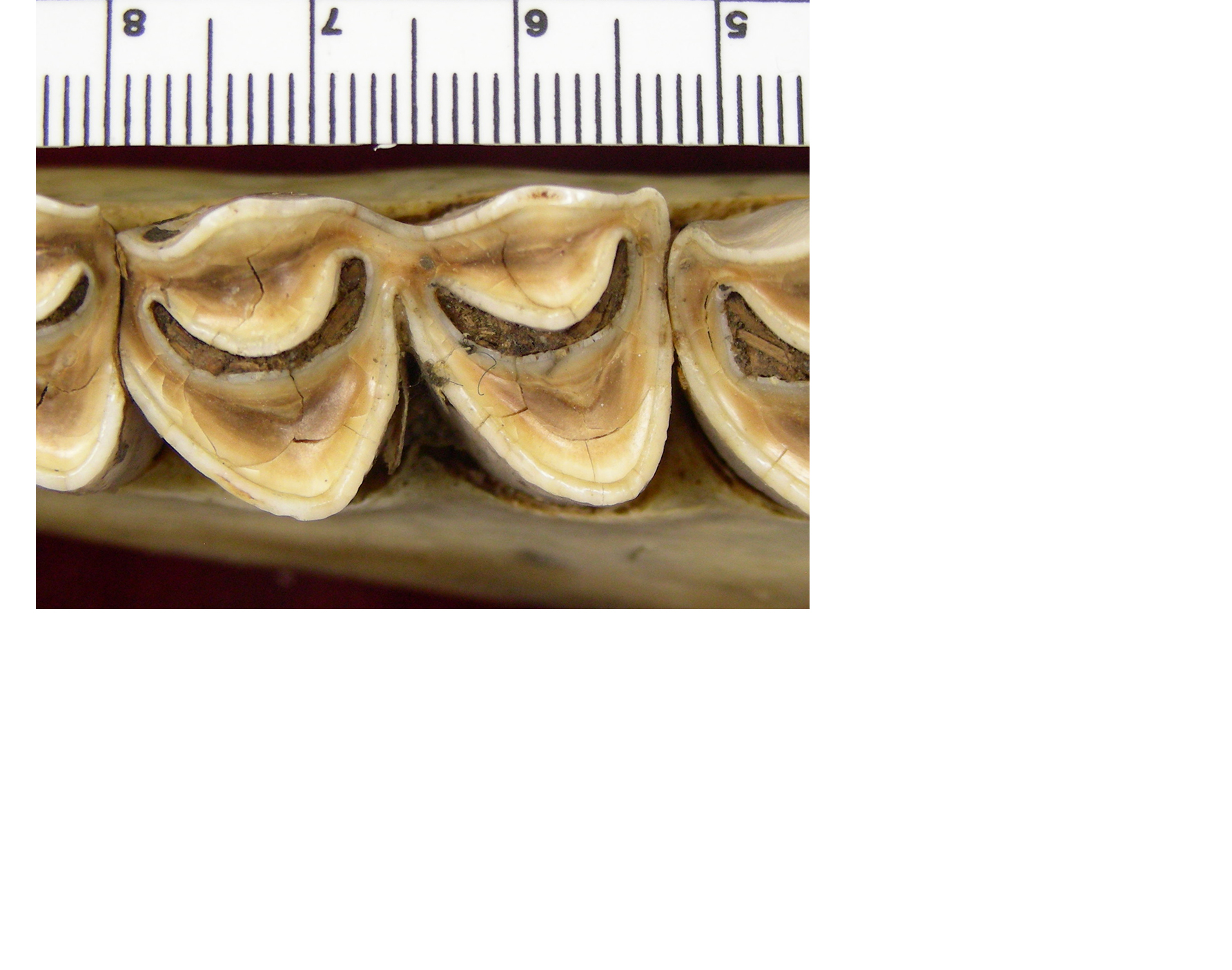}
    \\[-1mm]
    \includegraphics[width=0.45\columnwidth]{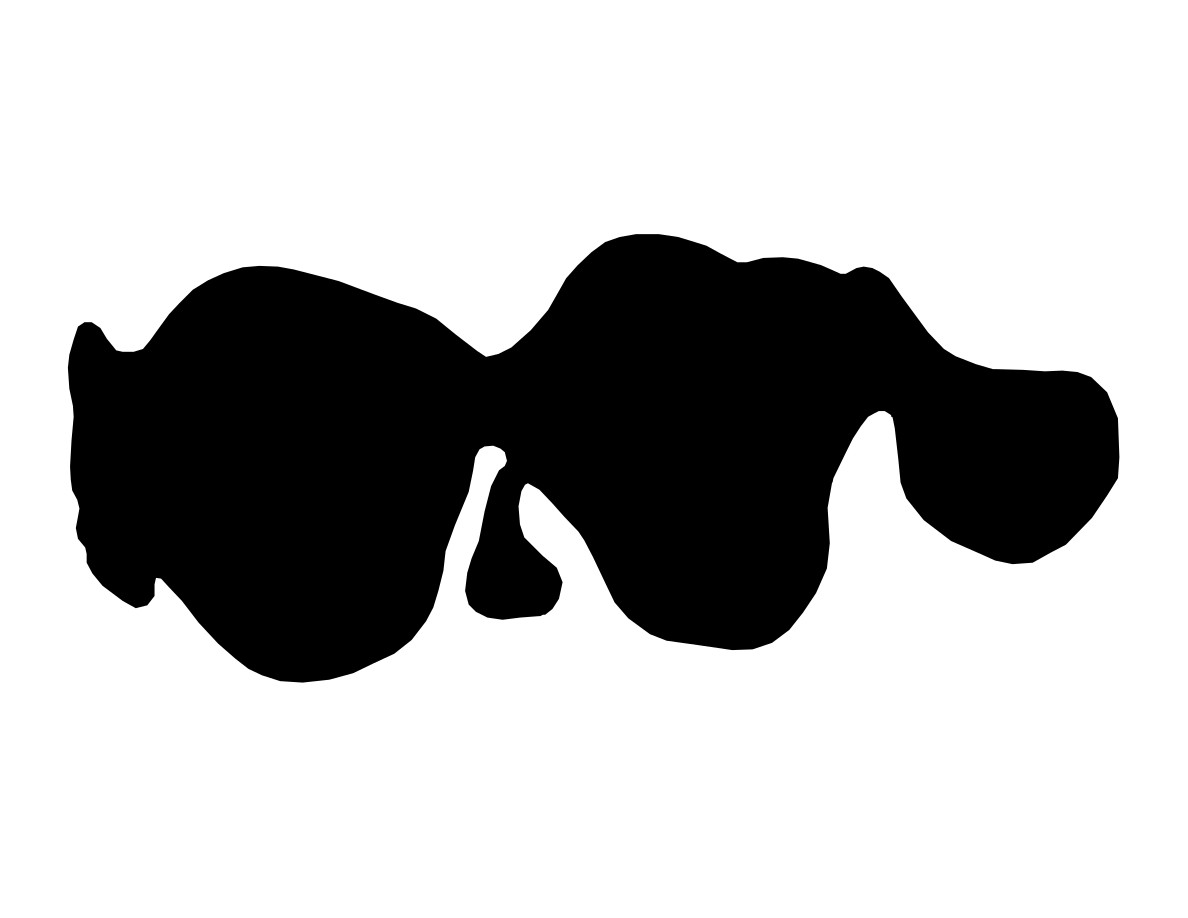}
    \includegraphics[width=0.45\columnwidth]{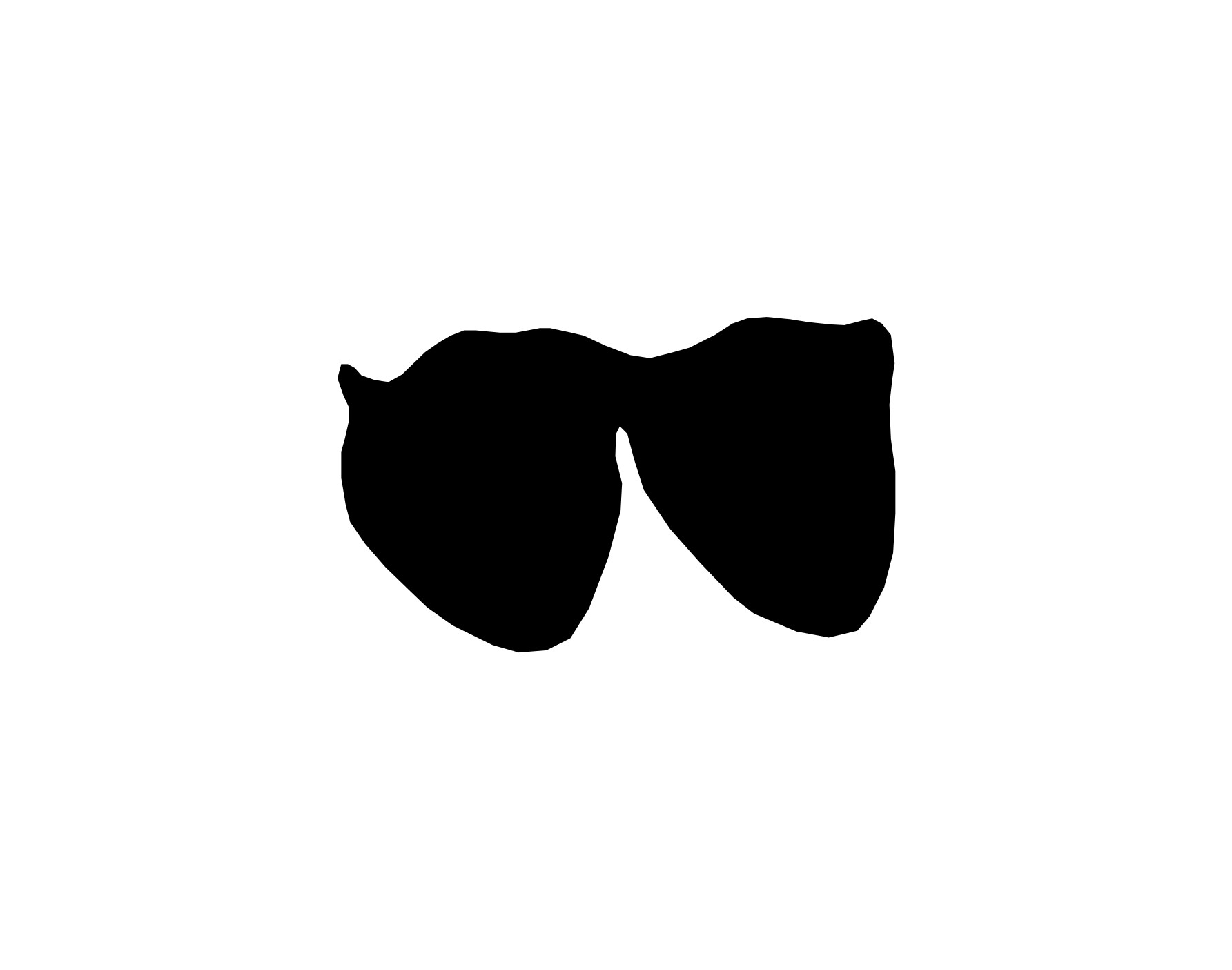}
    \caption{Examples of misaligned Raw Image, Mask Image pairs. Left-hand-side: The tooth in the raw image is in the upper third of the picture primarily, while it is centered in the mask. Right-hand-side: The raw image picture has whitespace padding surrounding it so the tooth is not in the center of the whole image, but this is the case for the mask.}
    \label{fig:misaligned_examples} 
\end{figure}

\noindent\textbf{Raw-Mask Misalignment.}
There are also cases of (Raw Image, Mask Image) pairs where, even if the images have the same resolution, the actual location of the mask in the Mask Image does not visibly align with the actual tooth in the Raw Image. Figure~\ref{fig:misaligned_examples} provides some sample example image pairs. This issue is more sinister than if the images are not proportional in resolution, as these (Raw Image, Mask Image) pairs can still be utilized for training, however, the spatial misalignment of the mask to the actual tooth will heavily disrupt the supervised learning process. 

The `Catch-22' to these issues is that adequate cropping one image to be proportional to another or to correct misalignment requires us to identify where the tooth and mask overlap and that is the objective we're aiming to fine-tune a segmentation model on in the first place. However, there are still some non-ML-based techniques we can utilize to alleviate these issues and improve generalizability. 

\section{Augmentations}
\label{sec:aug}

We discuss the techniques employed to train ML segmentation models on the B.O.V.I.D. dataset. First, we discuss pruning techniques to make the dataset cognizable by an ML model and perform well, before discussing different preprocessing techniques utilized to further increase segmentation performance. 

\subsection{Dataset Pruning \& Cropping}

The first step in our pruning process is to remove all (Raw Image, Mask Image) pairs where the resolutions are neither the same nor proportional to each other, i.e., they have different aspect ratios. This removes 283 image pairs, reducing the dataset down to 3325 image pairs for a reduction of less than $10\%$.

Second, to tackle the issue of misalignment in remaining pairs, raw images were spatially aligned to their corresponding binary masks using a foreground-centering procedure. Foreground pixels were identified by grayscale thresholding, a standard image-segmentation operation~\cite{gonzalez2009image}. Pixels with intensities below 240 were classified as foreground, and their centroid was calculated to estimate the object center. A rectangular crop centered on this centroid was then extracted, following the general principle of translating image content into a common spatial reference frame~\cite{zitova2003image}. 

The crop dimensions were defined as a scalar multiple $\alpha$ of the corresponding mask dimensions, with $\alpha=1.0$ in this study. Regions extending beyond the original image boundaries were filled with white background pixels. The resulting crop was resized using bilinear interpolation to match the exact height and width of its corresponding mask. This produced pixel-compatible image-mask pairs for downstream segmentation. Image loading, color conversion, resizing, and file writing were implemented using OpenCV~\cite{bradski2008learning}. %

This leaves us with a dataset of 3325 image pairs with an average image resolution is $(1308 \times 1458) \pm (338 \times 479)$ pixels, with the smallest being $(407 \times 238)$ pixels while the largest image is $(5712 \times 4288)$. This pruned and augmented dataset is the basis for all of our experiments. Next, we enumerate the online preprocessing techniques we utilize to enhance ML segmentation performance.

\subsection{Online Preprocessing and Training}

For each (Raw Image, Mask Image) pair, the Raw Image is loaded in RGB while the Mask Image is loaded in grayscale and then inverted so the masked region consists of white pixels, i.e., 

\[
y_{ij} =
\begin{cases}
1, & \text{if the grayscale mask value at } (i,j) = 0,\\
0, & \text{otherwise}.
\end{cases}
\]
Thus, the resulting masks contain one foreground channel with values in
$\{0,1\}$.

Next, we apply one of three contrast-processing modes:
\begin{enumerate}
    \item \textbf{Histogram Equalization (HE):} We convert the RGB image to the $YCrCb$ color space apply equalization to the luminance channel.
    \item \textbf{Contrast-Limited Adaptive Histogram Equalization (CLAHE):} We also convert the RGB image to $YCrCb$ and apply an equalization transform to the luminance channel using an $8 \times 8$ tile grid and a configurable clip limit $c$.
    \item \textbf{No Equalization (NE):} Equalization is not performed. 
\end{enumerate}

Next, in addition to several boilerplate or required preprocessing techniques, we optionally downsample images and masks by an integer factor $d$. Downsampling reduces GPU training and inference memory costs while also abating the effect of misalignment. Specifically, we downsample Raw Images using bilinear interpolation and downsample Mask Images by nearest-neighbor interpolation.

\begin{figure*}[t!] 
    \centering 
    \begin{subfigure}[t]{0.3\linewidth} \centering \includegraphics[width=\linewidth]{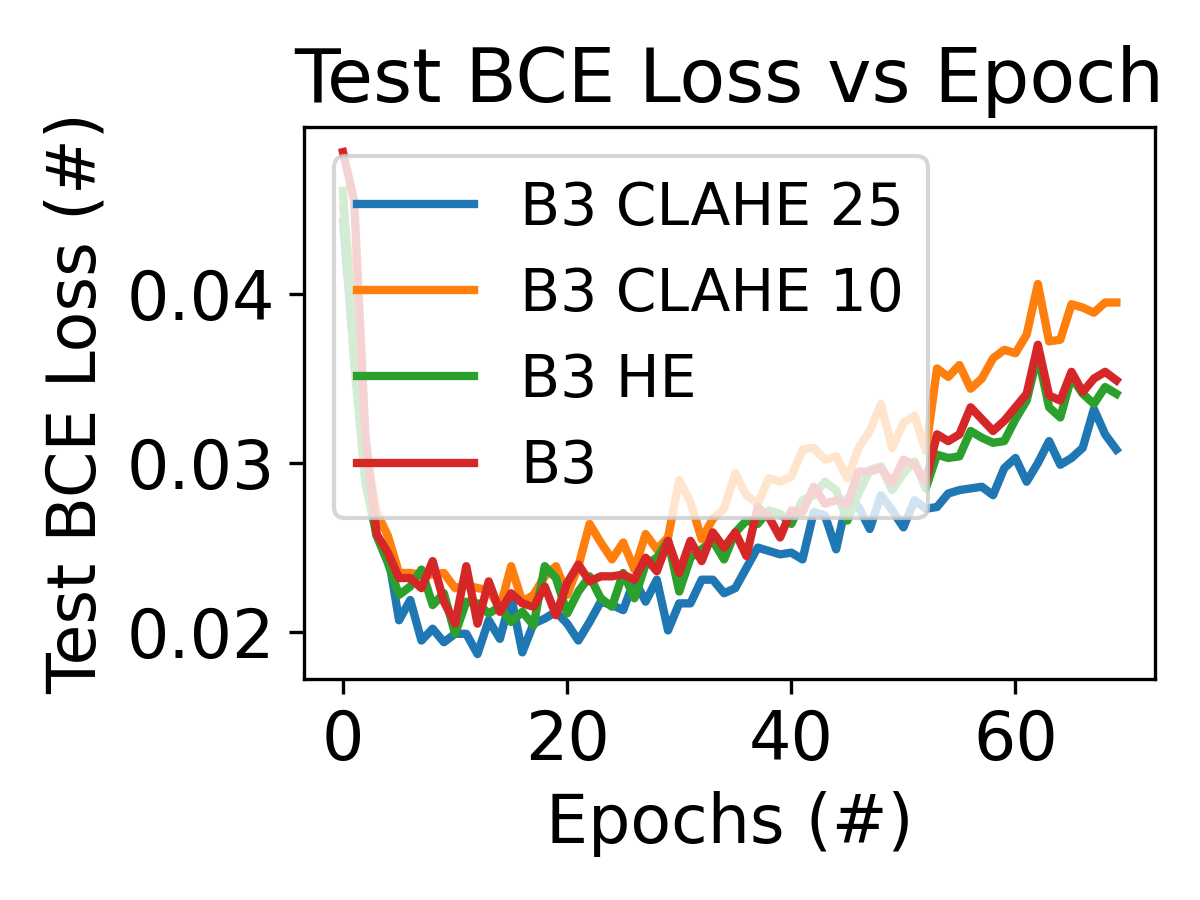} %
    \end{subfigure} \hfill \begin{subfigure}[t]{0.3\linewidth} \centering \includegraphics[width=\linewidth]{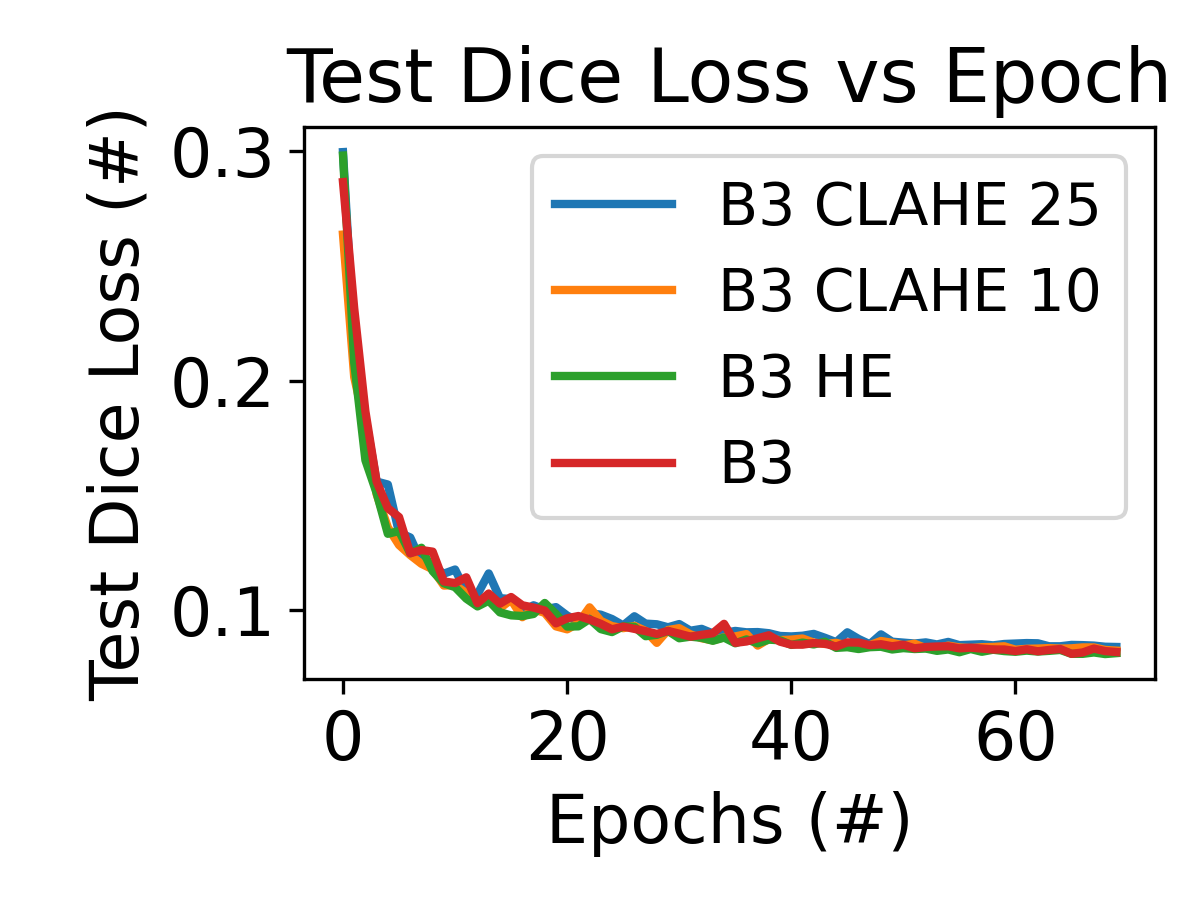} %
    \end{subfigure} \hfill \begin{subfigure}[t]{0.3\linewidth} \centering \includegraphics[width=\linewidth]{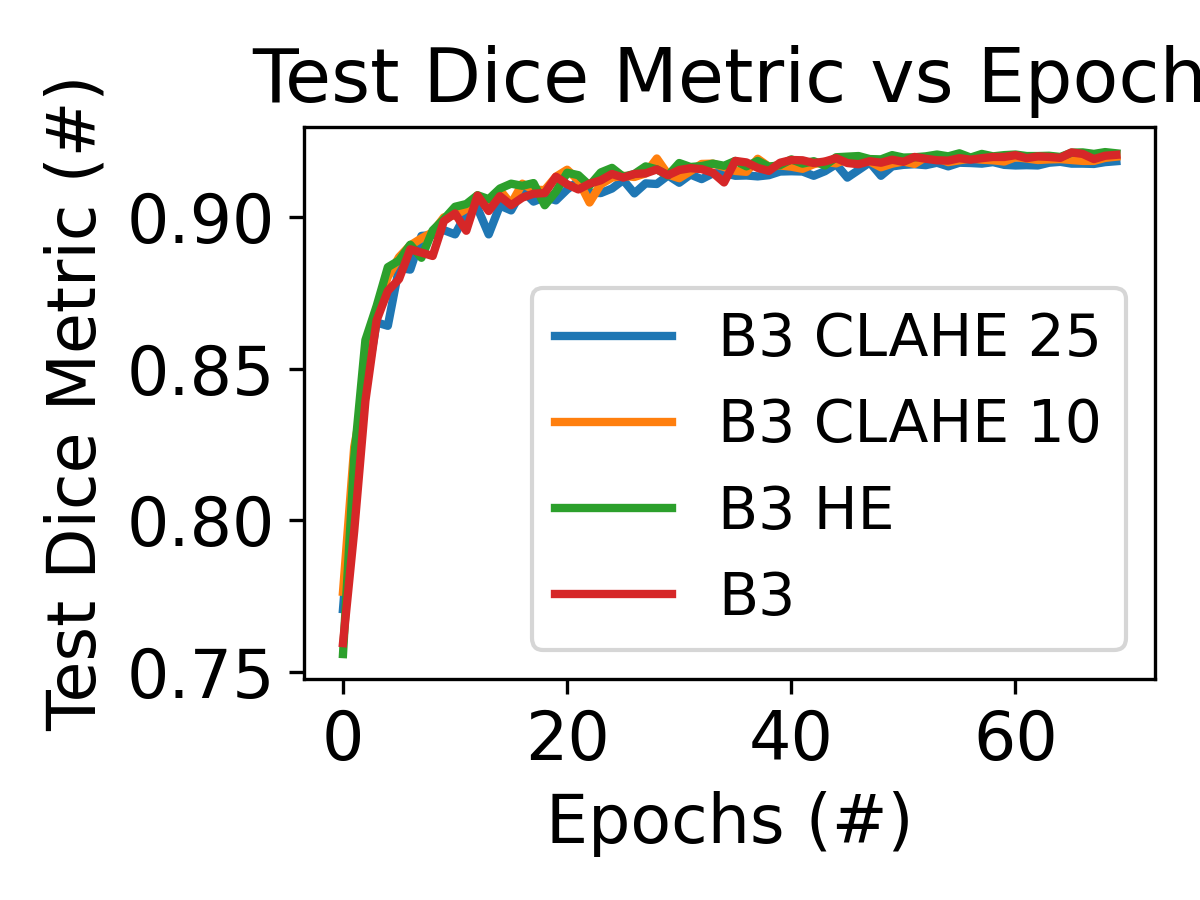} %
    \end{subfigure} \caption{BCE loss, Dice loss and Dice metric for SegFormer MiT-B3. For losses, lower is better, while higher is better for the Dice metric.} %
    \label{fig:segformer_main}
\end{figure*}

\begin{figure*}[t!] \centering \setlength{\tabcolsep}{2pt} \begin{tabular}{cccccc} \textbf{Raw Image} & \textbf{Mask Image} & \textbf{CLAHE 25} & \textbf{CLAHE 10} & \textbf{HE} & \textbf{NE} \\[2mm]     
    \includegraphics[width=0.15\textwidth,height=2.6cm,keepaspectratio]{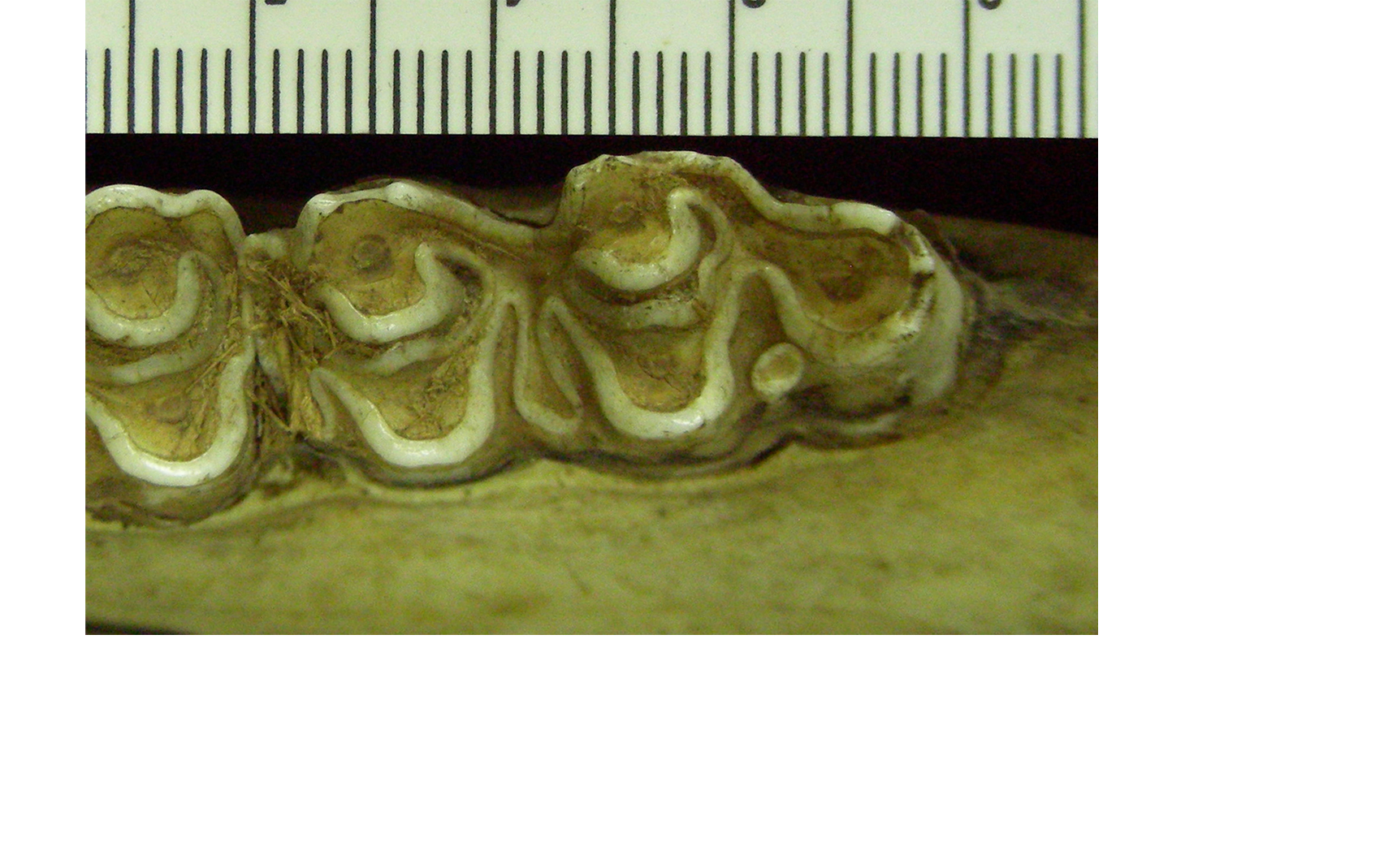} &\includegraphics[width=0.15\textwidth,height=2.6cm,keepaspectratio]{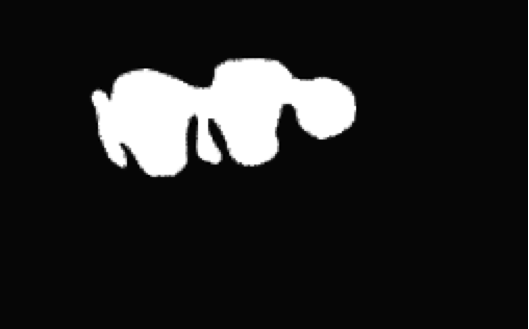} & \includegraphics[width=0.15\textwidth,height=2.6cm,keepaspectratio]{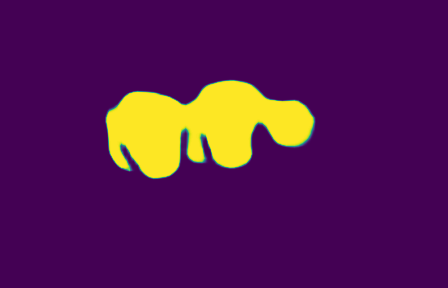} & \includegraphics[width=0.15\textwidth,height=2.6cm,keepaspectratio]{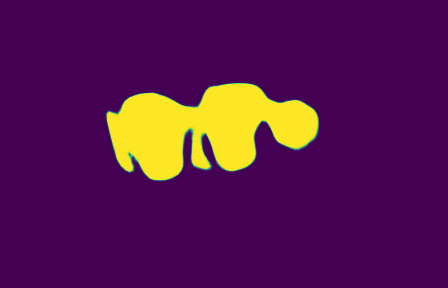} & \includegraphics[width=0.15\textwidth,height=2.6cm,keepaspectratio]{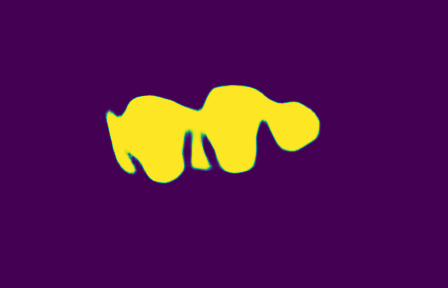} & \includegraphics[width=0.15\textwidth,height=2.6cm,keepaspectratio]{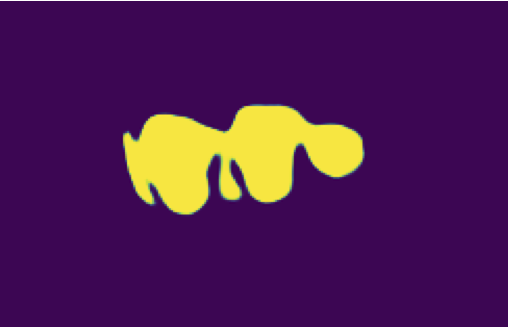} \\[1mm] \includegraphics[width=0.15\textwidth,height=2.6cm,keepaspectratio]{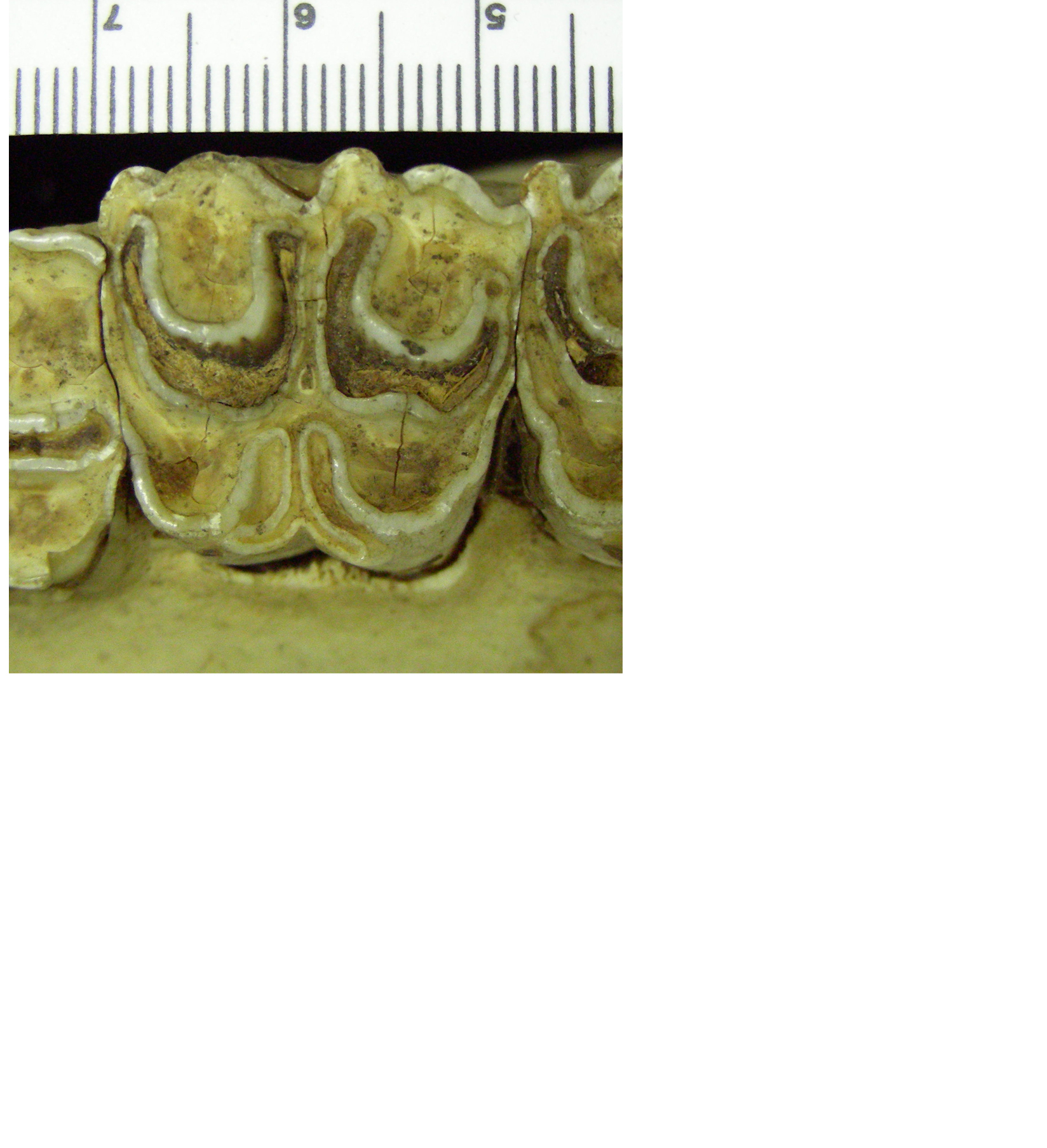} &\includegraphics[width=0.15\textwidth,height=2.6cm,keepaspectratio]{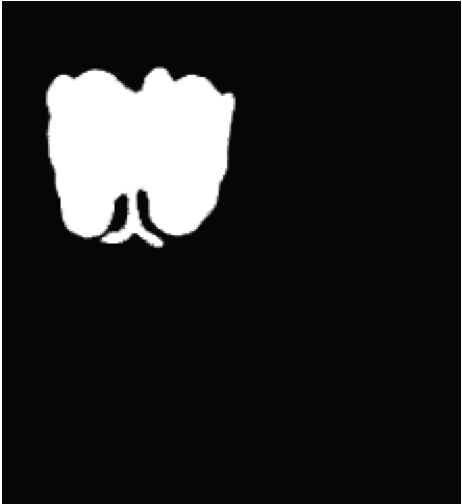} & \includegraphics[width=0.15\textwidth,height=2.6cm,keepaspectratio]{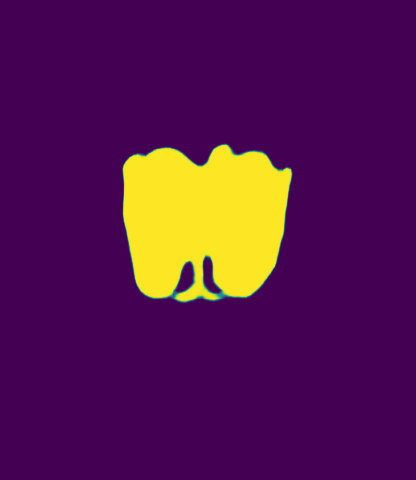} & \includegraphics[width=0.15\textwidth,height=2.6cm,keepaspectratio]{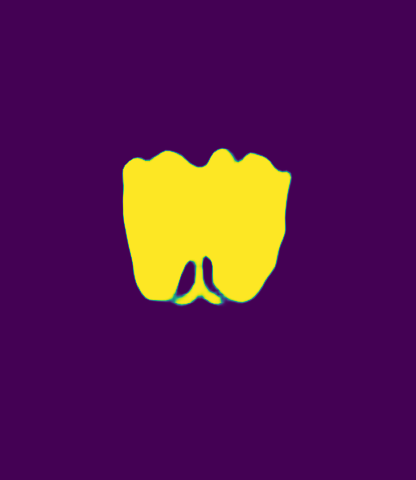} & \includegraphics[width=0.15\textwidth,height=2.6cm,keepaspectratio]{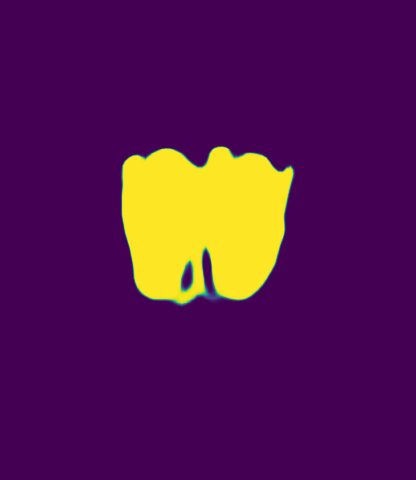} & \includegraphics[width=0.15\textwidth,height=2.6cm,keepaspectratio]{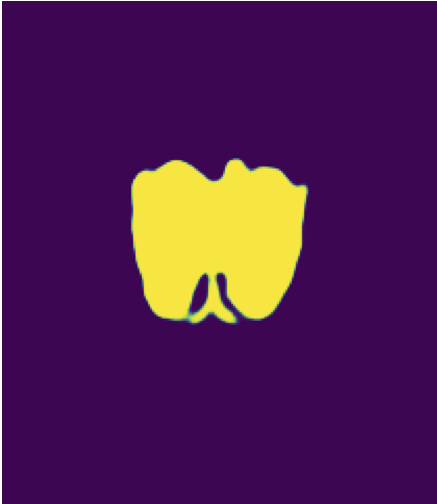} \\[1mm] \includegraphics[width=0.15\textwidth,height=2.6cm,keepaspectratio]{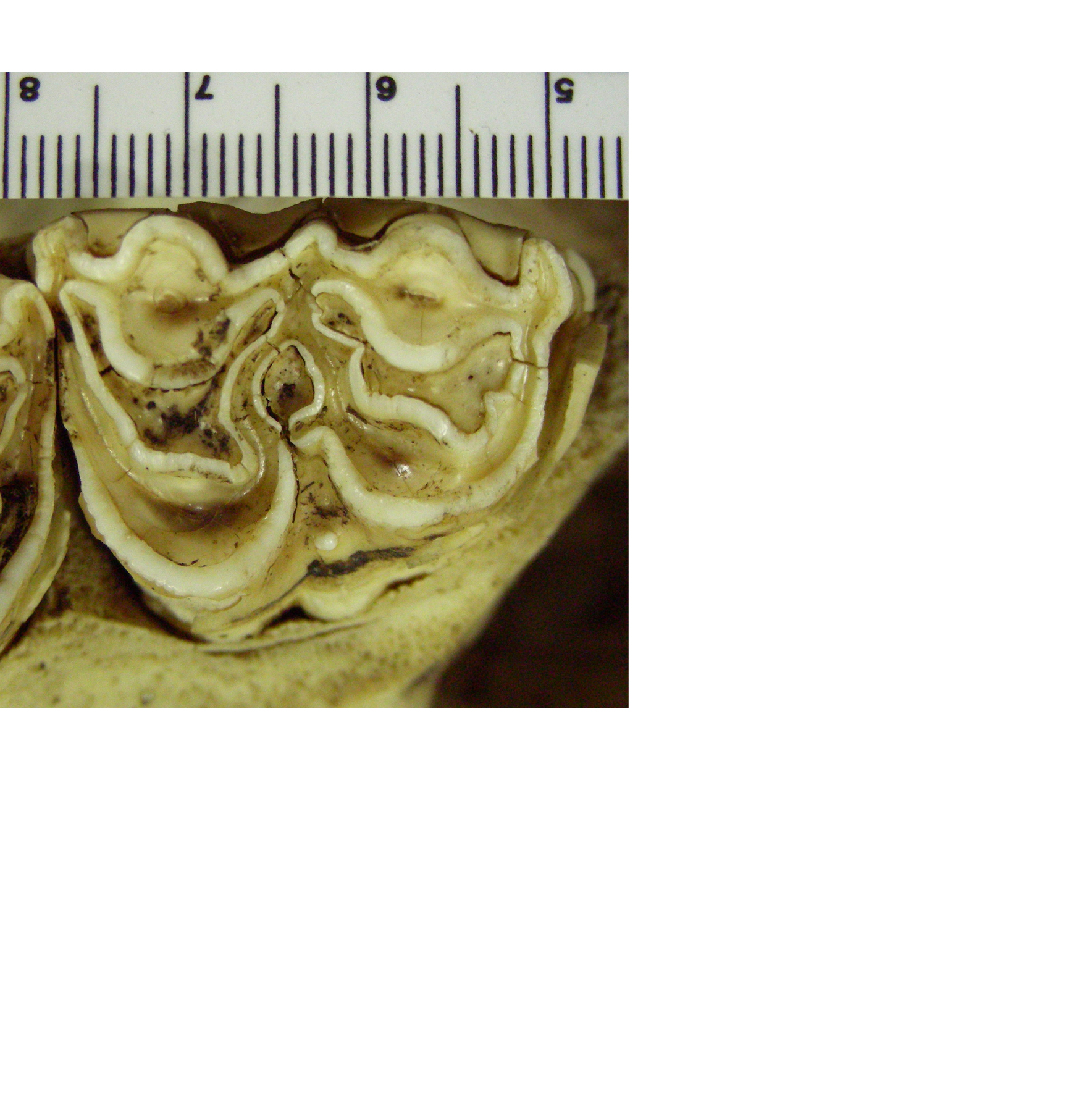} &\includegraphics[width=0.15\textwidth,height=2.6cm,keepaspectratio]{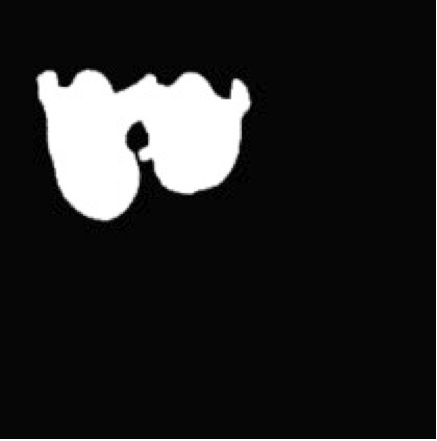} & \includegraphics[width=0.15\textwidth,height=2.6cm,keepaspectratio]{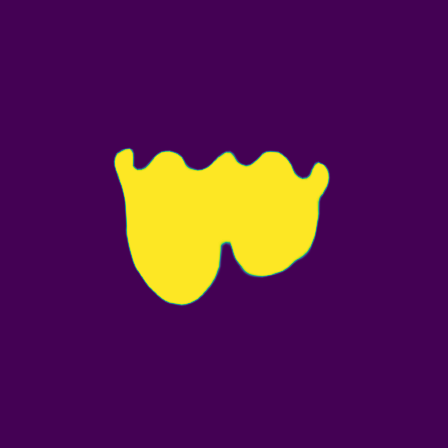} & \includegraphics[width=0.15\textwidth,height=2.6cm,keepaspectratio]{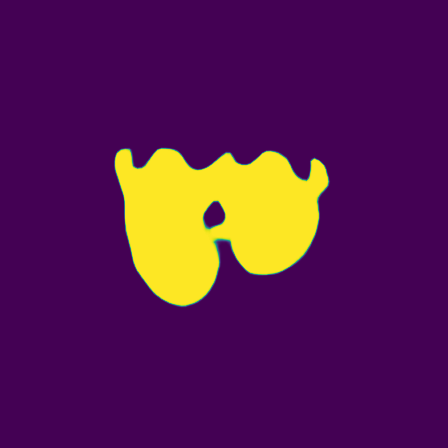} & \includegraphics[width=0.15\textwidth,height=2.6cm,keepaspectratio]{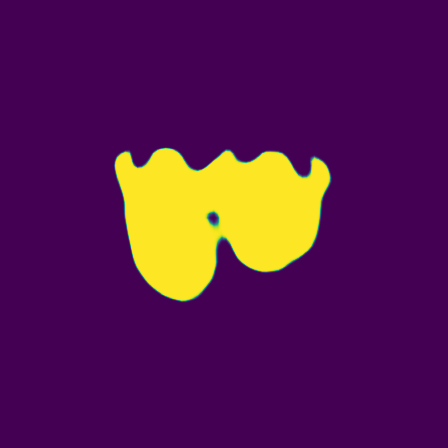} & \includegraphics[width=0.15\textwidth,height=2.6cm,keepaspectratio]{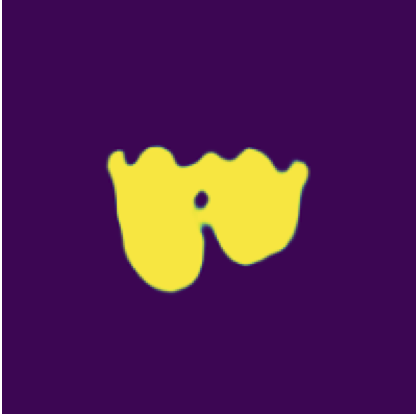} \\[1mm] \includegraphics[width=0.15\textwidth,height=2.6cm,keepaspectratio]{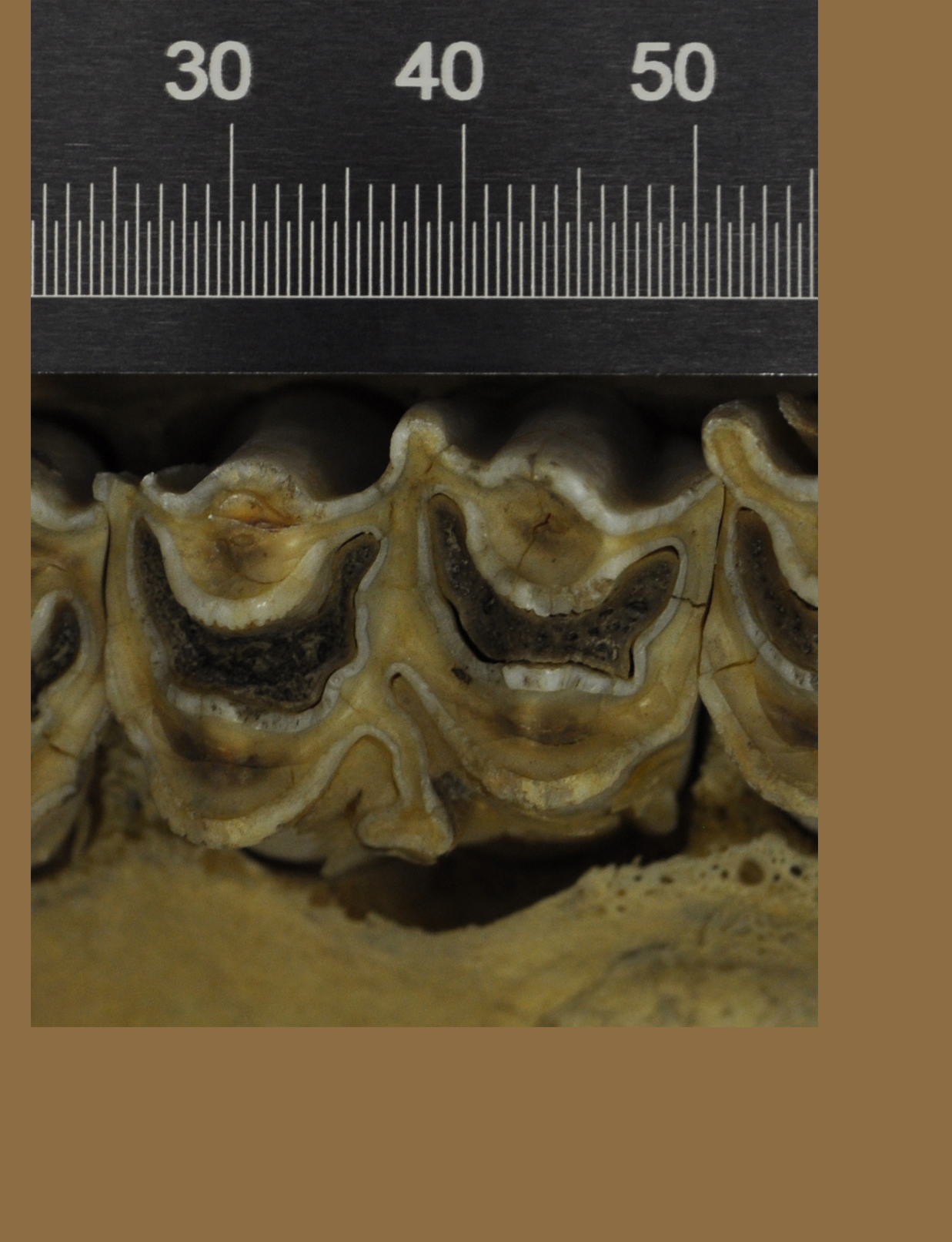} &\includegraphics[width=0.15\textwidth,height=2.6cm,keepaspectratio]{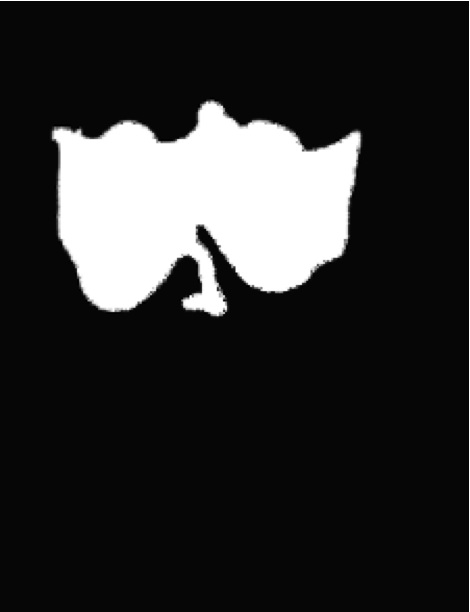} & \includegraphics[width=0.15\textwidth,height=2.6cm,keepaspectratio]{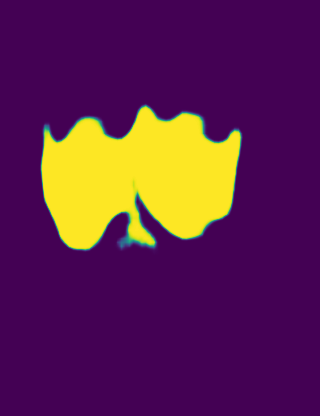} & \includegraphics[width=0.15\textwidth,height=2.6cm,keepaspectratio]{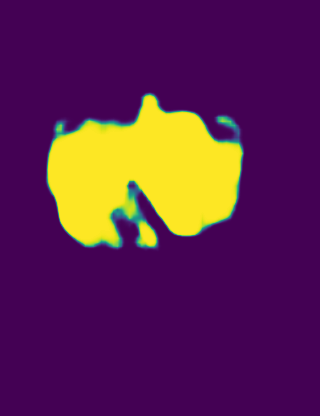} & \includegraphics[width=0.15\textwidth,height=2.6cm,keepaspectratio]{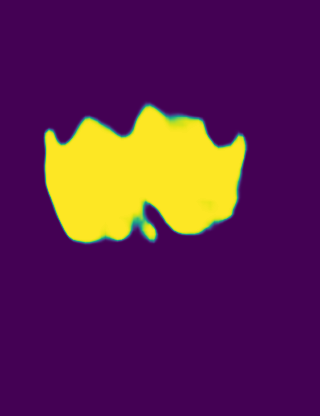} & \includegraphics[width=0.15\textwidth,height=2.6cm,keepaspectratio]{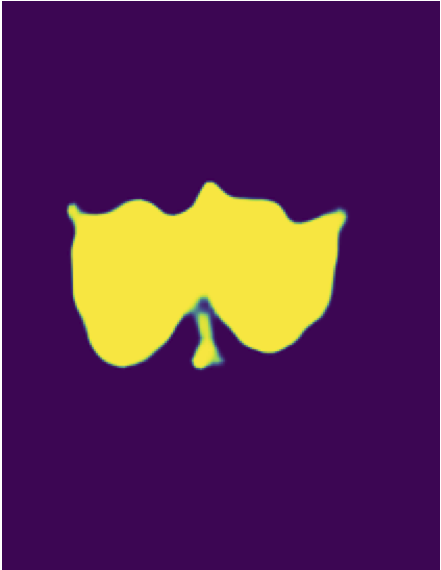} \\ \end{tabular} 
    \caption{Example Raw Images (including whitespace) and ground-truth Mask Images as well as inferred heatmaps by SegFormer MiT-B3 using different histogram equalization techniques. Yellow represents regions with high ($\sim1$) likelyhood of being part of the tooth; purple represents low likelihood. Best viewed in color.} 
    \label{fig:masks_segformer} 
\end{figure*}

Following contrast processing, we pass the image through the encoder-specific ImageNet~\cite{deng2009imagenet} preprocessing function by the segmentation model library~\cite{iakubovskii2019segmentation} and convert it to a float-point tensor with channel-first ordering.  Samples within each minibatch are zero-padded to the largest spatial height and width dimensions within the minibatch. Further, images and masks were optionally downsampled by an integer factor $d$. This reduces GPU training and inference memory consumption and also helps to abate the affect of misalignment. Specifically, we downsample Raw Images using bilinear interpolation and downsample Mask Images by nearest-neighbor interpolation to preserve binary labels. Finally, if necessary, we pad the height and width to be a multiple of 32 to match spatial requirements of encoder-decoder architectures. 

Finally, we train our segmentation models using a combination of the binary cross-entropy and weighted Dice~\cite{kato2024dice} losses as follows

\begin{equation}
    \centering
    \mathcal{L} = \mathcal{L}_{\mathrm{BCE}} + \lambda_{\mathrm{Dice}}\mathcal{L}_{\mathrm{Dice}},
    \label{eq:loss}
\end{equation}
where $\lambda_{\mathrm{Dice}} \in [0, 1]$ is a configurable scalar.

\section{Results and Discussion}
\label{sec:results}

We now present our experimental results and findings. Specifically, we first briefly describe our experimental setup, before delving into the effectiveness of different histogram equalization preprocessing techniques, for both Transformers and CNNs. Next, we provide additional per-tribe (e.g., like per-class in an ML-sense) segmentation results based on different species of bovine, before finally showing some results on brand new, out of distribution data samples captured in the summer of 2026. Additional ablations exist in the supplementary.

\begin{figure*}[t!] 
    \centering 
    \begin{subfigure}[t]{0.3\linewidth} \centering \includegraphics[width=\linewidth]{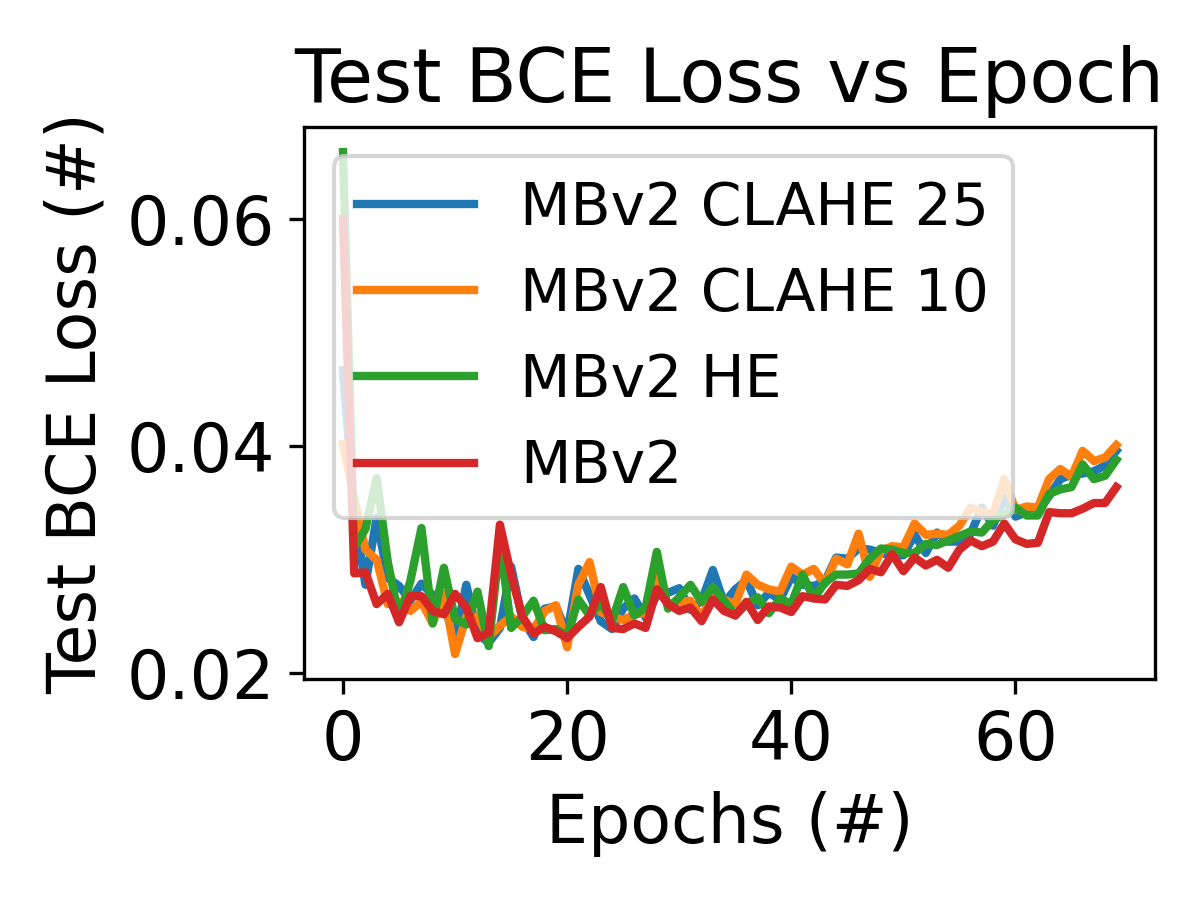} %
    \end{subfigure} \hfill \begin{subfigure}[t]{0.3\linewidth} \centering \includegraphics[width=\linewidth]{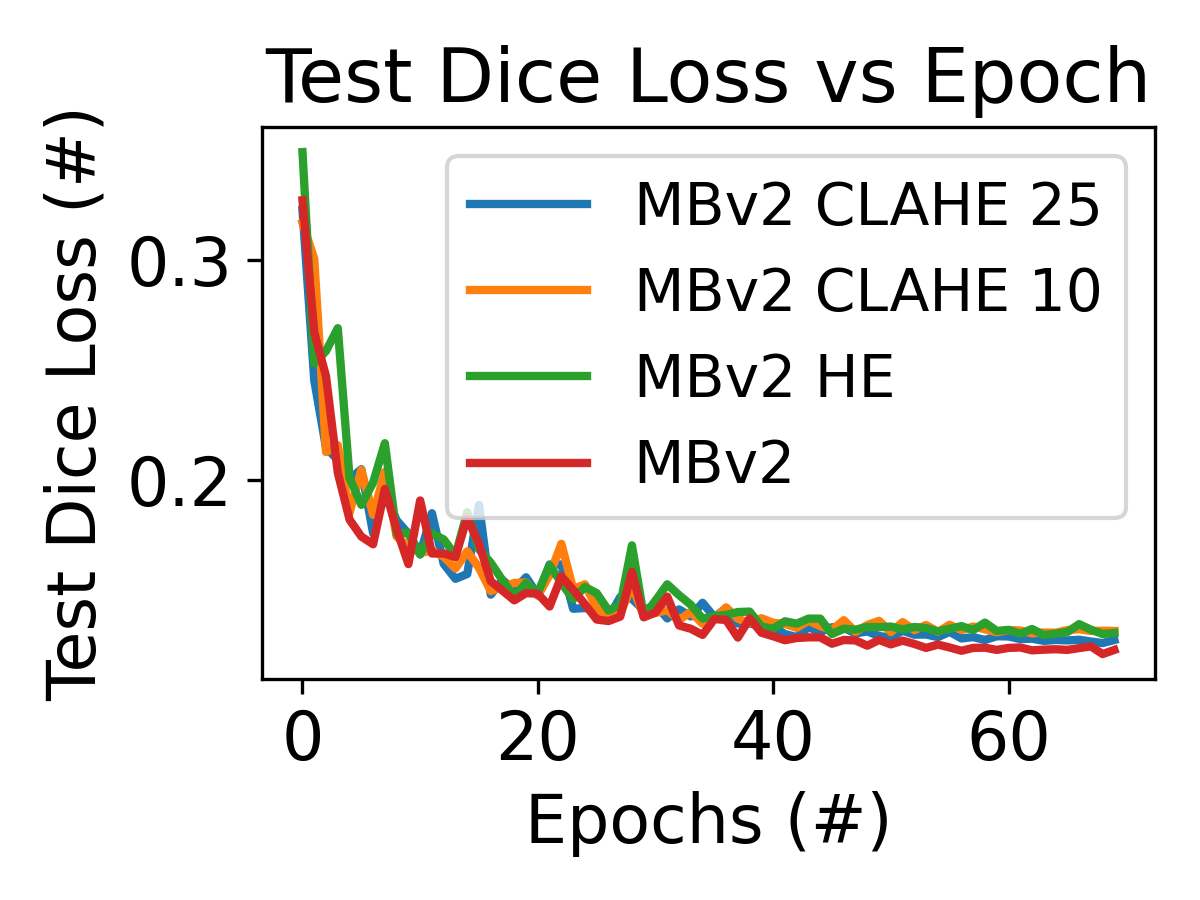} %
    \end{subfigure} \hfill \begin{subfigure}[t]{0.3\linewidth} \centering \includegraphics[width=\linewidth]{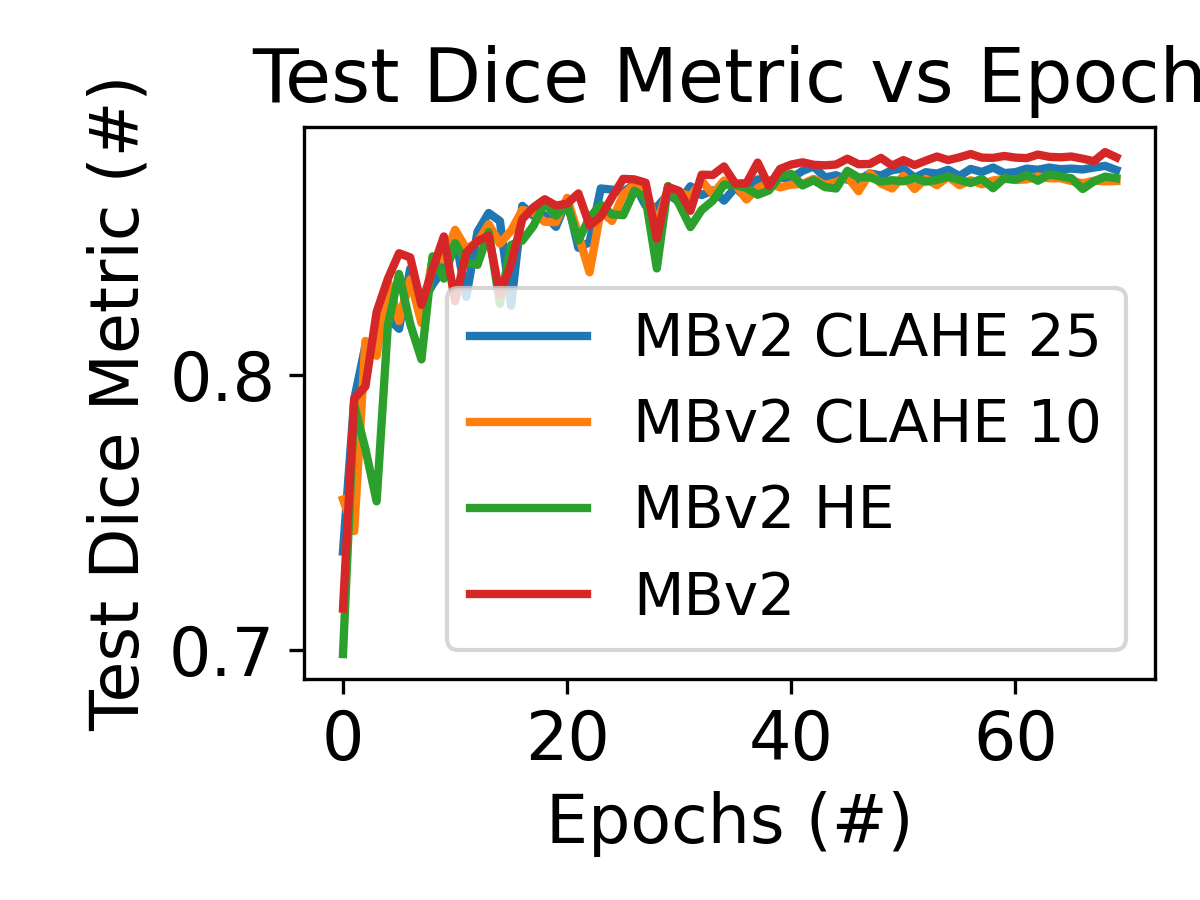} %
    \end{subfigure} \caption{BCE loss, Dice loss and Dice metric for U-Net++ MBv2. For losses, lower is better, while higher is better for the Dice metric.} \label{fig:mbv2_main}
\end{figure*}

\begin{figure*}[t!] \centering \setlength{\tabcolsep}{2pt} \begin{tabular}{cccccc} \textbf{Raw Image} & \textbf{Mask Image} & \textbf{CLAHE 25} & \textbf{CLAHE 10} & \textbf{HE} & \textbf{NE} \\[2mm] 
    
    \includegraphics[width=0.15\textwidth,height=2.6cm,keepaspectratio]{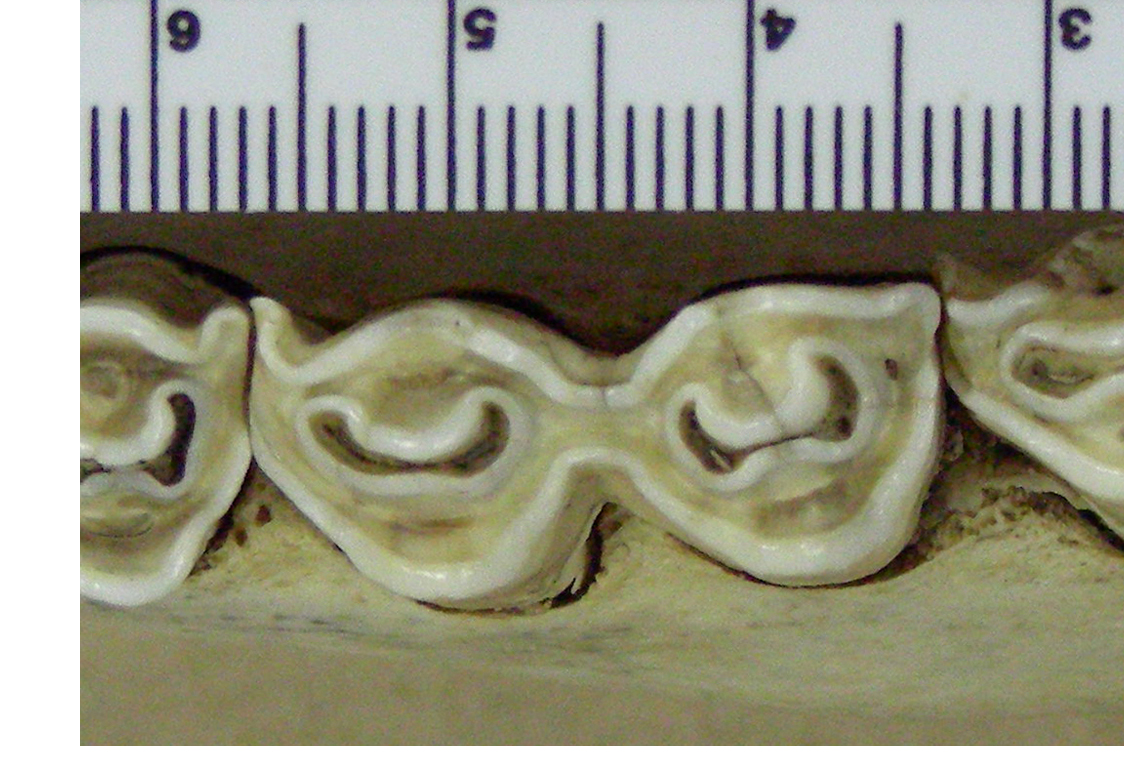} & \includegraphics[width=0.15\textwidth,height=2.6cm,keepaspectratio]{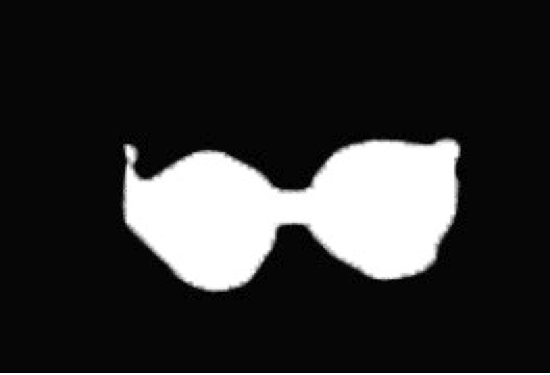} & \includegraphics[width=0.15\textwidth,height=2.6cm,keepaspectratio]{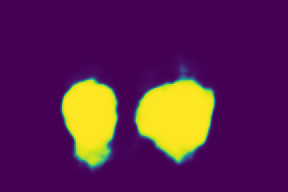} & \includegraphics[width=0.15\textwidth,height=2.6cm,keepaspectratio]{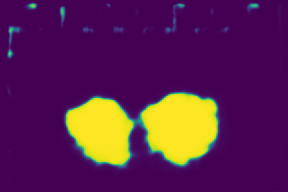} & \includegraphics[width=0.15\textwidth,height=2.6cm,keepaspectratio]{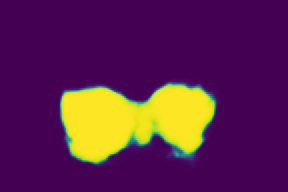} & \includegraphics[width=0.15\textwidth,height=2.6cm,keepaspectratio]{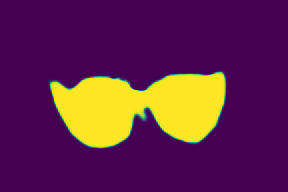} \\[1mm]
    
    \includegraphics[width=0.15\textwidth,height=2.6cm,keepaspectratio]{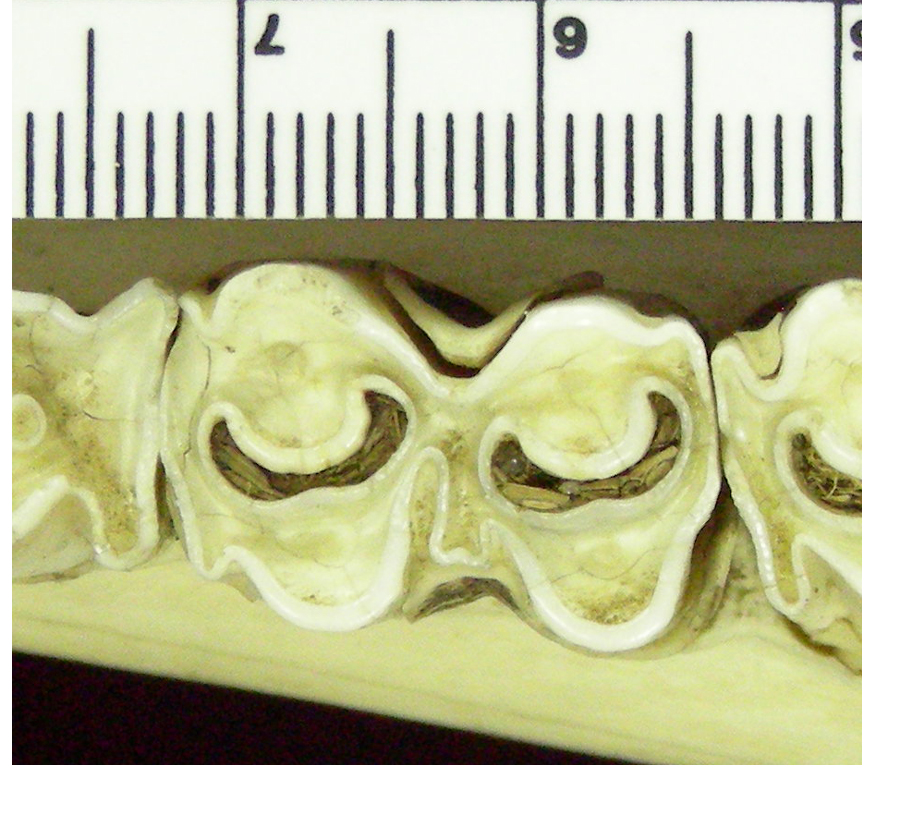} &\includegraphics[width=0.15\textwidth,height=2.6cm,keepaspectratio]{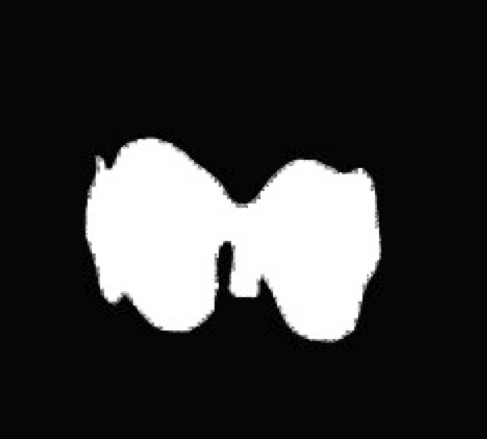} & \includegraphics[width=0.15\textwidth,height=2.6cm,keepaspectratio]{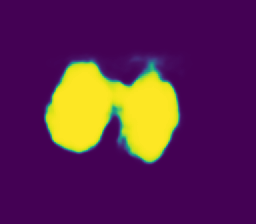} & \includegraphics[width=0.15\textwidth,height=2.6cm,keepaspectratio]{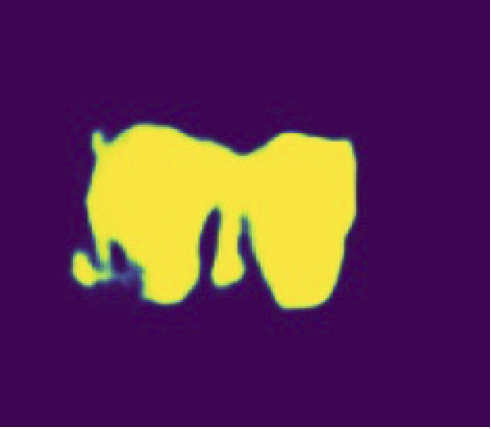} & \includegraphics[width=0.15\textwidth,height=2.6cm,keepaspectratio]{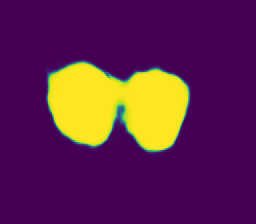} & \includegraphics[width=0.15\textwidth,height=2.6cm,keepaspectratio]{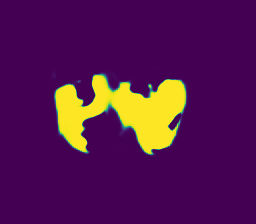} \\[1mm] 
    
    \includegraphics[width=0.15\textwidth,height=2.6cm,keepaspectratio]{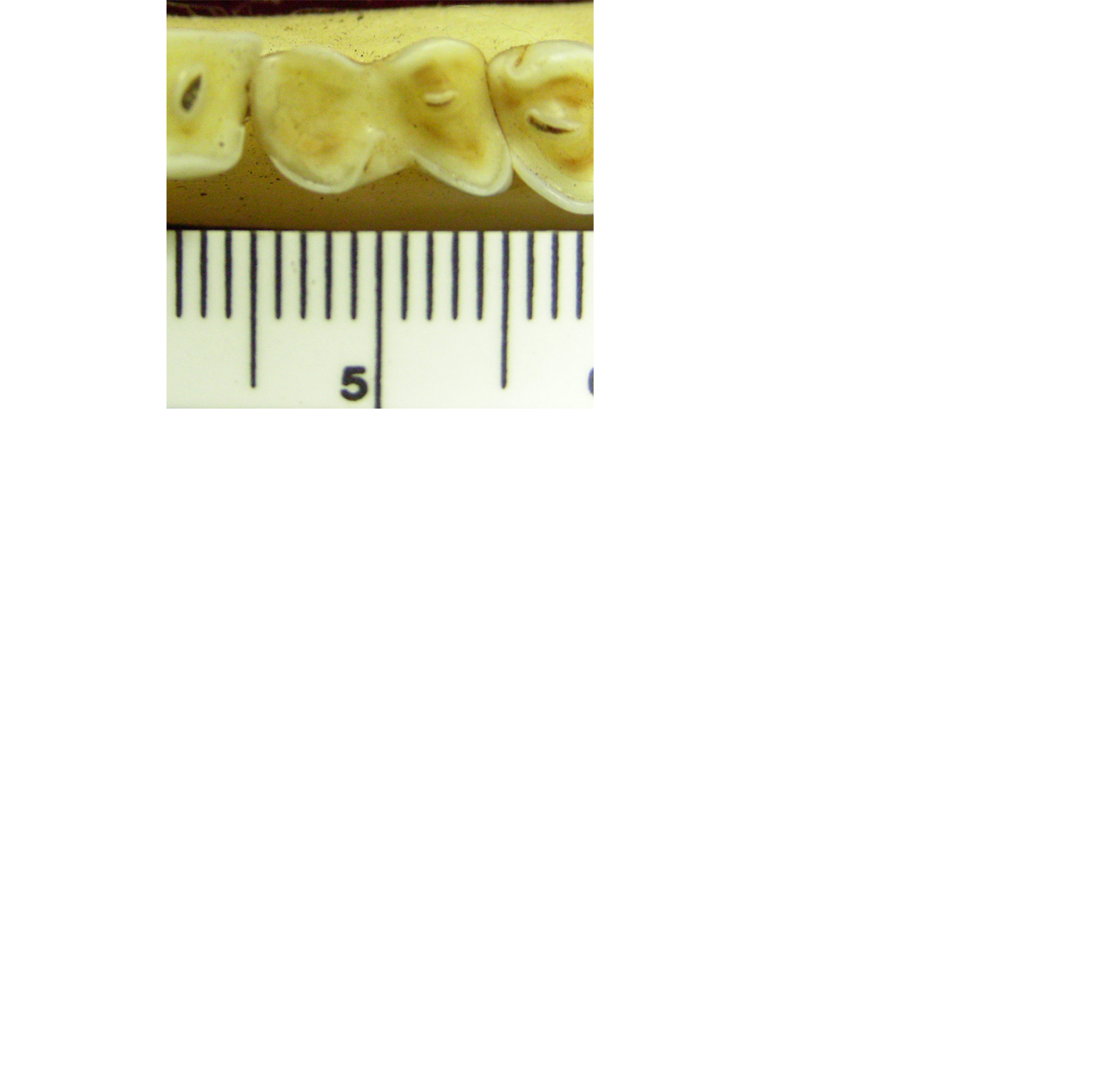} &\includegraphics[width=0.15\textwidth,height=2.6cm,keepaspectratio]{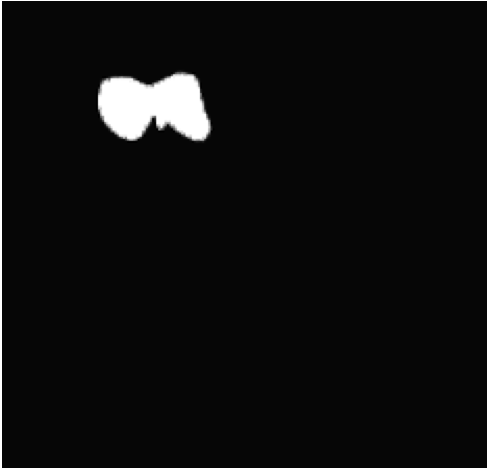} & \includegraphics[width=0.15\textwidth,height=2.6cm,keepaspectratio]{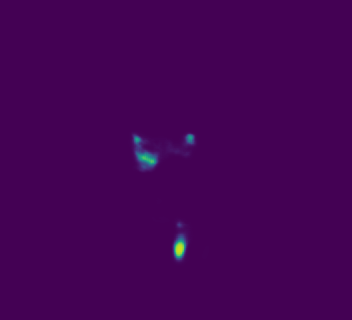} & \includegraphics[width=0.15\textwidth,height=2.6cm,keepaspectratio]{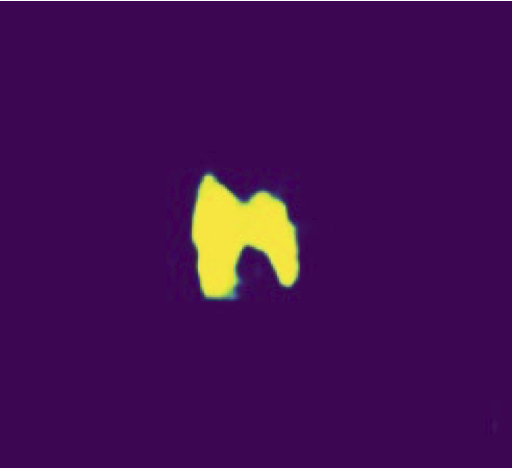} & \includegraphics[width=0.15\textwidth,height=2.6cm,keepaspectratio]{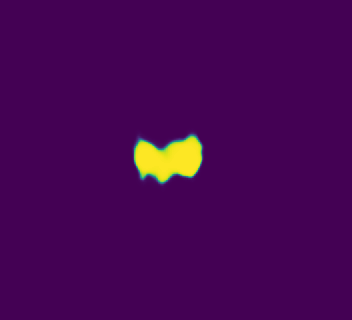} & \includegraphics[width=0.15\textwidth,height=2.6cm,keepaspectratio]{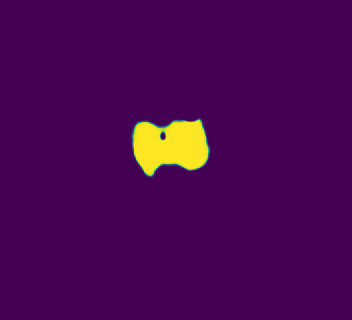} \\[1mm]
    
    \includegraphics[width=0.15\textwidth,height=2.6cm,keepaspectratio]{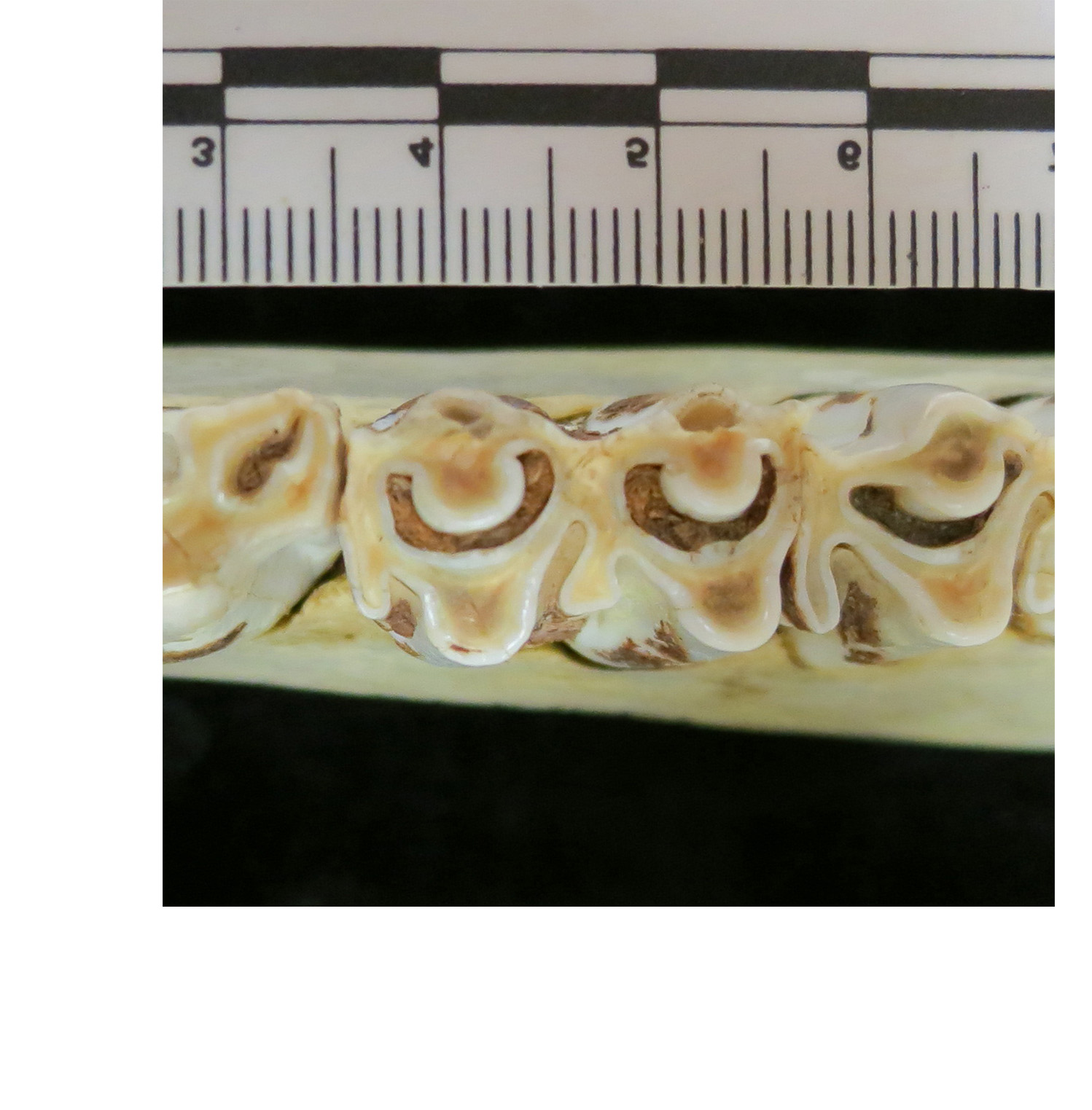} &\includegraphics[width=0.15\textwidth,height=2.6cm,keepaspectratio]{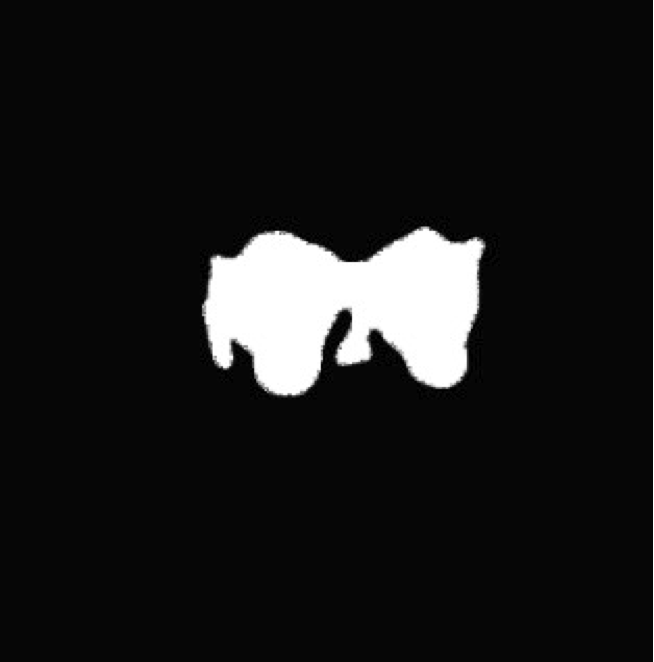} & \includegraphics[width=0.15\textwidth,height=2.6cm,keepaspectratio]{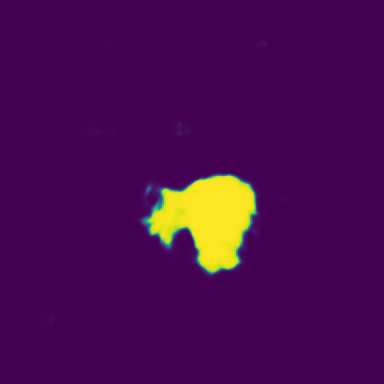} & \includegraphics[width=0.15\textwidth,height=2.6cm,keepaspectratio]{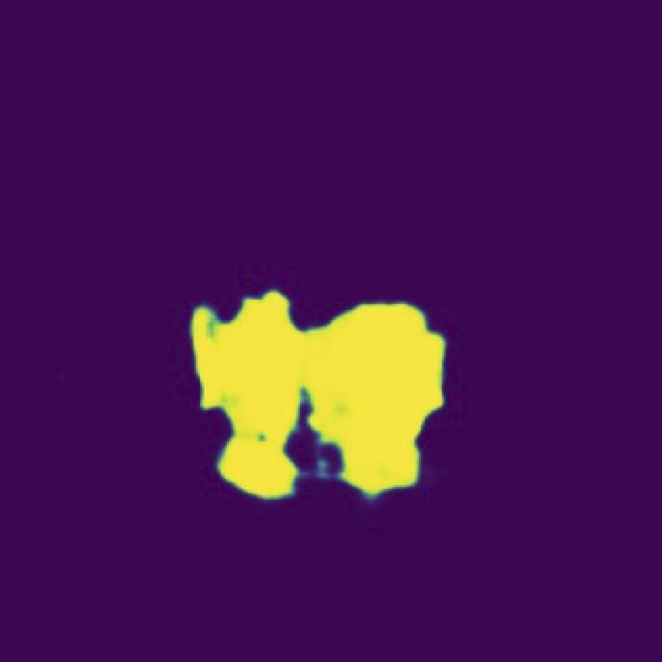} & \includegraphics[width=0.15\textwidth,height=2.6cm,keepaspectratio]{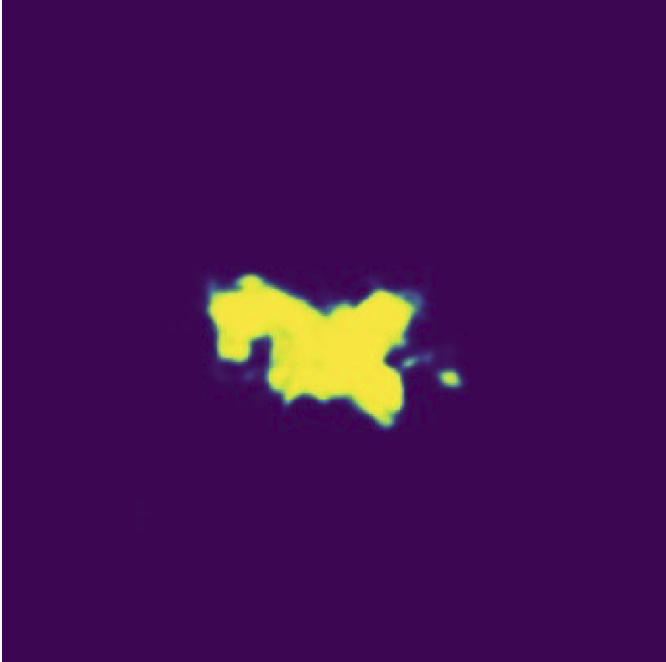} & \includegraphics[width=0.15\textwidth,height=2.6cm,keepaspectratio]{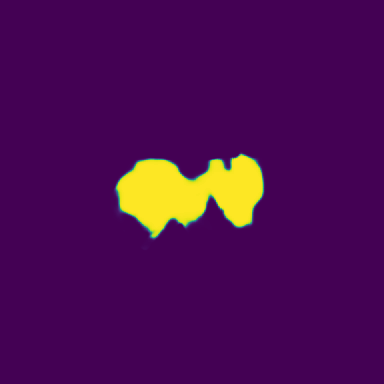} \\[1mm] \end{tabular} \caption{Example Raw Images (including whitespace) and ground-truth Mask Images as well as inferred heatmaps by MBv2 using different histogram equalization techniques. Yellow represents regions with high ($\sim1$) likelihood of being part of the tooth; purple represents low likelihood. Best viewed in color.} 
    \label{fig:masks_mbv2} 
    \vspace{-4mm}
\end{figure*}

\subsection{Experimental Setup}

We randomly shuffle the B.O.V.I.D. dataset 80\%/20\% into disjoint training and testing sets. Semantic segmentation models come from the segmentation-models-pytorch~\cite{iakubovskii2019segmentation} library. For transformers we utilize SegFormer MiT-B3~\cite{xie2021segformer} while for CNNs we utilize the U-Net++~\cite{zhou2018unet++} model with either MobileNetV2 (MBv2)~\cite{sandler2018mobilenetv2}, ResNet-34~\cite{he2016deep} or EfficientNet-B0~\cite{tan2019efficientnet} serving as the image encoder backbone. We train each model for 70 epochs and set $\lambda_{Dice}=0.3$. Due to space constraints, the most salient hyperparameter choices are part of our methodology in Section~\ref{sec:aug} while we provide additional details and ablations, i.e., for $\lambda_{Dice}$ in the supplementary. %

\subsection{SegFormer Quantitative/Qualitative Results}

We train SegFormer MiT-B3 for 70 epochs, varying the type of preprocessing from CLAHE with $c\in [10, 25]$ as the clip limit, HE and NE. Quantitatively, we measure the binary cross entropy (BCE) loss, Dice loss, and Dice metric on the test set after every epoch. Figure~\ref{fig:segformer_main} plots the results. All four configurations show similar trends: Sharp decrease in the BCE loss followed by a gradual rise, paired with a continuous decrease or increase of the Dice loss or metric, respectively. In terms of Dice, all four configurations are relatively equal and it is hard to discern which is superior, but this is not the case for test BCE loss as CLAHE with $c=25$ consistently achieves a lower loss, even during the upswing after epoch 15 or so. Overall, these results demonstrate the feasibility of fine-tuning a SegFormer model on B.O.V.I.D. despite the segmentation imperfections in the data. However, what is more important is the actual predicted masks. 

During training we track the epoch that corresponds to the best Dice loss and save the weights at this epoch as the checkpoint for inference. After training, we generate \textit{heatmaps} where values range between 0 and 1, rather than being 0 or 1, to visualize model decision making. Figure~\ref{fig:masks_segformer} provides some sample sets of ground truth masks alongside the respective heatmaps generated by each preprocessing technique for SegFormer. For each type of preprocessing, the model predicts a mask in roughly the center of the region even though the ground truth may have the mask location shifted. What is more interesting is the shape of the mask and how well it conforms to the ground truth. Here, we can see that all methods draw a similar region for the first picture, while for the second, HE and NE struggle a bit more with capturing the small masked region shaped like an upside-down `Y' in terms of heatmap values. For the third image, CLAHE 25 draws over the middle of the mask incorrectly while CLAHE 10 best captures the overall shape. For the final image, most methods except CLAHE 25 and NE tend to destabilize in some way, either missing critical shape details. Overall though, this model is capable of producing competent heatmaps, which can the be thresholded into accurate masks, for the B.O.V.I.D. dataset.

\subsection{MBv2 Quantitative/Qualitative Results}

Next, we study the capacity of older, but more computationally lightweight convolutional neural network architectures, to segment the B.O.V.I.D. dataset. Due to space constraints, we primarily consider MBv2 in the main body of this manuscript but provide ablations to demonstrate its performance over other CNNs in the supplementary. Figure~\ref{fig:mbv2_main} mirrors Fig.~\ref{fig:segformer_main} for MBv2 by graphing the test metrics across training epochs. We make a few observations: First, early on, learning is more stochastic on MBv2 than for SegFormer MiT-B3, with larger spikes in the test metrics by epoch, although this eventually smooths out by epoch 50 or so. Second, like SegFormer, the BCE loss drops sharply then gradually increases, though not as quickly as SegFormer, while the Dice loss and Dice metric fall and rise, respectively. Third, the lower-bound of the Dice loss is not as low as it is for SegFormer, nor is the upper bound of the Dice metric. All of these metric observations suggest a model that may also be competent at this task, but less capable than SegFormer.

Figure~\ref{fig:masks_mbv2} mirrors Fig.~\ref{fig:masks_segformer} and largely corroborates this hypothesis. Here, we can see that MBv2 relies on a good preprocessing histogram equalization strategy to generate accurate masks. Specifically, CLAHE $c=10$ stands out as best representing the 2nd and 4th images which have the most complex shapes to accurately mask. In contrast, NE best represents the 1st image and does well on the 3rd, which are simpler, but falls apart completely for the more complex 2nd and 4th images, as evidenced by the large gaps in the mask constructed by NE for the 2nd image.

\subsection{Stratified Per-Tribe Results}
\label{sec:strat}

Next, we provide results and comparison on a per-tribe level, i.e., rather than lumping all (Raw Image, Mask Image) pairs together for an ML dataset, we consider the different tribes, e.g., Alcelaphini, Neotragini, etc.~\cite{brophy2022bovid}. When performing this kind of experiment, we further change the dataloader to stratify the data, ensuring that when we perform an 80\%/20\% data split, both partitions contain roughly the same proportion of data per tribe as the overall dataset. Further, in addition to the Dice metric, we also evaluate mean Intersection over Union (mIoU). 

Table~\ref{tab:segformer_b3_per_class} shows the results on SegFormer MiT-B3. Quantitatively, out of 7 different tribes, CLAHE 25 achieves the best Dice and mIoU in a majority of cases. Specifically, CLAHE $c=25$ excels at the Alcelaphini, Antilopini and Hippotragini tribes, all of which tend to carry higher Dice and mIoU scores. Moreover, these results tend not to be correlated with the dataset size, but likely other underlying traits of the tribes. However, like before, these quantitative results are best paired with qualitative visual examples. 

\begin{table*}[t!]
    \centering
    \scalebox{0.85}{
    \begin{tabular}{l|cccc|cccc} \toprule
    \textbf{Metric} &  \multicolumn{4}{c|}{\textbf{Dice}} & \multicolumn{4}{c}{\textbf{mIoU}} \\ \midrule
    \textbf{Tribe} & \textbf{CLAHE 25} & \textbf{CLAHE 10} & \textbf{HE} & \textbf{NE} & \textbf{CLAHE 25} & \textbf{CLAHE 10} & \textbf{HE} & \textbf{NE} \\ \midrule 
    Alcelaphini (665) & $\mathbf{0.9271}_{0.008}$  & $0.9230_{0.013}$  & $0.9248_{0.005}$ & $0.9269_{0.011}$ & $\mathbf{0.8639}_{0.008}$ & $0.8572_{0.022}$ & $0.8602_{0.009}$ & $0.8637_{0.019}$ \\
    Antilopini (180) & $\mathbf{0.9022}_{0.009}$   & $0.8960_{0.010}$ & $0.8994_{0.002}$ & $0.9006_{0.004}$ & $\mathbf{0.8220}_{0.009}$ & $0.8117_{0.018}$ & $0.8171_{0.003}$ & $0.8193_{0.006}$ \\
    Bovini (138) & $\mathbf{0.9413}_{0.006}$       & $0.9398_{0.006}$ & $0.9370_{0.008}$ & $0.9400_{0.008}$ & $0.8864_{0.006}$ & $\mathbf{0.8893}_{0.011}$ & $0.8816_{0.013}$ & $0.8866_{0.014}$ \\
    Hippotragini (483) & $\mathbf{0.9207}_{0.008}$ & $0.9162_{0.009}$ & $0.9133_{0.010}$ & $0.9183_{0.005}$ & $\mathbf{0.8531}_{0.008}$ & $0.8454_{0.015}$ & $0.8405_{0.016}$ & $0.8491_{0.008}$ \\
    Neotragini (757) & $0.8697_{0.003}$   & $0.8645_{0.013}$  & $0.8708_{0.001}$ & $\mathbf{0.8637}_{0.010}$ & $0.7694_{0.003}$ & $0.7614_{0.020}$ & $\mathbf{0.7712}_{0.001}$ & $0.7601_{0.016}$\\
    Reduncini (489)& $0.9275_{0.002}$    & $0.9269_{0.004}$  & $\mathbf{0.9290}_{0.003}$ & $0.9315_{0.003}$ & $0.8648_{0.002}$ & $0.8638_{0.007}$ & $\mathbf{0.8718}_{0.006}$ & $0.8674_{0.005}$ \\
    Tragelaphini (489) & $0.9427_{0.003}$ & $0.9396_{0.004}$ & $0.9390_{0.002}$ & $\mathbf{0.9440}_{0.004}$ & $\mathbf{0.8940}_{0.003}$ & $0.8861_{0.008}$ & $0.8850_{0.003}$ & $0.8916_{0.006}$\\
    \bottomrule
    \end{tabular}
    }
    \caption{Stratified per-tribe segmentation Dice metric and mean Intersection over Union (mIoU) scores for SegFormer MiT-B3. We compare CLAHE at $c\in{10, 25}$, HE and NE. Tribes annotated with number of images. Best results in \textbf{bold}. Results across 3 seeds.}
    \label{tab:segformer_b3_per_class}
\end{table*}

\begin{figure*}[t!] \centering \setlength{\tabcolsep}{2pt} \begin{tabular}{ccccc} \textbf{Tribe} & \textbf{Raw Image} & \textbf{Mask Image} & \textbf{SegFormer MiT-B3} & \textbf{U-Net++ MBv2} \\[2mm] 
    
    \textbf{Alcelaphini} & \includegraphics[width=0.24\textwidth,height=2.6cm,keepaspectratio]{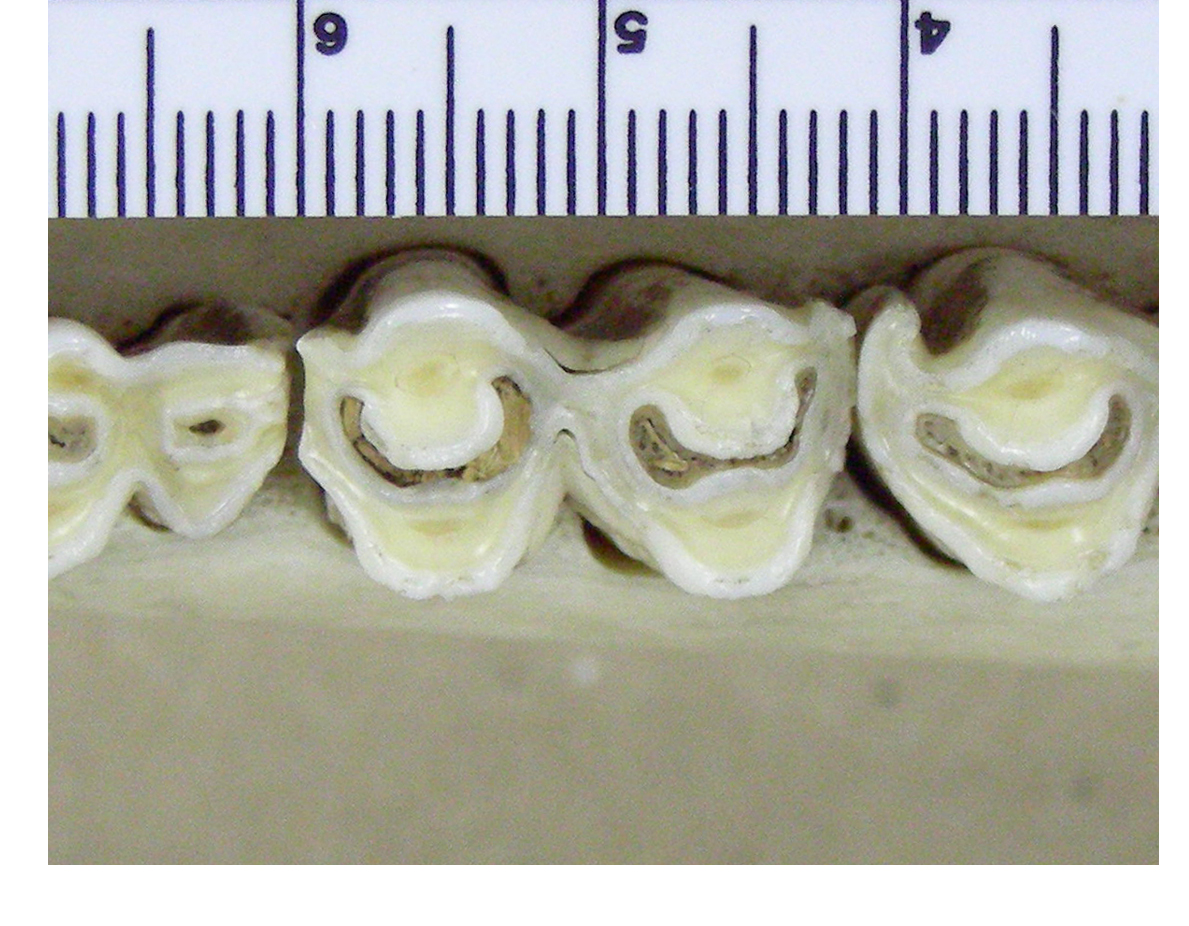} & \includegraphics[width=0.24\textwidth,height=2.6cm,keepaspectratio]{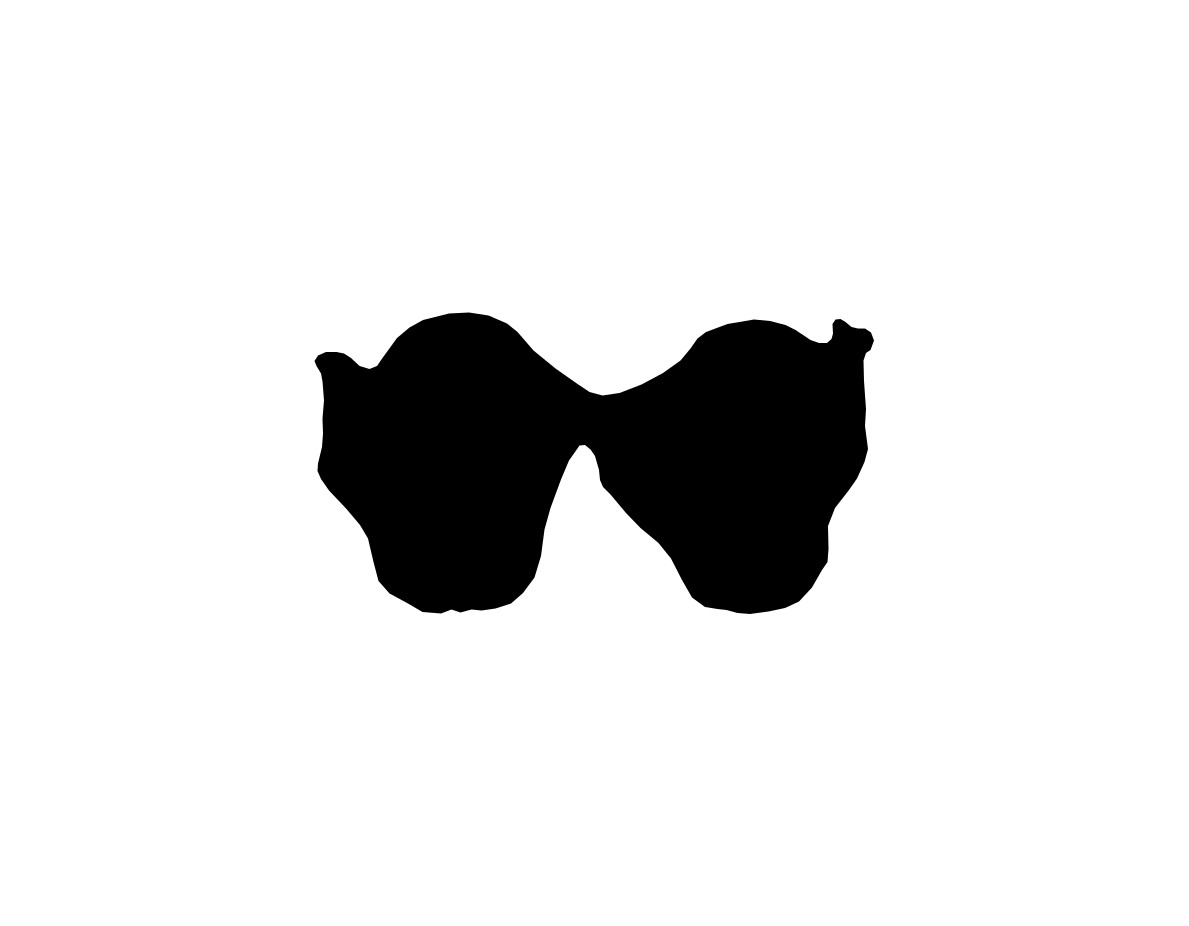} & \includegraphics[width=0.24\textwidth,height=2.6cm,keepaspectratio]{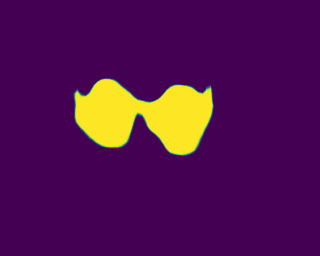} & \includegraphics[width=0.24\textwidth,height=2.6cm,keepaspectratio]{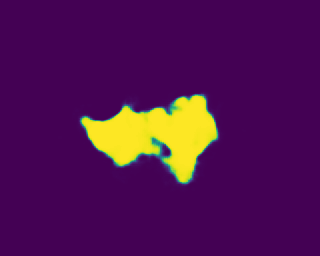} \\[1mm]

    \textbf{Bovini} & \includegraphics[width=0.24\textwidth,height=2.6cm,keepaspectratio]{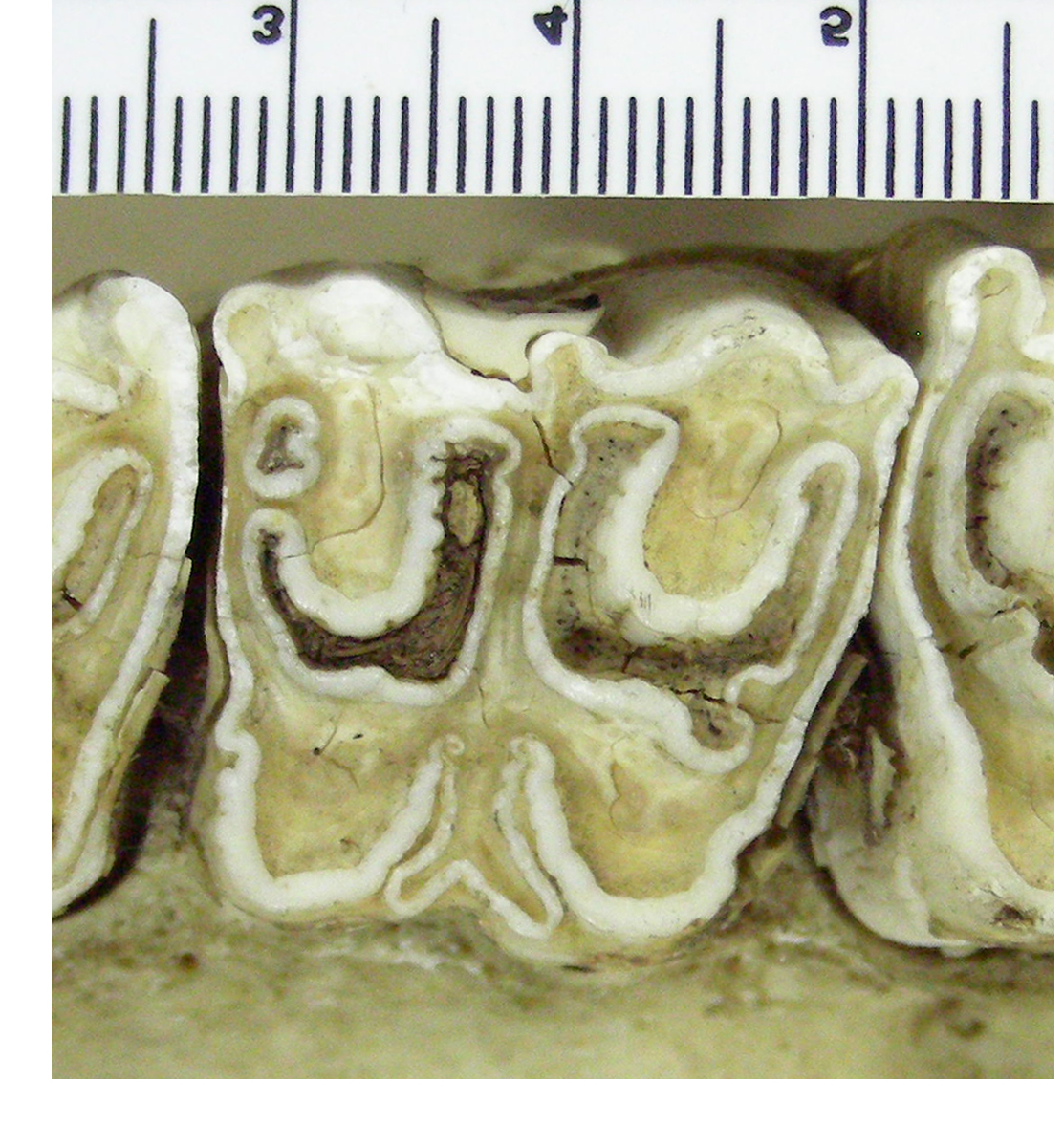} & \includegraphics[width=0.24\textwidth,height=2.6cm,keepaspectratio]{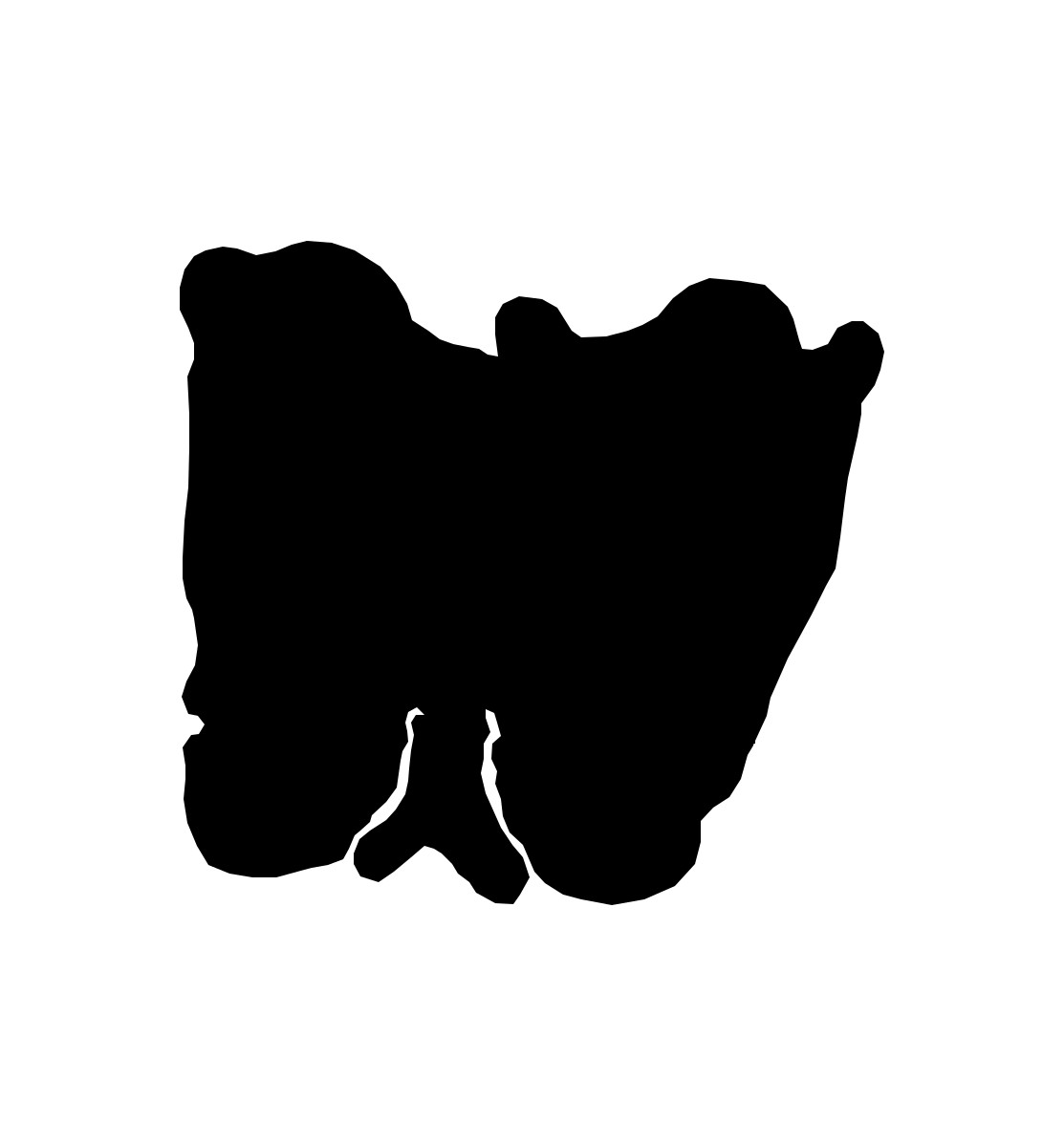} & \includegraphics[width=0.24\textwidth,height=2.6cm,keepaspectratio]{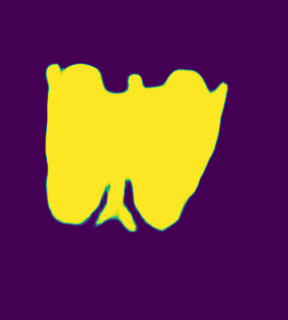} & \includegraphics[width=0.24\textwidth,height=2.6cm,keepaspectratio]{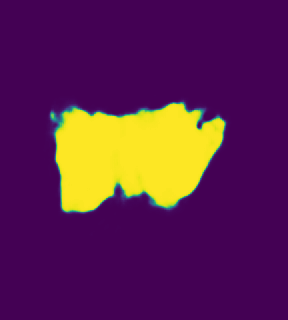} \\[1mm]

    \textbf{Neotragini} & \includegraphics[width=0.24\textwidth,height=2.6cm,keepaspectratio]{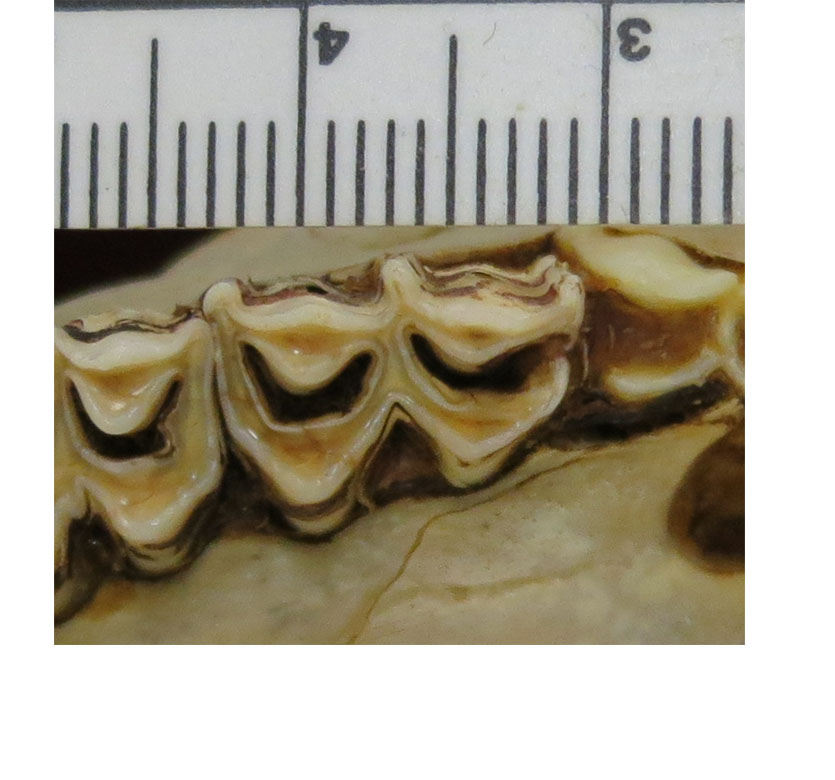} & \includegraphics[width=0.24\textwidth,height=2.6cm,keepaspectratio]{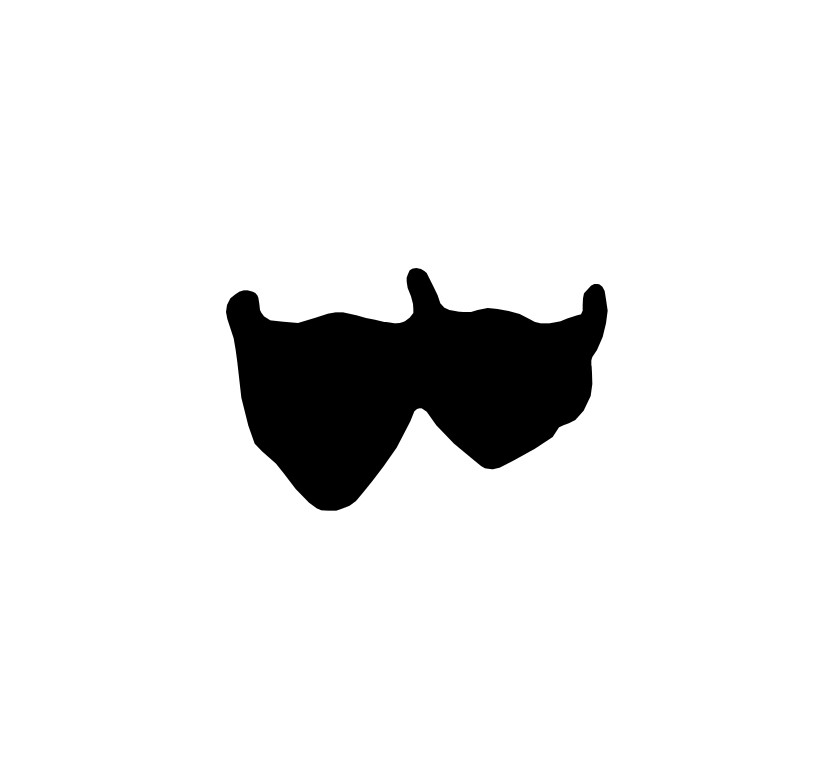} & \includegraphics[width=0.24\textwidth,height=2.6cm,keepaspectratio]{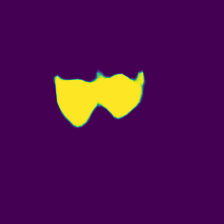} & \includegraphics[width=0.24\textwidth,height=2.6cm,keepaspectratio]{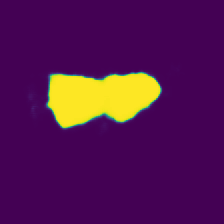} \\[1mm]

    \textbf{Tragelaphini} & \includegraphics[width=0.24\textwidth,height=2.6cm,keepaspectratio]{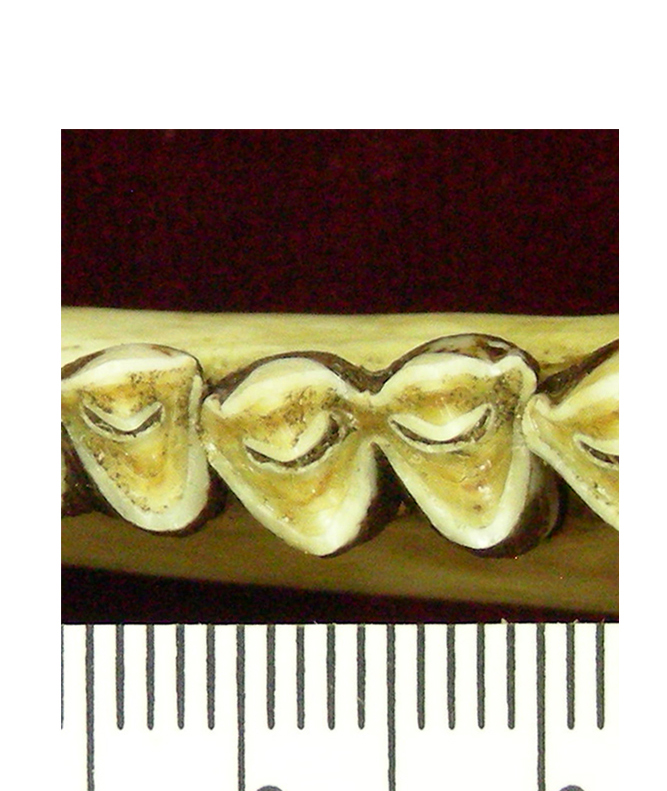} & \includegraphics[width=0.24\textwidth,height=2.6cm,keepaspectratio]{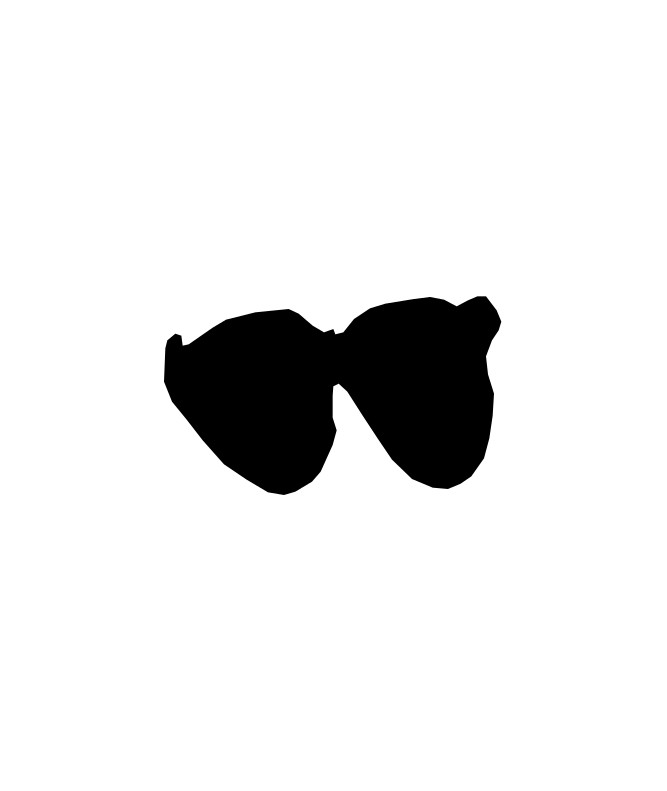} & \includegraphics[width=0.24\textwidth,height=2.6cm,keepaspectratio]{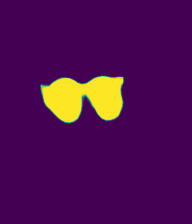} & \includegraphics[width=0.24\textwidth,height=2.6cm,keepaspectratio]{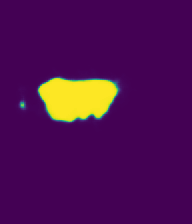} \\[1mm]
    
    \end{tabular} %
    \caption{Example (Raw Image, Mask Image) samples for different B.O.V.I.D. tribes as well as inference results for SegFormer MiT-B3 CLAHE $c=25$ and U-Net++ MBv2 $c=10$. Best viewed in color.} 
    \label{fig:masks_perclass} 
    \vspace{-4mm}
\end{figure*}

\begin{figure*}[t!]
    \centering
    \includegraphics[width=0.19\linewidth]{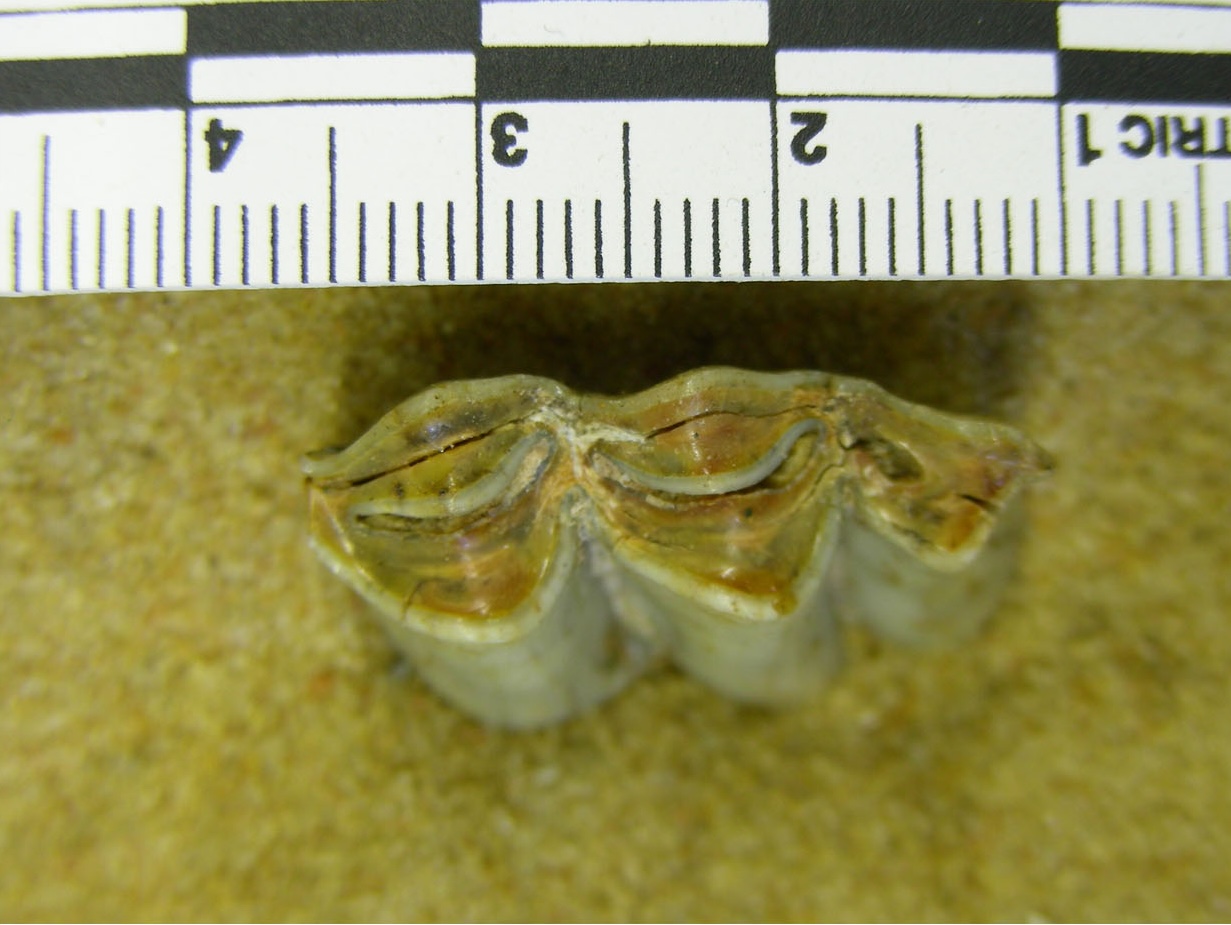} 
    \includegraphics[width=0.19\linewidth]{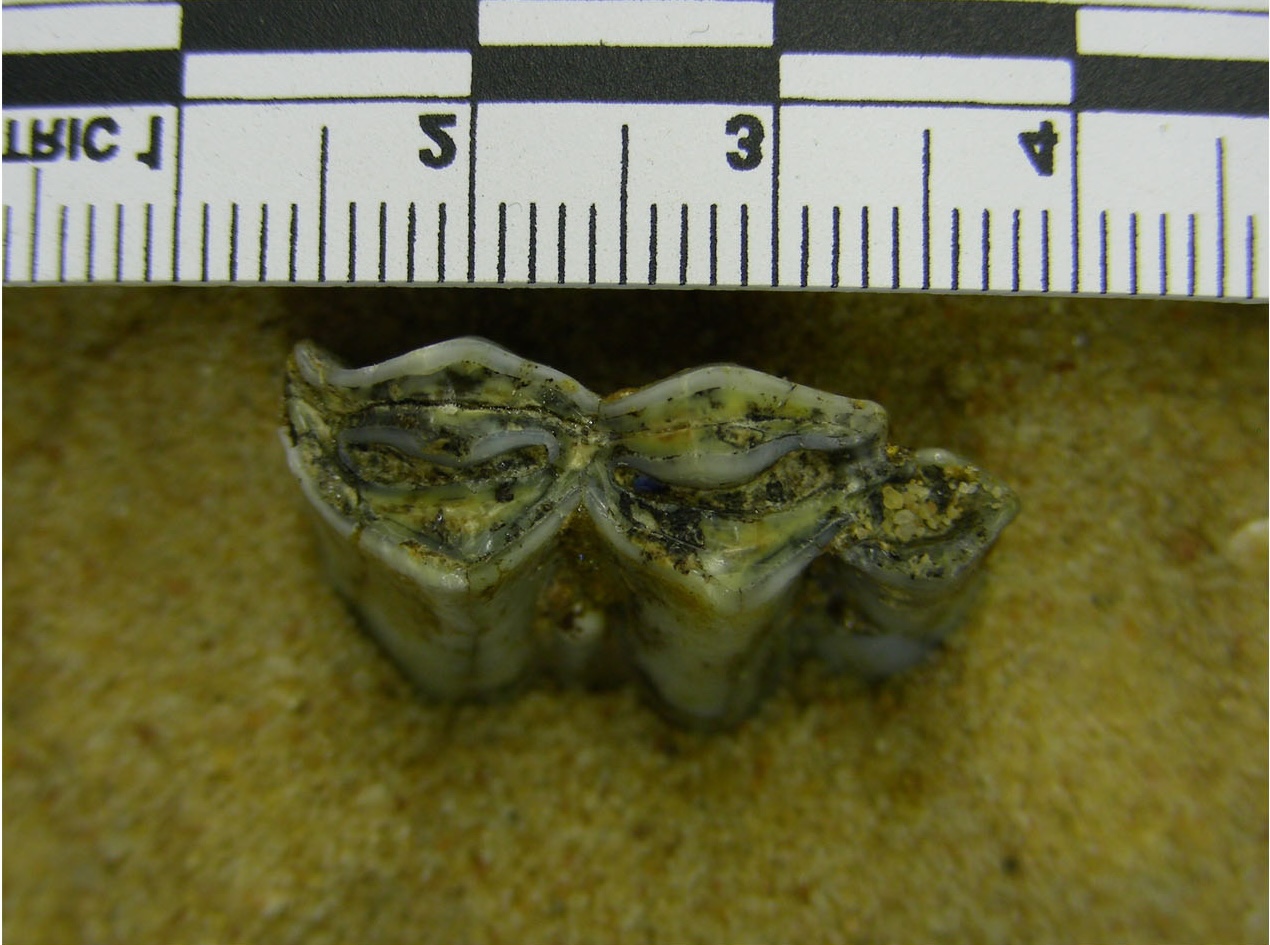}
    \includegraphics[width=0.19\linewidth]{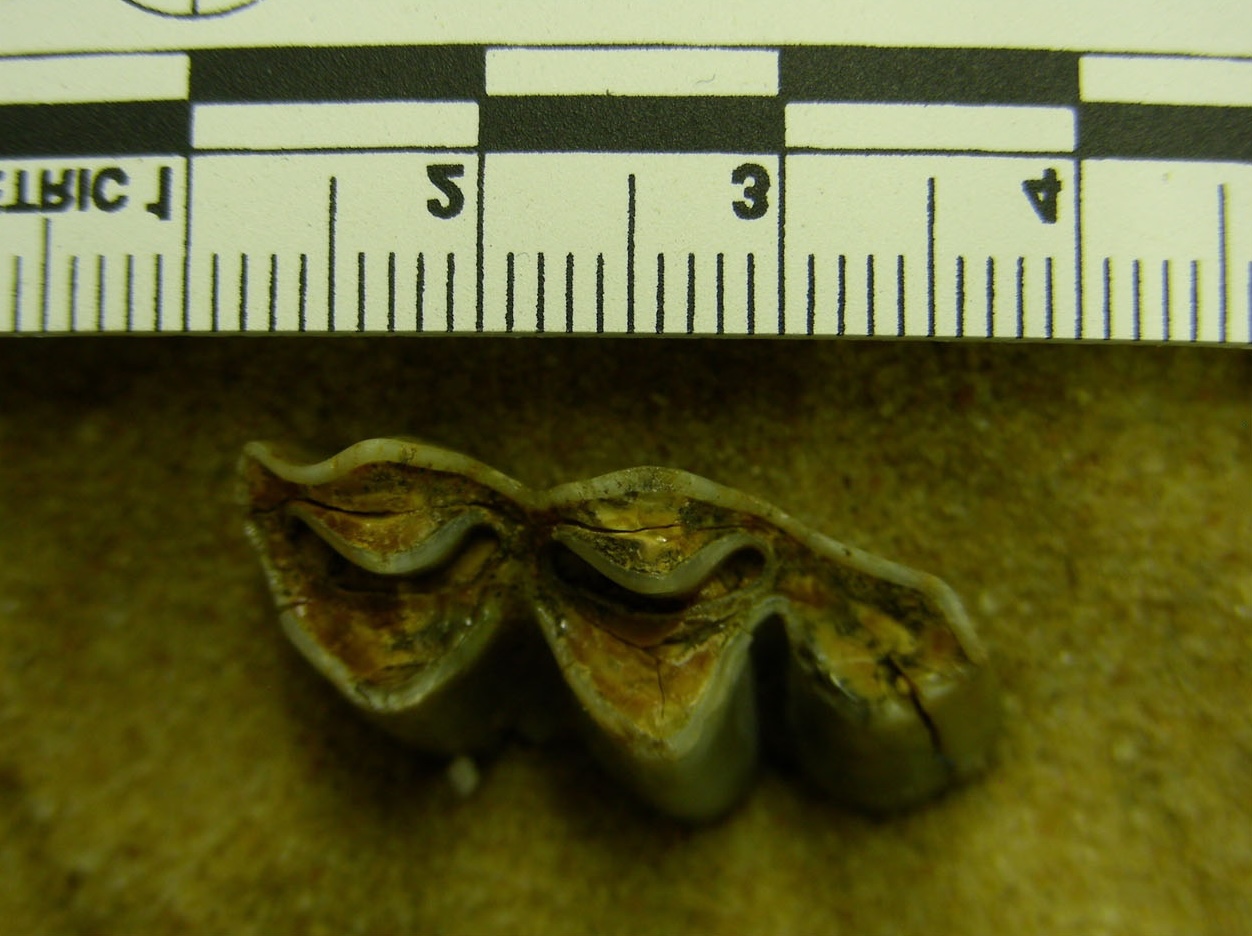}
    \includegraphics[width=0.19\linewidth]{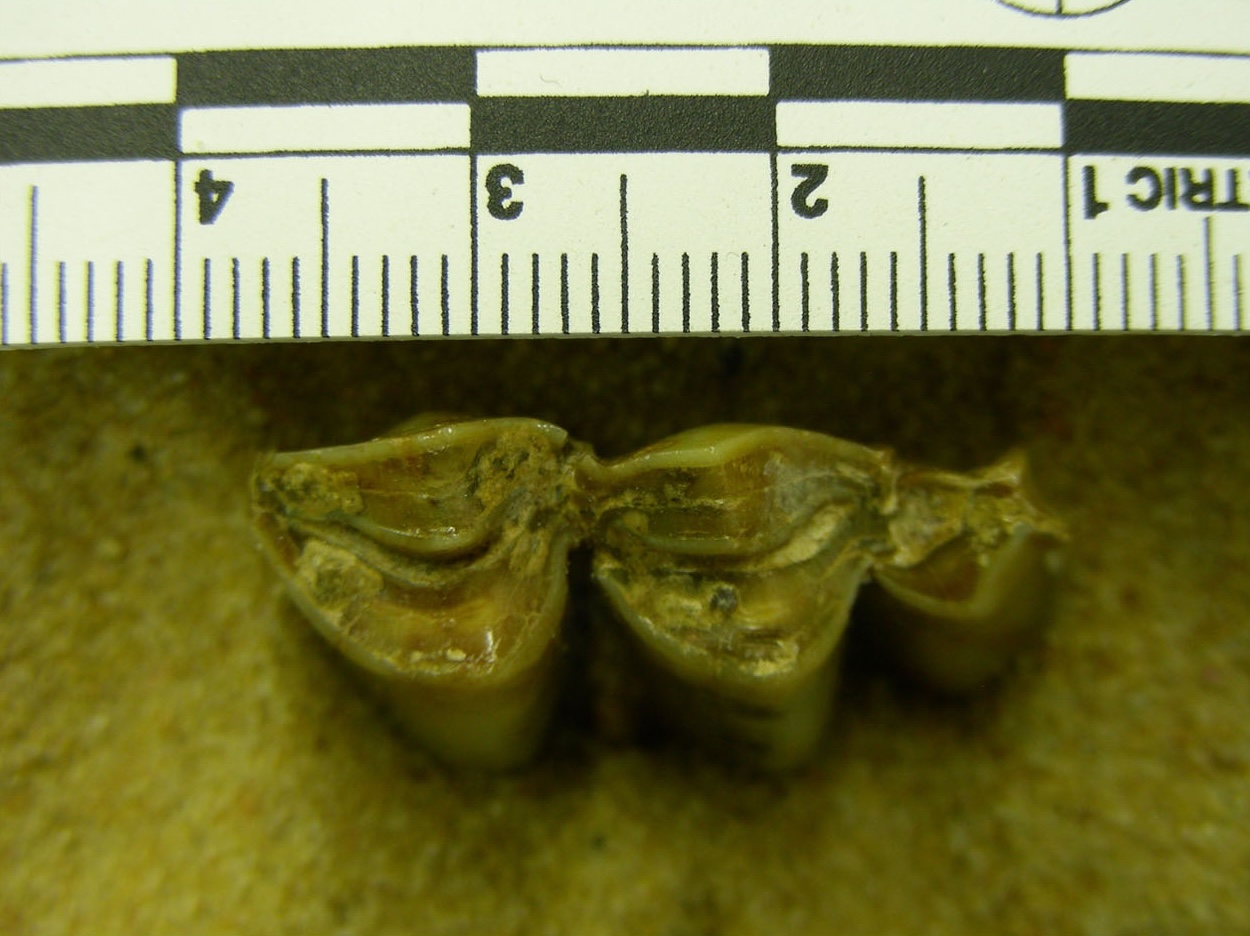}
    \includegraphics[width=0.19\linewidth]{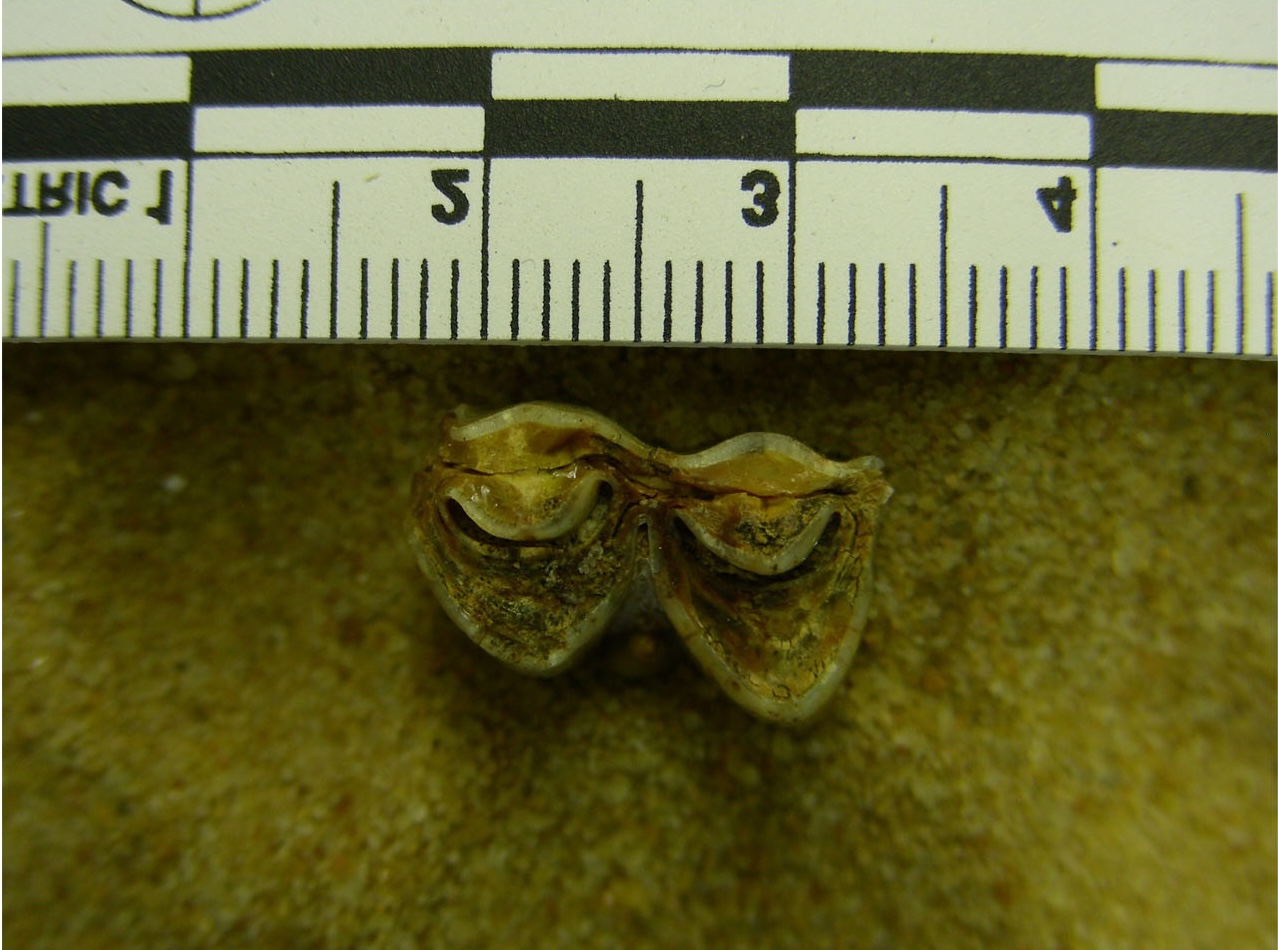}
    \\[-1mm]
    \includegraphics[width=0.19\linewidth]{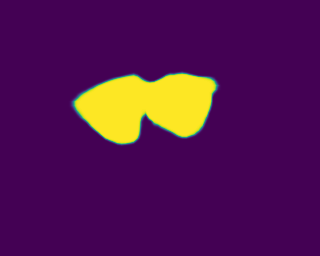} 
    \includegraphics[width=0.19\linewidth]{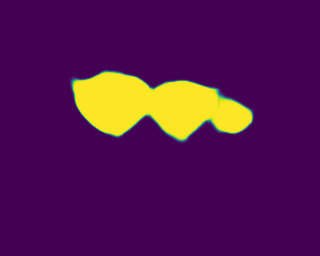}
    \includegraphics[width=0.19\linewidth]{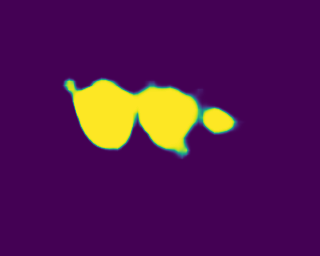}
    \includegraphics[width=0.19\linewidth]{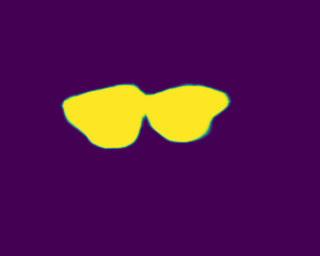}
    \includegraphics[width=0.19\linewidth]{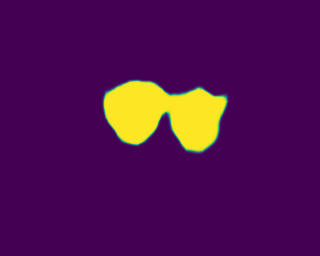}
    \caption{Inference on out-of-distribution samples captured during a summer 2026 research trip to South Africa using SegFormer MiT-B3 CLAHE $c=25$. Best viewed in color.}
    \label{fig:summer_2026_1} 
    \vspace{-4mm}
\end{figure*}

Next, Figure~\ref{fig:masks_perclass} provides visual examples of different (Raw Image, Mask Image) pairs for four distinct tribes, as well as the heatmap masks generated by SegFormer MiT-B3 and U-Net++ MBv2. Here we can more directly compare the Transformer and CNN: Alcelaphini teeth are relatively simple and despite the picture showing other teeth flanking the region of interest, both MiT-B3 and MBv2 capture the general area, however, MBv2 fails to properly reconstruct the shape convincingly. For Bovini, MiT-B3 almost perfectly captures the odd shape of the tooth, including the upside-down 'Y' on the bottom of it. For Neotragini, both models capture the general shape of the image, however, only MiT-B3 captures the finer details, such as the protrusions on top of the molars. The same holds true for Tragelaphini, although here, MBv2 falls short by a larger margin. 

\noindent\textbf{Model Size Consideration.} This type of behaviour, arguably, is expected, as it has been well-documented that Transformers tend to outperform CNNs on visual tasks~\cite{dosovitskiy2020image, xie2021segformer}. However, we must also consider that the MiT-B3 SegFormer is a much larger and more powerful computational model, containing 44.6M parameters to the 6.8M of MBv2, which is expressly designed for mobile deployment. When processing a $512\times512$ RGB image, SegFormer consumes 41.96 GFLOPs~\cite{mills2023aiop} to the 17.91 GFLOPs of MBv2, which an increase by a factor of 2.34. Given this, when coupled the dataset imperfections of B.O.V.I.D., the segmentation performance of the model is arguably still impressive.

\subsection{Generalization to New Data}

Finally, we provide some visual results about how our B.O.V.I.D. segmentation model can generalize to new, out-of-distribution (OOD) data. Specifically, we consider several new images, expressly not part of the original B.O.V.I.D. dataset, captured during a summer 2026 research trip to South Africa. Masks have not been drawn out for these images, so we are purely relying on the model, SegFormer MiT-B3 CLAHE $c=25$, to draw accurate masks of the teeth.

Figure~\ref{fig:summer_2026_1} provides sample pictures of raw images and estimated masks. In contrast to prior pictures, the background of these photos is different and the lighting can be darker. Despite this, the SegFormer MiT-B3 model with CLAHE does manage to competently mask off the appropriate regions of the image with two caveats: First, the model has a bias towards drawing the mask in the center of the image; second, the 1st and 4th images have a small third lobe (these are pictures of molar teeth) on their right hand side which are not adequately captured by the heatmap. Nevertheless, the model does manage to capture the protrusions on the 5th image's right hand side. Overall, these results further corroborate the efficacy of segmentation on this data.

\section{Related Work}

\noindent\textbf{Semantic Segmentation Deep Neural Networks.} Convolutional encoder-decoder networks remain a staple of pixel-level semantic segmentation, with residual networks such as ResNets~\cite{he2016deep}, MobileNets~\cite{sandler2018mobilenetv2} and EfficientNets~\cite{tan2019efficientnet} serving as a common CNN backbone for mobile and resource-constrained inference rather than training-data robustness. More recently, Vision Transformers~\cite{dosovitskiy2020image} and transformer-based segmentation heads such as SegFormer~\cite{xie2021segformer} demonstrate the efficacy of attention-based encoders. Our study draws directly on this architectural lineage by isolating architecture choice as a variable of interest.

\noindent\textbf{Segmentation Benchmarks and Annotation Quality.}
The design of widely used segmentation datasets such as Cityscapes~\cite{Cordts2016Cityscapes} and ADE20K~\cite{zhou2017ADE} is geared towards ML training, with annotation protocols designed around pixel-accurate, model-ready masks. This is a meaningful point of departure for the B.O.V.I.D. dataset, whose masks were produced by domain experts for morphometric analysis~\cite{brophy2021comparative} rather than for supervised learning, and which consequently exhibit unique issues uncommon in ML-geared semantic segmentation datasets. Rather than proposing a new benchmark or new correction algorithm, we contribute a comparative evaluation of how existing, well-established architectures tolerate this brand of annotation noise, a question that is largely orthogonal to, and complements, prior benchmark-driven architecture comparisons.

\noindent\textbf{Preprocessing and Alignment.} Contrast-limited adaptive histogram equalization is a standard technique for improving local contrast in domains with inconsistent or not ideal lighting or acquisition conditions, and we adopt it as one of the several preprocesing modes evaluated in this work. To address spatial misalignment specifically, we draw on foreground-centering and common-reference-frame registration principles, applying a lightweight, non-learned centroid-alignment step rather than a full deformable registration pipeline, consistent with the scale of our dataset and the goal of the ablation.

\section{Conclusion}

This paper presents a comparative study of convolutional and transformer-based segmentation architectures as well as preprocessing techniques on B.O.V.I.D., a dentation dataset with handcrafted masks for morphometric analysis instead of ML training. We find that lightweight preprocessing makes the dataset viable for supervised segmentation despite its resolution mismatches and raw-mask misalignment, and that quantative metrics like Dice and mIoU are largely insensitive to the choice of contrast-processing strategy even though qualitative differences in predicted masks are substantial. Finally, we find that a lightweight CNN encoder can achieve segmentation quality broadly comparable to a larger transformer-based model. Results on out-of-distribution field data demonstrate method generalizability. %

{
    \small
    \bibliographystyle{ieeenat_fullname}
    \bibliography{main}
}

\newpage
\begin{figure*}[t!] 
    \centering 
    \begin{subfigure}[t]{0.3\linewidth} \centering \includegraphics[width=\linewidth]{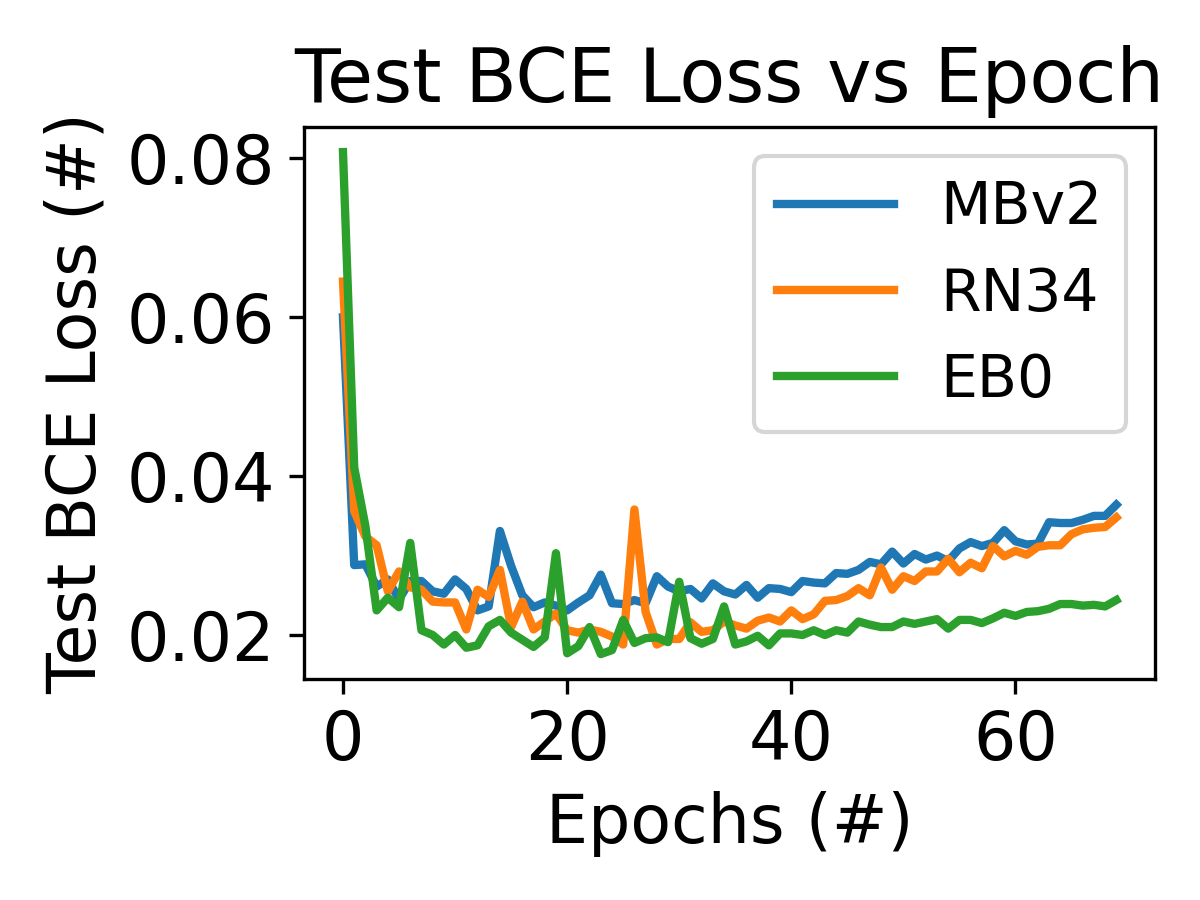} %
    \end{subfigure} \hfill \begin{subfigure}[t]{0.3\linewidth} \centering \includegraphics[width=\linewidth]{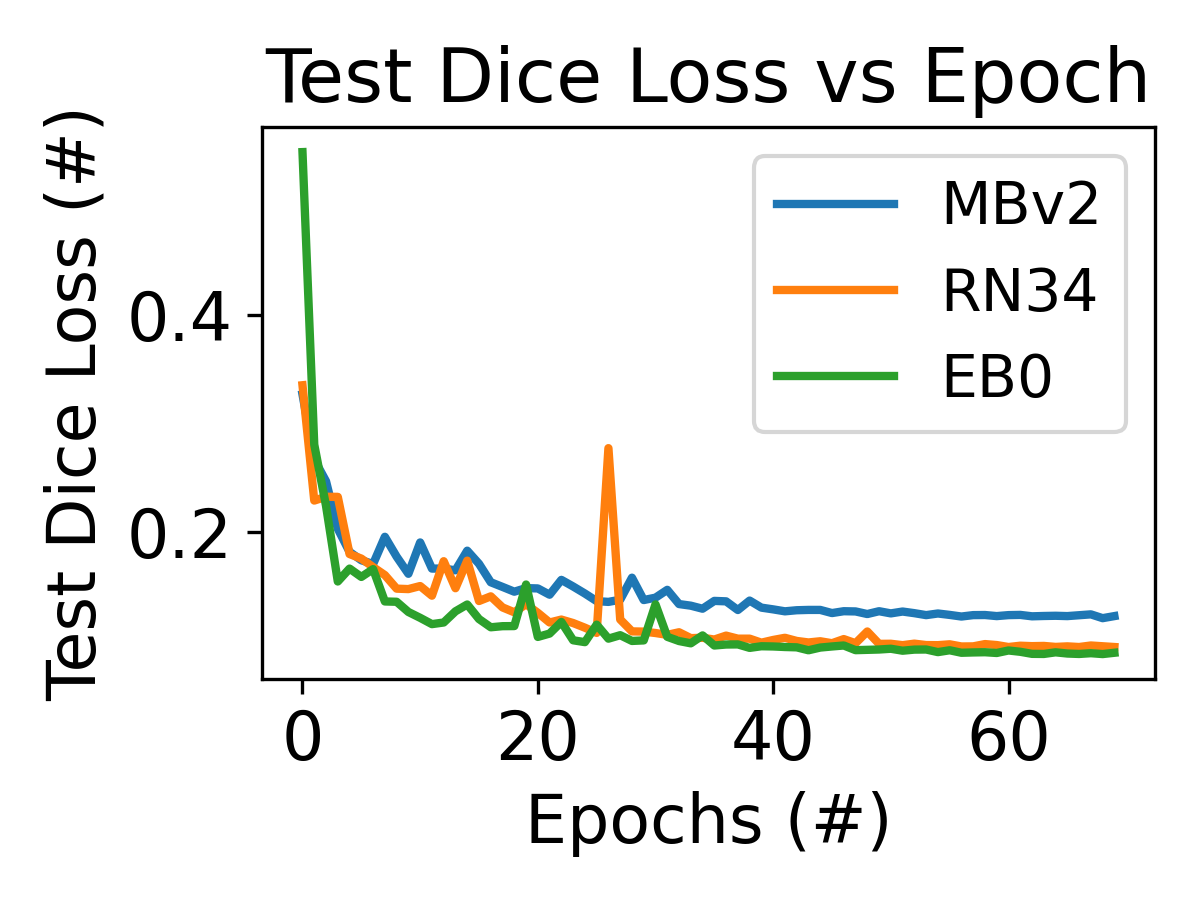} %
    \end{subfigure} \hfill \begin{subfigure}[t]{0.3\linewidth} \centering \includegraphics[width=\linewidth]{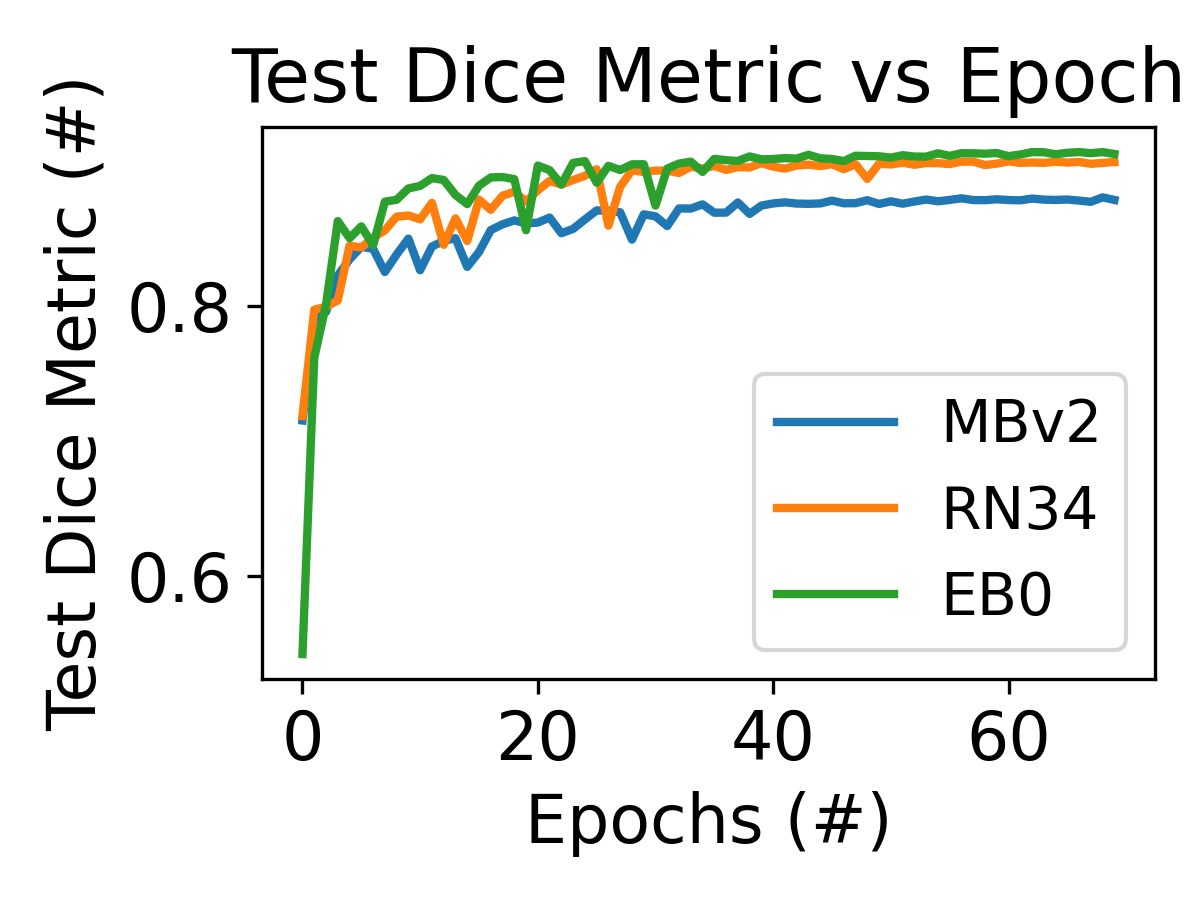} %
    \end{subfigure} \caption{BCE loss, Dice loss and Dice metric for U-Net++ MBv2 vs. RN34 vs. EB0. For losses, lower is better, while higher is better for the Dice metric.} \label{fig:cnns}
\end{figure*}

\begin{figure*}[t!] \centering \setlength{\tabcolsep}{2pt} \begin{tabular}{ccccc} \textbf{Raw Image} & \textbf{Mask Image} & \textbf{MobileNetV2} & \textbf{ResNet-34} & \textbf{EfficientNet-B0} \\[2mm]

    \includegraphics[width=0.19\textwidth,height=2.6cm,keepaspectratio]{fig/mbv2_main_img/DSCN2907.JPG} & \includegraphics[width=0.19\textwidth,height=2.6cm,keepaspectratio]{fig/mbv2_main_img/DSCN2907.png} &  \includegraphics[width=0.19\textwidth,height=2.6cm,keepaspectratio]{fig/mbv2_main_img/ne/DSCN2907_mask.png} & \includegraphics[width=0.19\textwidth,height=2.6cm,keepaspectratio]{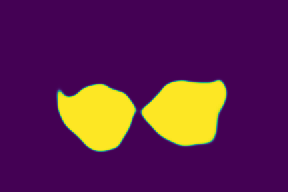} & \includegraphics[width=0.19\textwidth,height=2.6cm,keepaspectratio]{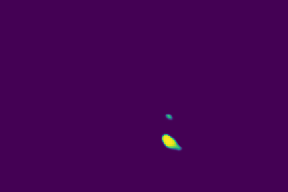} \\[1mm]

    \includegraphics[width=0.19\textwidth,height=2.6cm,keepaspectratio]{fig/mbv2_main_img/DSCN3547.JPG} & \includegraphics[width=0.19\textwidth,height=2.6cm,keepaspectratio]{fig/mbv2_main_img/DSCN3547.png} &  \includegraphics[width=0.19\textwidth,height=2.6cm,keepaspectratio]{fig/mbv2_main_img/ne/DSCN3547_mask.png} & \includegraphics[width=0.19\textwidth,height=2.6cm,keepaspectratio]{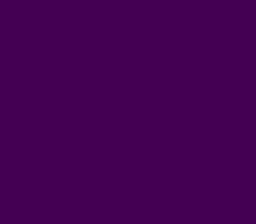} & \includegraphics[width=0.19\textwidth,height=2.6cm,keepaspectratio]{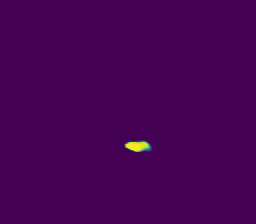} \\[1mm]

    \includegraphics[width=0.19\textwidth,height=2.6cm,keepaspectratio]{fig/mbv2_main_img/DSCN5104.JPG} & \includegraphics[width=0.19\textwidth,height=2.6cm,keepaspectratio]{fig/mbv2_main_img/DSCN5104.png} &  \includegraphics[width=0.19\textwidth,height=2.6cm,keepaspectratio]{fig/mbv2_main_img/ne/DSCN5104_mask.png} & \includegraphics[width=0.19\textwidth,height=2.6cm,keepaspectratio]{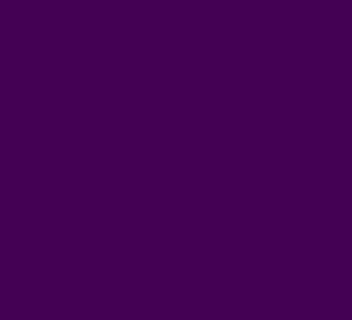} & \includegraphics[width=0.19\textwidth,height=2.6cm,keepaspectratio]{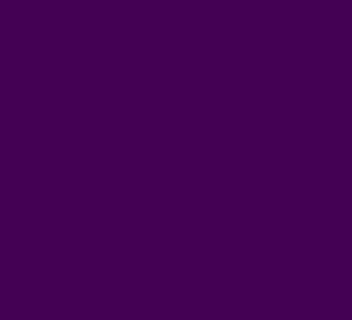} \\[1mm]

    \includegraphics[width=0.19\textwidth,height=2.6cm,keepaspectratio]{fig/mbv2_main_img/IMG_5300.JPG} & \includegraphics[width=0.19\textwidth,height=2.6cm,keepaspectratio]{fig/mbv2_main_img/IMG_5300.png} &  \includegraphics[width=0.19\textwidth,height=2.6cm,keepaspectratio]{fig/mbv2_main_img/ne/IMG_5300_mask.png} & \includegraphics[width=0.19\textwidth,height=2.6cm,keepaspectratio]{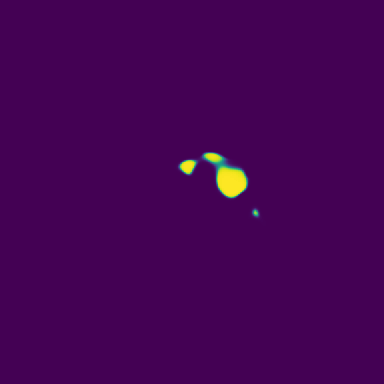} & \includegraphics[width=0.19\textwidth,height=2.6cm,keepaspectratio]{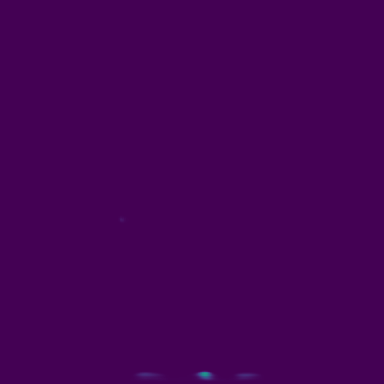} \\[1mm]
    
    \end{tabular} 
    \caption{Example Raw Images (including whitespace) and ground-truth Mask Images as well as inferred heatmaps by MBv2, ResNet-34 and EfficientNet-B0 without any histogram equalization. Yellow represents regions with high ($\sim1$) likelihood of being part of the tooth; purple represents low likelihood. Best viewed in color.} \label{fig:masks_models} 
\end{figure*}

\appendix

\section*{Supplementary Materials}

We enumerate the hyperparameters considered in our experiments as well as provide additional ablations and visualizations. 

\section{Experimental Hyperparameters}

We use the segmentation-models-pytorch~\cite{iakubovskii2019segmentation} library to implement our models and conduct all of our experiments on a single Nvidia DGX Spark unit. We train our models for 70 epochs with a batch size of 8 or 6 (depends on model size, larger if possible) using the AdamW optimizer. For CNNs, we use a default learning rate of $1e^{-3}$ and for transformers a default learning rate of $1e^{-4}$ is utilized, and in both cases a weight decay factor of $1e^{-5}$ is utilized alongside a gradient norm clipping parameter of $1.0$. Per Sec.~\ref{sec:aug} we downsample images and masks by a factor of $4$.

We maintain an 80\%/20\% training/testing split. When per-tribe stratification is enabled, filenames are mapped to tribes using an external CSV file, and then a split is performed using this information to preserve the proportion of data belonging to each tribe. Prior to this, data is shuffled using a random seed which we control. All stratification experiments in Sec.~\ref{sec:strat} are conducted using three distinct random seeds where we take the mean and standard deviation of the results, while mask images are simply pulled from the model produced by the first random seed.

\section{CNN Model Comparison}

Due to space constraints, we now provide an comparative study of different CNN backbones, specifically comparing MobileNetV2 (MBv2)~\cite{sandler2018mobilenetv2} to the older, yet larger ResNet-34 (RN34)~\cite{he2016identity} and more advanced EfficientNet-B0 (EB0)~\cite{tan2019efficientnet}. These experiments are conducted without any preprocessing, i.e., they are all `NE'. Figure~\ref{fig:cnns} mirrors Figs.~\ref{fig:segformer_main} and \ref{fig:mbv2_main} by plotting the test BCE loss, Dice loss and Dice score metric across model training. The most interesting phenomenon here is that MBv2 lags behind both ResNet-34 and EfficientNet-B0 on all three metrics to a noticeable degree. Specifically, EfficientNet-B0 achieves by far the lowest BCE loss but is roughly tied with ResNet-34 on the Dice loss and score metric, while MBv2 scores noticeably worse in these cases. This finding suggests that both ResNet-34 and EfficientNet-B0 would be better suited to masking B.O.V.I.D., however, we must also consider the qualitative aspect as well.

Figure~\ref{fig:masks_models} provides sample images of the mask heatmaps produced by different CNN backbones on this task. As we can see, MBv2 is the only model that can consistently produce a result which vaguely resembles the ground truth Mask Image. ResNet-34 does a decent job on the first image, but then either produces nothing or just a blunch of blotches on the heatmap, while EfficientNet-B0 fails to even visualize the first mask at all.

\section{Ablation Study on $\lambda_{Dice}$}

\begin{figure*}[t!] 
    \centering 
    \begin{subfigure}[t]{0.3\linewidth} \centering \includegraphics[width=\linewidth]{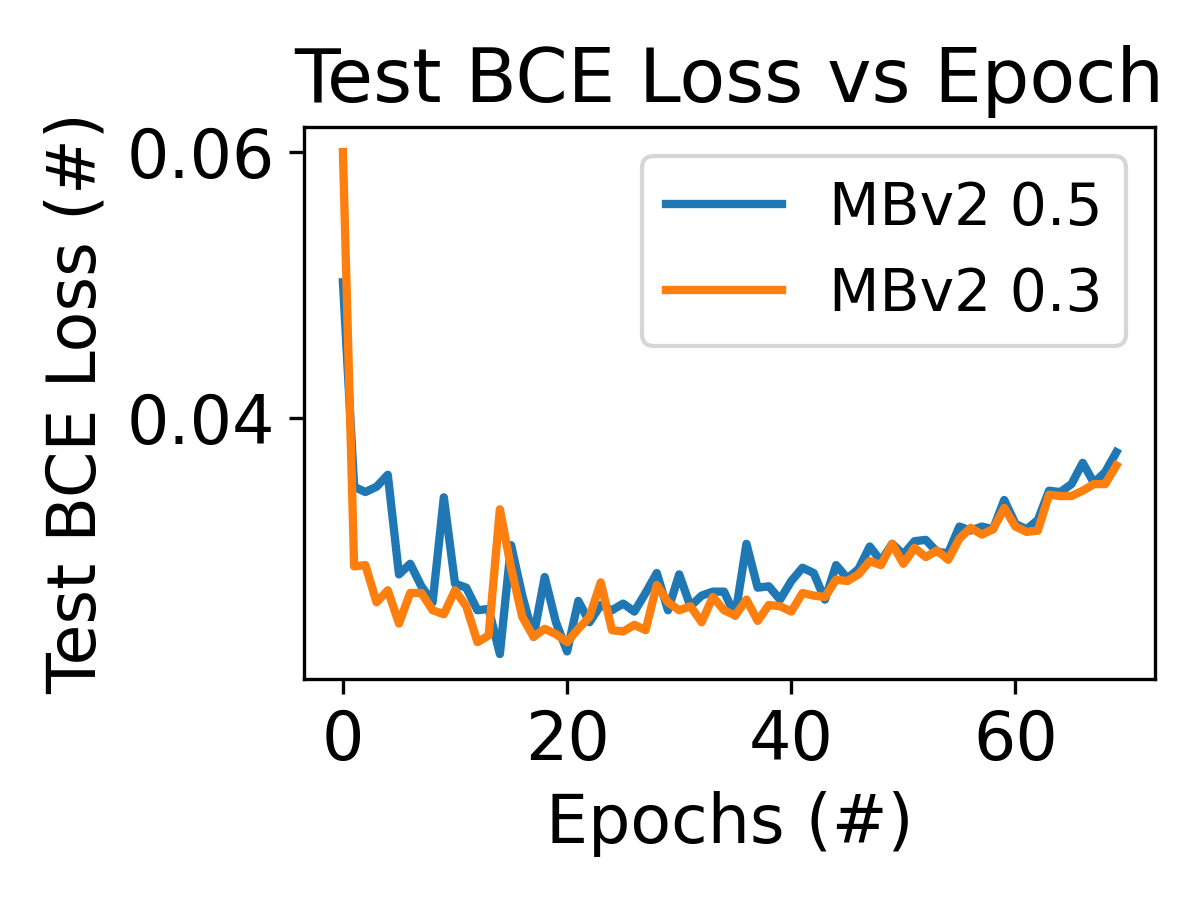} %
    \end{subfigure} \hfill \begin{subfigure}[t]{0.3\linewidth} \centering \includegraphics[width=\linewidth]{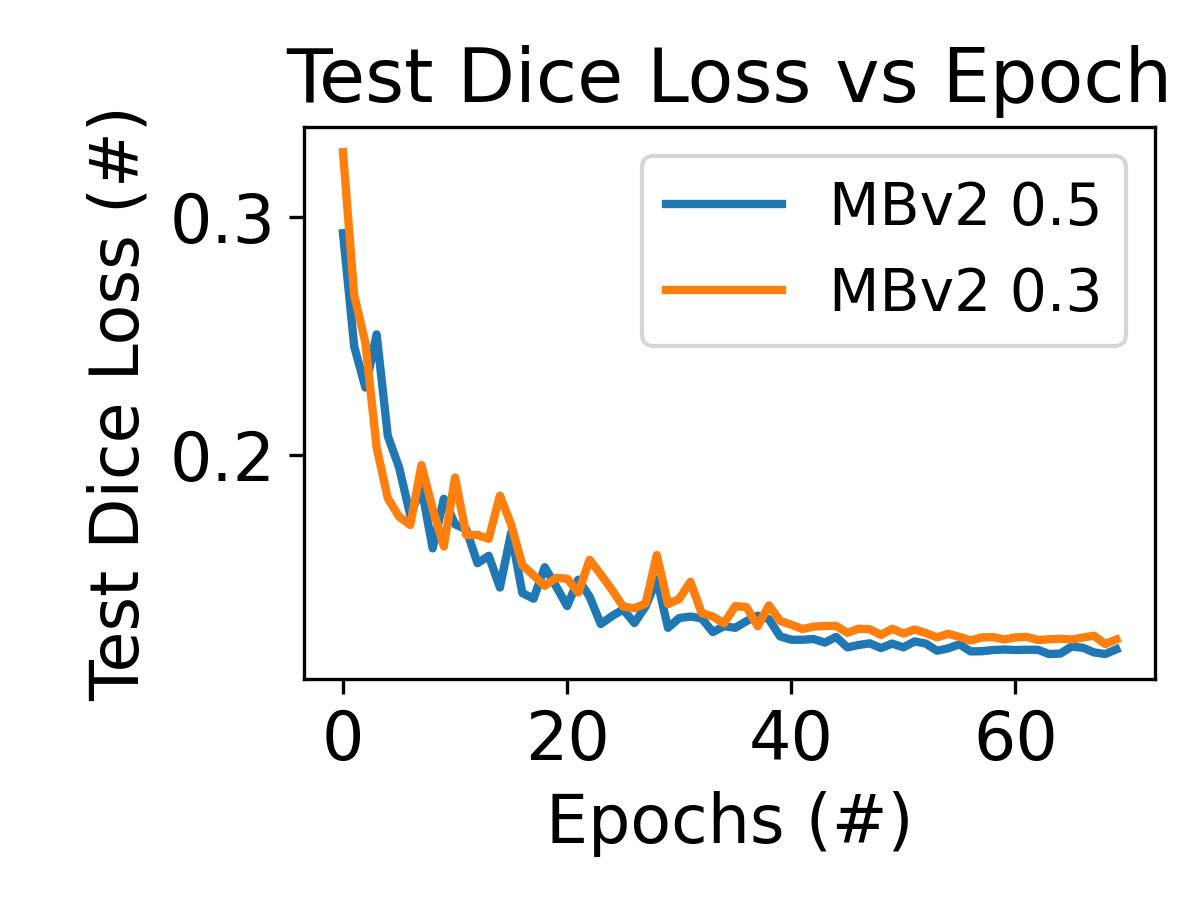} %
    \end{subfigure} \hfill \begin{subfigure}[t]{0.3\linewidth} \centering \includegraphics[width=\linewidth]{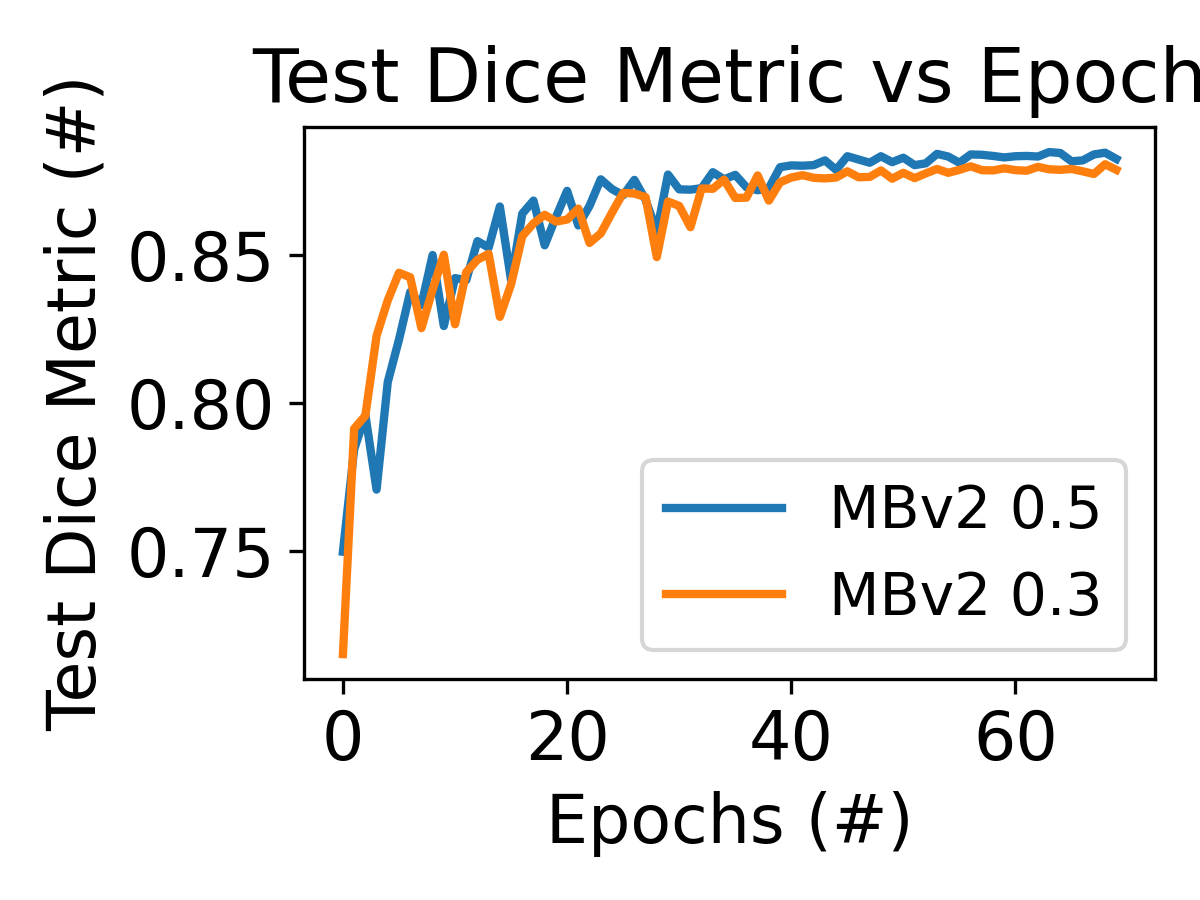} %
    \end{subfigure} \caption{BCE loss, Dice loss and Dice metric for U-Net++ MBv2 while varying $\lambda_{Dice}$. For losses, lower is better, while higher is better for the Dice metric. Best viewed in color.} \label{fig:mbv2_ld}
\end{figure*}

We now provide an ablation over the hyperparameter $\lambda_{Dice}$, primarily on MBv2. Figure~\ref{fig:mbv2_ld} plots our findings for the BCE loss, Dice loss and Dice score metric for $\lambda_{Dice}\in\{0.3, 0.5\}$. Unsurprisingly, the stronger value of $\lambda_{Dice}=0.5$ provides lower Dice loss and higher dice metric, yet higher BCE loss. 

\begin{figure*}[t] \centering \setlength{\tabcolsep}{2pt} \begin{tabular}{ccccc} \textbf{Raw Image} & \textbf{Mask Image} & $\mathbf{\lambda_{Dice}=1.0}$ & $\mathbf{\lambda_{Dice}=0.5}$ & $\mathbf{\lambda_{Dice}=0.3}$  \\[2mm]     
    \includegraphics[width=0.19\textwidth,height=2.6cm,keepaspectratio]{fig/mbv2_main_img/DSCN2907.JPG} &
    \includegraphics[width=0.19\textwidth,height=2.6cm,keepaspectratio]{fig/mbv2_main_img/DSCN2907.png} & \includegraphics[width=0.19\textwidth,height=2.6cm,keepaspectratio]{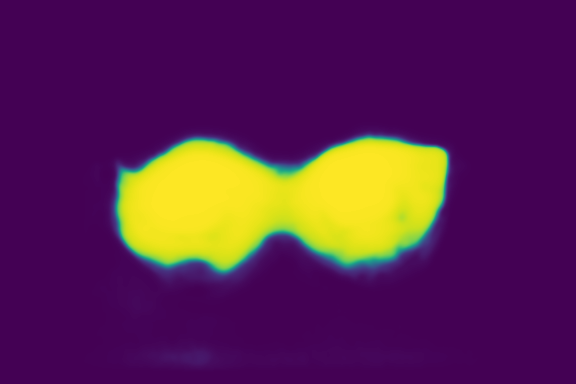} & \includegraphics[width=0.19\textwidth,height=2.6cm,keepaspectratio]{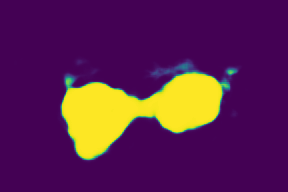} & \includegraphics[width=0.19\textwidth,height=2.6cm,keepaspectratio]{fig/mbv2_main_img/ne/DSCN2907_mask.png} \\[1mm]

    \includegraphics[width=0.19\textwidth,height=2.6cm,keepaspectratio]{fig/mbv2_main_img/DSCN3547.JPG} &
    \includegraphics[width=0.19\textwidth,height=2.6cm,keepaspectratio]{fig/mbv2_main_img/DSCN3547.png} & \includegraphics[width=0.19\textwidth,height=2.6cm,keepaspectratio]{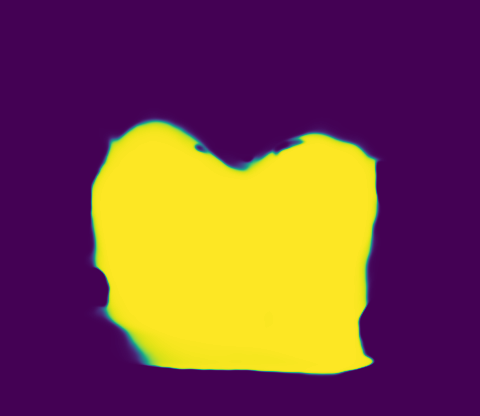} & \includegraphics[width=0.19\textwidth,height=2.6cm,keepaspectratio]{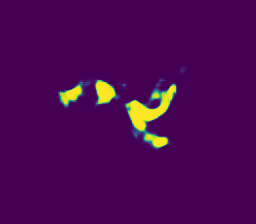} & \includegraphics[width=0.19\textwidth,height=2.6cm,keepaspectratio]{fig/mbv2_main_img/ne/DSCN3547_mask.png} \\[1mm]

    \includegraphics[width=0.19\textwidth,height=2.6cm,keepaspectratio]{fig/mbv2_main_img/DSCN5104.JPG} &
    \includegraphics[width=0.19\textwidth,height=2.6cm,keepaspectratio]{fig/mbv2_main_img/DSCN5104.png} & \includegraphics[width=0.19\textwidth,height=2.6cm,keepaspectratio]{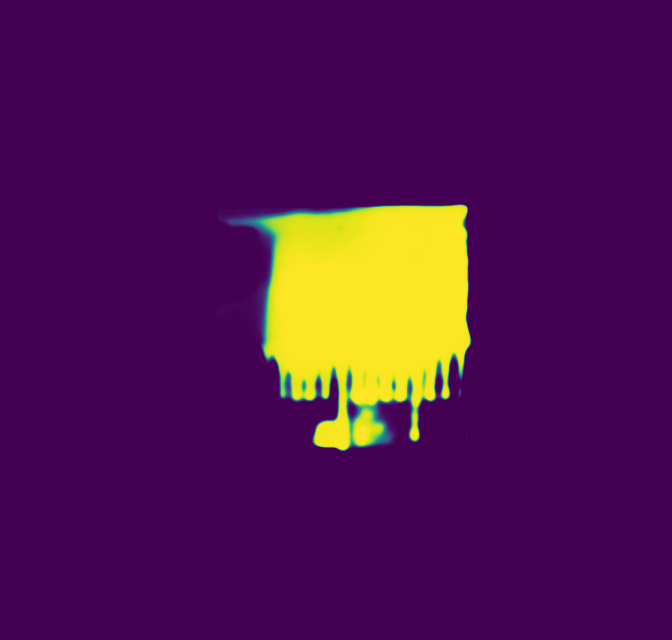} & \includegraphics[width=0.19\textwidth,height=2.6cm,keepaspectratio]{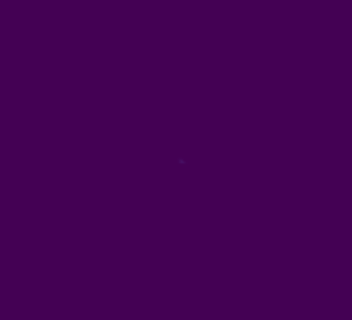} & \includegraphics[width=0.19\textwidth,height=2.6cm,keepaspectratio]{fig/mbv2_main_img/ne/DSCN5104_mask.png} \\[1mm]

    \includegraphics[width=0.19\textwidth,height=2.6cm,keepaspectratio]{fig/mbv2_main_img/IMG_5300.JPG} &
    \includegraphics[width=0.19\textwidth,height=2.6cm,keepaspectratio]{fig/mbv2_main_img/IMG_5300.png} & \includegraphics[width=0.19\textwidth,height=2.6cm,keepaspectratio]{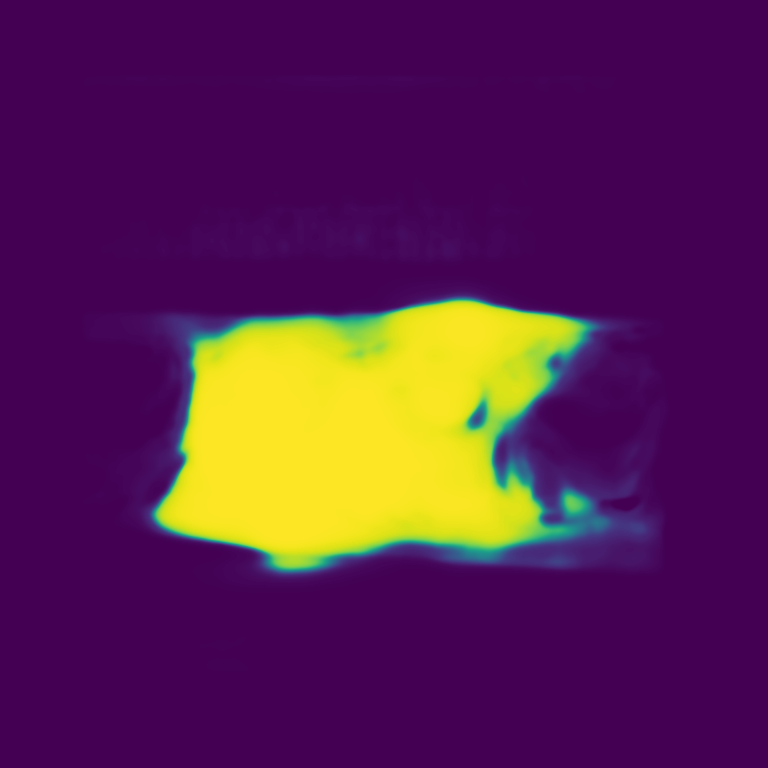} & \includegraphics[width=0.19\textwidth,height=2.6cm,keepaspectratio]{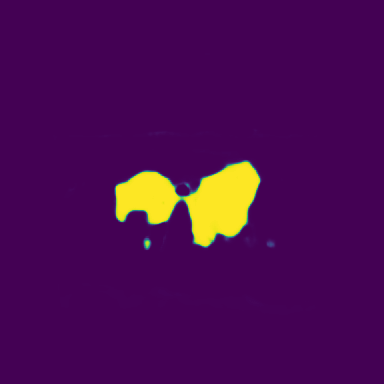} & \includegraphics[width=0.19\textwidth,height=2.6cm,keepaspectratio]{fig/mbv2_main_img/ne/IMG_5300_mask.png} \\[1mm] 
    \end{tabular} 
    \caption{Example Raw Images (including whitespace) and ground-truth Mask Images as well as inferred heatmaps by MBv2 without any histogram equalization while varying $\lambda_{Dice}$. Yellow represents regions with high ($\sim1$) likelihood of being part of the tooth; purple represents low likelihood. Best viewed in color.} \label{fig:masks_mbv2_ld} 
\end{figure*}

Next, Figure~\ref{fig:masks_mbv2_ld} provides sample mask heatmap inferences for MBv2 without histogram equalization for different values of $\lambda_{Dice}$. When we set the Dice coefficient too high, i.e., to $1.0$, the model tends to overcompensate and draw excessively large masks that cover more than the region of interest and also have badly defined borders, as evidenced by the amount of green intermittent value pixels in the heatmaps. However, when $\lambda_{Dice}=0.5$, the model will sometimes not even draw a mask heatmap, and when it does, it will usually be inaccurate. That leaves the most optimal value we considered, $\lambda_{Dice}=0.3$ which suggests that the BCE loss may actually play a silent, but important role in properly identifying regions of interest for this problem. 

\section{Per-Tribe MBv2 Statistics}

Table~\ref{tab:mbv2_b3_per_class} reports stratified, per-class statistics for MBv2. Unlike SegFormer MiT-B3, the results are less competitive as CLAHE $c=25$ does not obtain a clear majority for either metric, as both CLAHE $c=10$ and NE achieve the best performance on some tribes, while HE is never the best. Moreover, the results are consistent from Dice to mIoU, which was not always the case for SegFormer MiT-B3.

\section{Additional OOD Examples}

\begin{figure*}[t!]
    \centering
    \includegraphics[width=0.19\linewidth]{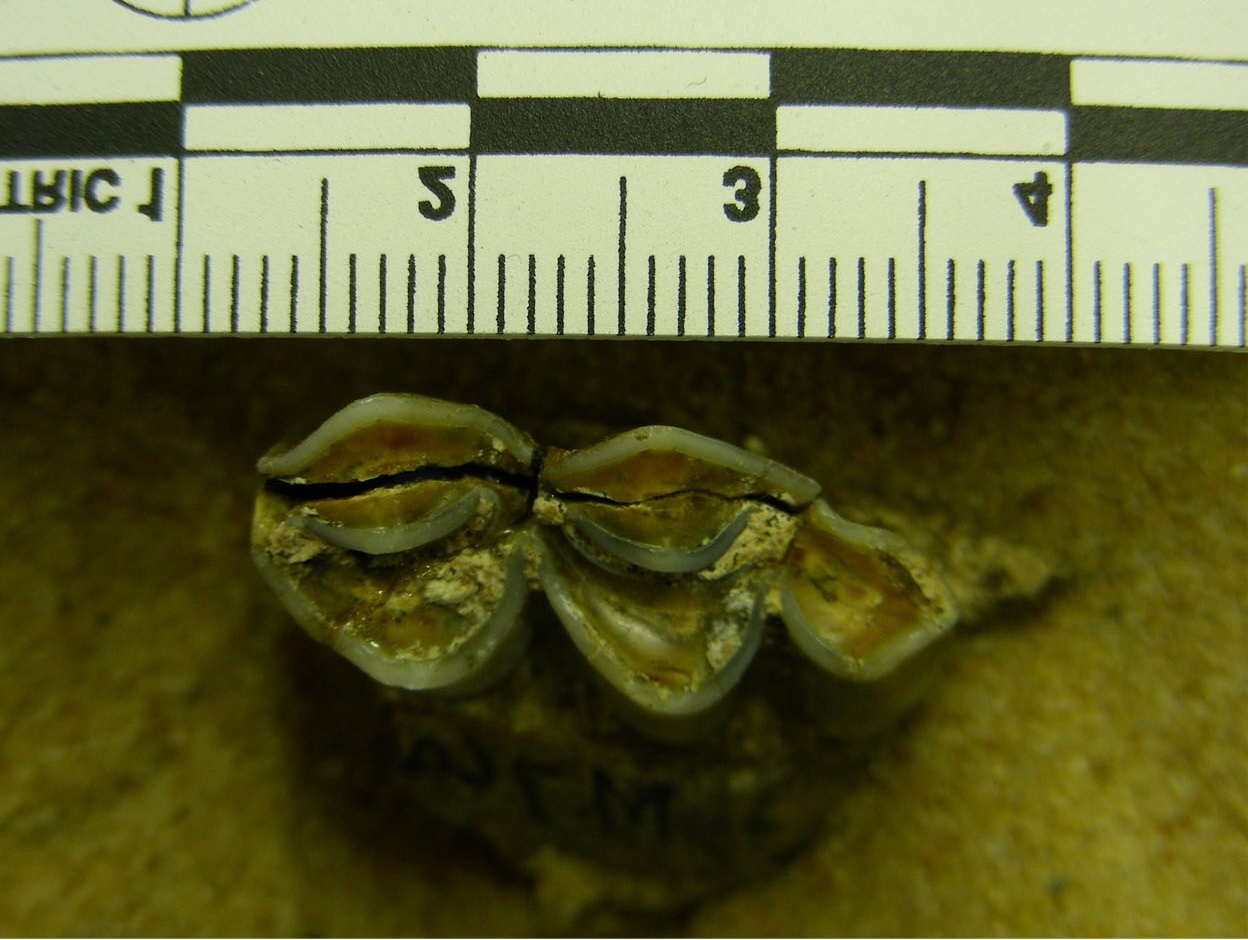} 
    \includegraphics[width=0.19\linewidth]{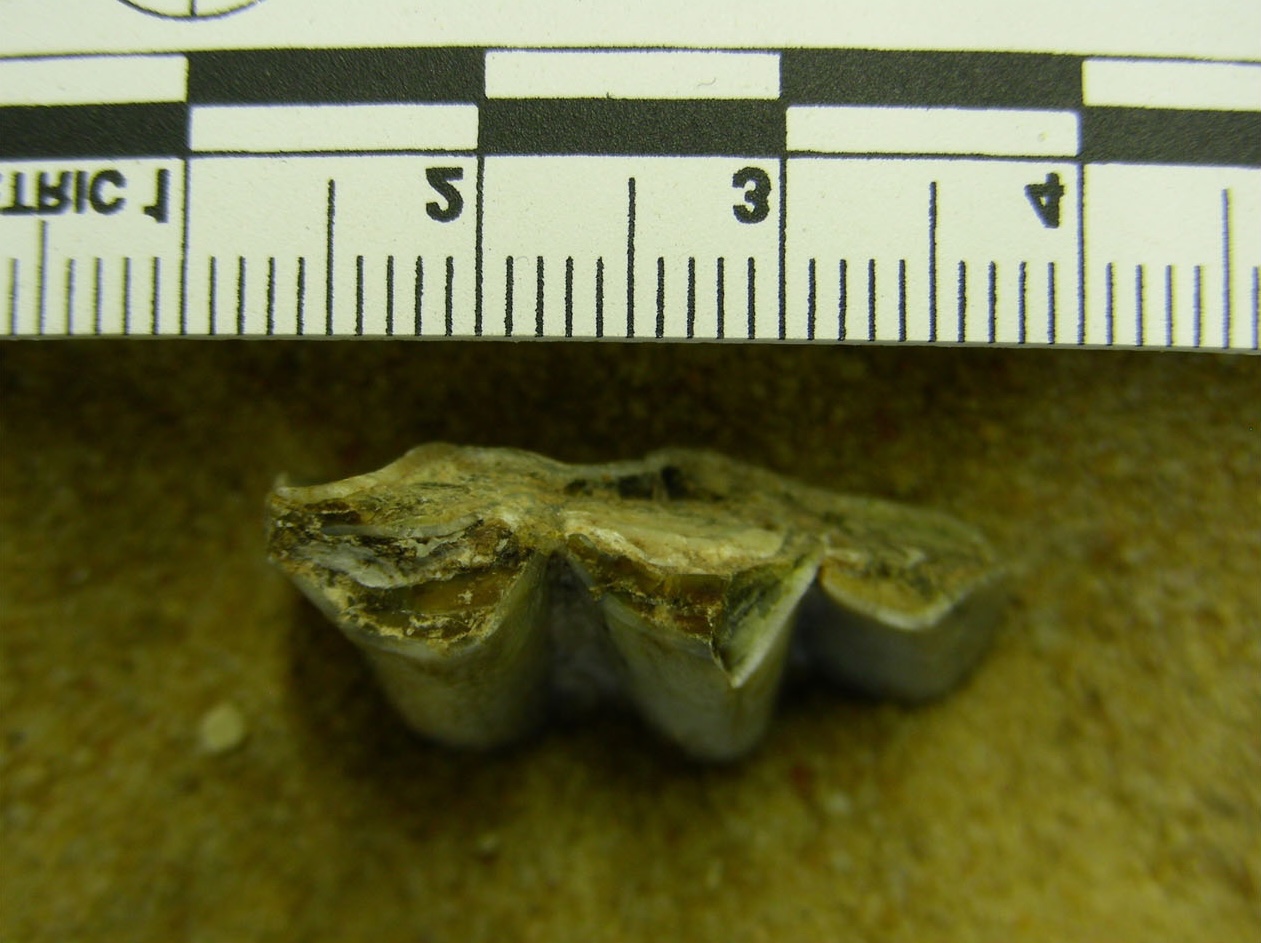}
    \includegraphics[width=0.19\linewidth]{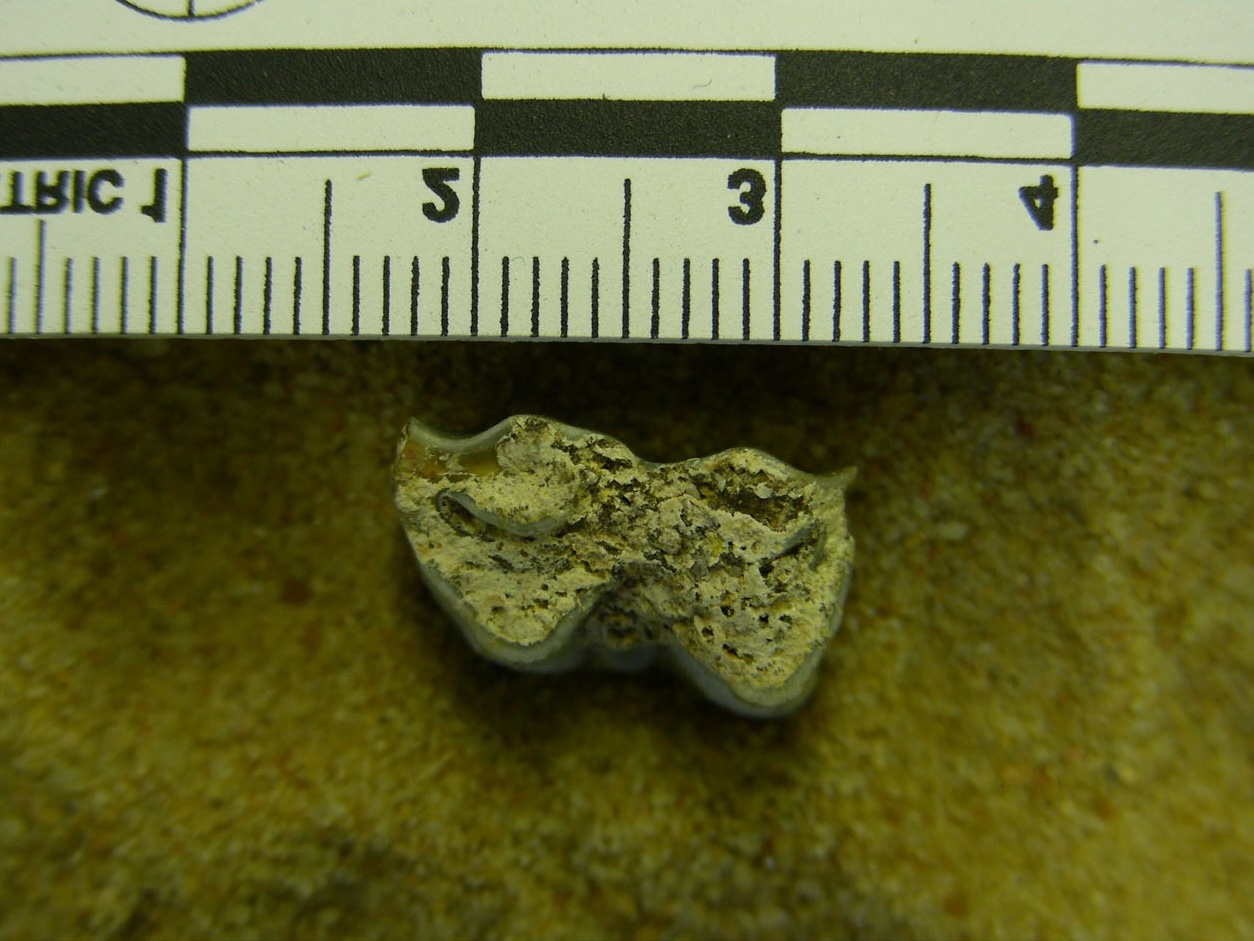}
    \includegraphics[width=0.19\linewidth]{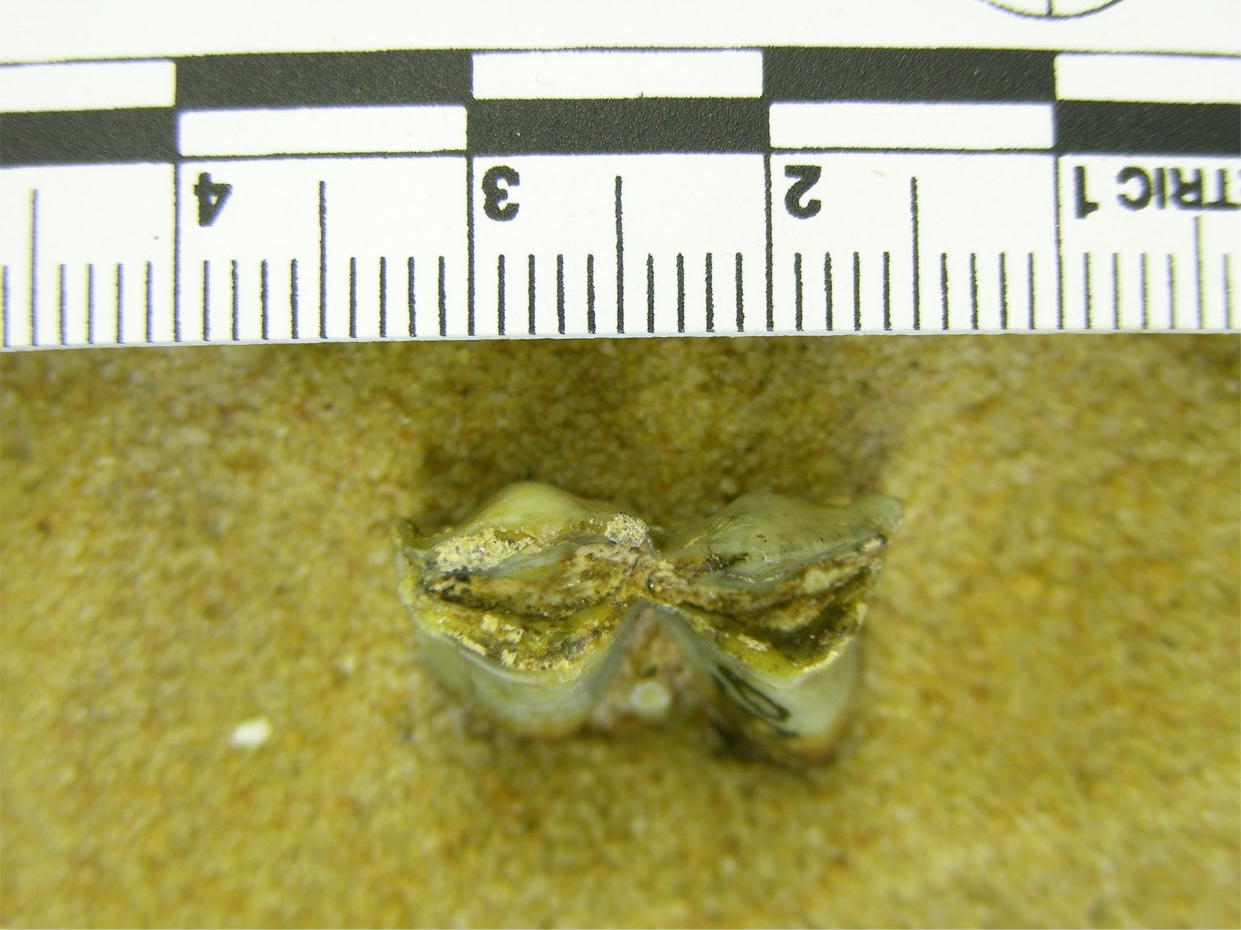}
    \includegraphics[width=0.19\linewidth]{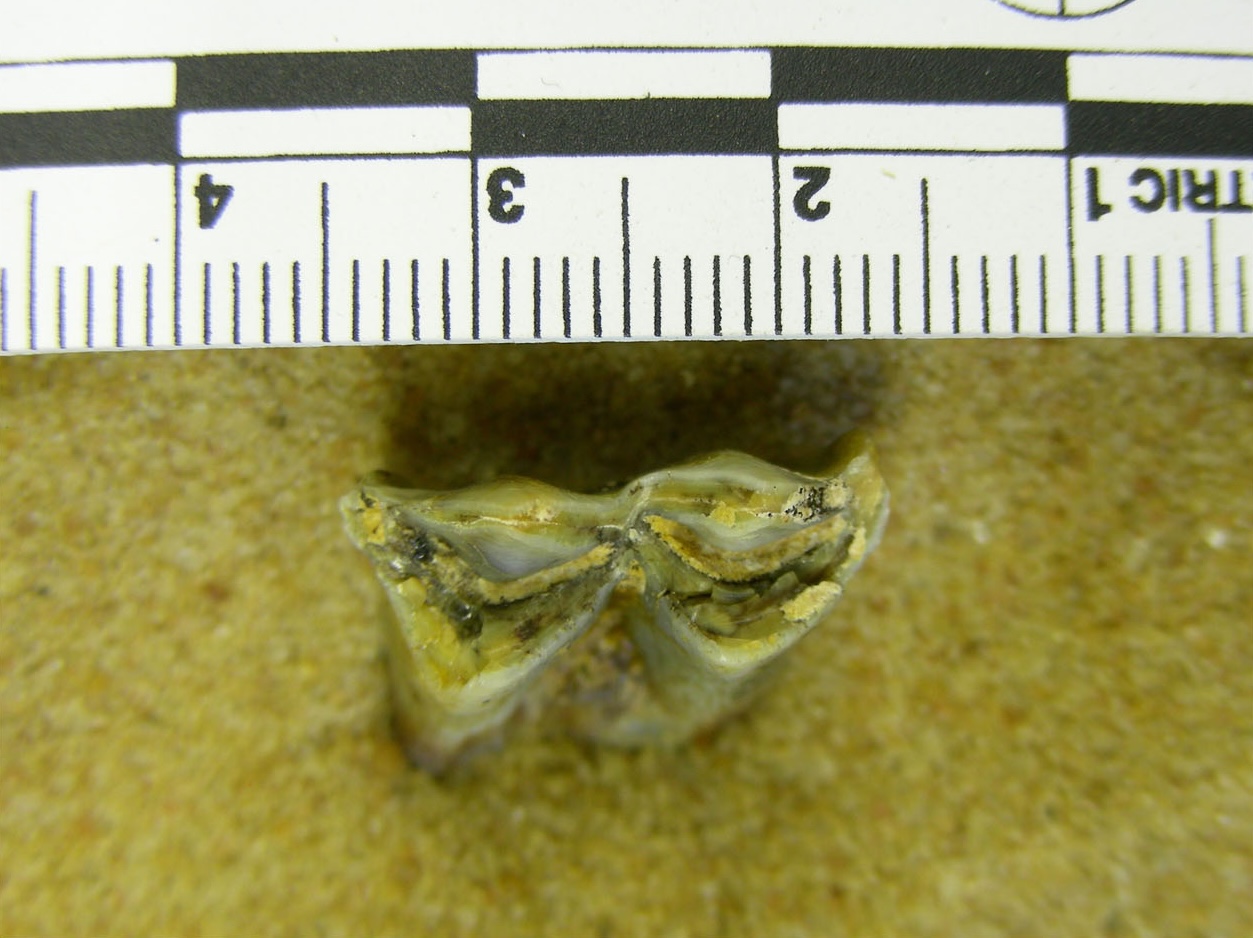}
    \\[-1mm]
    \includegraphics[width=0.19\linewidth]{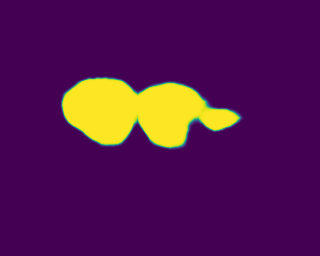} 
    \includegraphics[width=0.19\linewidth]{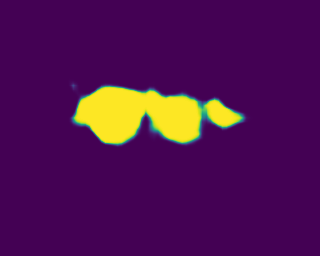}
    \includegraphics[width=0.19\linewidth]{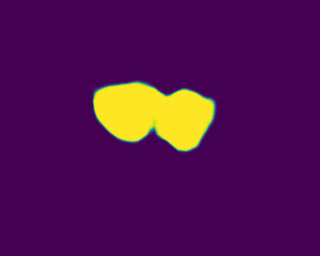}
    \includegraphics[width=0.19\linewidth]{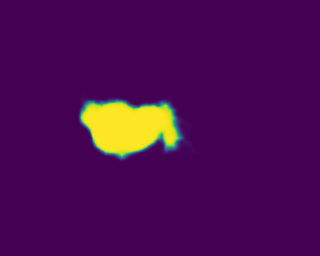}
    \includegraphics[width=0.19\linewidth]{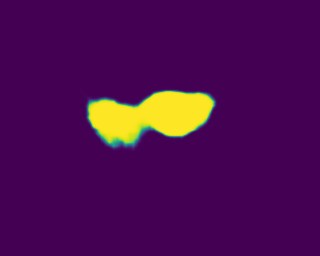}
    \\[-1mm]
    \includegraphics[width=0.19\linewidth]{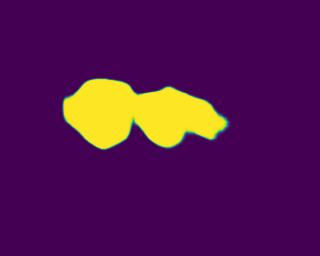} 
    \includegraphics[width=0.19\linewidth]{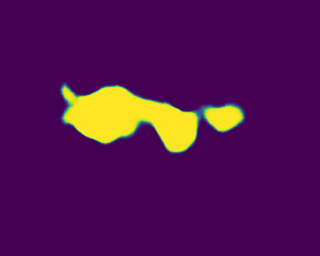}
    \includegraphics[width=0.19\linewidth]{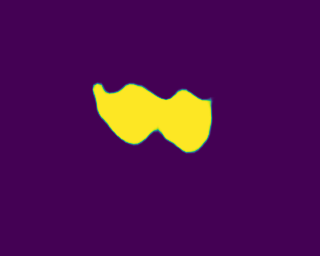}
    \includegraphics[width=0.19\linewidth]{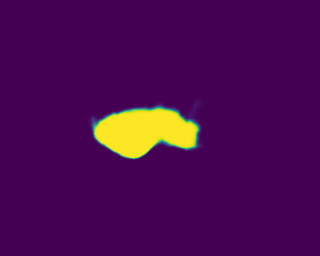}
    \includegraphics[width=0.19\linewidth]{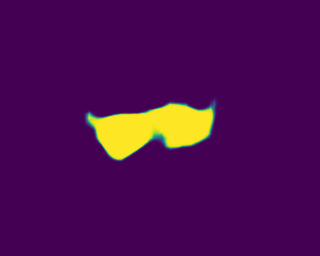}
    \caption{Additional inference on out-of-distribution samples captured during a summer 2026 research trip to South Africa using SegFormer MiT-B3 CLAHE $c=25$ (middle row) and $c=10$ (bottom row). Best viewed in color.}
    \label{fig:summer_2026_2} 
\end{figure*}

Finally, Figure~\ref{fig:summer_2026_2} provides additional out-of-distribution mask heatmap inferences for Bovidae teeth photos captured in the summer of 2026. The segmentation model utilized is SegFormer MiT-B3 with CLAHE $c\in{10, 25}$ (see caption). We see that both clip $c$ values are able to competently draw mask heatmaps, however, $c=25$ has a slight edge on certain samples, such as the first, second, and final image, where its heatmap seems more well-defined and a better match to the actual photo.

\begin{table*}[t!]
    \centering
    \scalebox{0.85}{
    \begin{tabular}{l|cccc|cccc} \toprule
    \textbf{Metric} &  \multicolumn{4}{c|}{\textbf{Dice}} & \multicolumn{4}{c}{\textbf{mIoU}} \\ \midrule
    \textbf{Tribe} & \textbf{CLAHE 25} & \textbf{CLAHE 10} & \textbf{HE} & \textbf{NE} & \textbf{CLAHE 25} & \textbf{CLAHE 10} & \textbf{HE} & \textbf{NE} \\ \midrule 
    Alcelaphini (665)  & $\mathbf{0.8710}_{0.019}$ & $0.8708_{0.015}$ & $0.8658_{0.017}$ & $0.8659_{0.015}$ & $\mathbf{0.7718}_{0.031}$ & $0.7714_{0.023}$ & $0.7637_{0.027}$ & $0.7637_{0.024}$ \\
    Antilopini (180)   & $0.8571_{0.015}$ & $\mathbf{0.8598}_{0.012}$ & $0.8428_{0.010}$ & $0.8570_{0.003}$ & $0.7500_{0.023}$ & $\mathbf{0.7542}_{0.019}$ & $0.7284_{0.015}$ & $0.7498_{0.005}$ \\
    Bovini (138)       & $0.4018_{0.008}$ & $0.9030_{0.006}$ & $0.9000_{0.009}$ & $\mathbf{0.9032}_{0.008}$ & $0.8213_{0.013}$ & $0.8232_{0.010}$ & $0.8182_{0.015}$ & $\mathbf{0.8235}_{0.013}$ \\
    Hippotragini (483) & $\mathbf{0.8657}_{0.015}$ & $0.8651_{0.013}$ & $0.8573_{0.017}$ & $0.8630_{0.010}$ & $\mathbf{0.7634}_{0.023}$ & $0.7625_{0.021}$ & $0.7505_{0.026}$ & $0.7591_{0.016}$ \\
    Neotragini (757)   & $0.8131_{0.013}$ & $\mathbf{0.8165}_{0.005}$ & $0.8020_{0.006}$ & $0.8075_{0.013}$ & $0.6852_{0.018}$ & $\mathbf{0.6899}_{0.007}$ & $0.6696_{0.008}$ & $0.6772_{0.018}$\\
    Reduncini (489)    & $\mathbf{0.8869}_{0.008}$ & $0.8863_{0.009}$ & $0.8857_{0.007}$ & $0.8848_{0.005}$ & $\mathbf{0.7969}_{0.014}$ & $0.7959_{0.015}$ & $0.7948_{0.011}$ & $0.7936_{0.010}$ \\
    Tragelaphini (489) & $0.8929_{0.003}$ & $0.8462_{0.005}$ & $0.8916_{0.004}$ & $\mathbf{0.8970}_{0.007}$ & $0.8066_{0.004}$ & $0.8120_{0.008}$ & $0.8045_{0.007}$ & $\mathbf{0.8134}_{0.011}$\\
    \bottomrule
    \end{tabular}
    }
    \caption{Stratified per-tribe segmentation Dice metric and mean Intersection over Union (mIoU) scores for MBv2. We compare CLAHE at $c\in{10, 25}$, HE and NE. Tribes annotated with number of images. Best results in \textbf{bold}. Results averaged across 3 random seeds.}
    \label{tab:mbv2_b3_per_class}
\end{table*}

\end{document}